\documentclass[sigconf,nonacm,balance=false]{aamas}

\usepackage{multirow}
\usepackage{array}
\usepackage{fvextra}
\usepackage{needspace}
\newcommand{\PromptSource}[1]{\noindent Source file: \path{#1}.\par\smallskip}

\newif\ifarxiv
\arxivtrue
\newcommand{\mainref}[2]{#2}
\newcommand{\suppref}[2]{#2}
\newcommand{\mirror}{MirrorCraft}
\newcommand{\direct}{Direct LLM}

\title[\mirror{}: Paired Evaluation under Hidden Rule Changes in Minecraft]{\mirror{}: Paired Evaluation under Hidden \\ Rule Changes in Minecraft}

\author{Jianxin Gao}
\authornote{These authors contributed equally to this work.}
\affiliation{
  \institution{China Agricultural University}}
\email{jxgao@cau.edu.cn}

\author{Beini Hu}
\authornotemark[1]
\affiliation{
  \institution{Beijing Normal University}}
\email{beinihu160714@gmail.com}

\author{Runze Li}
\authornotemark[1]
\affiliation{
  \institution{Jilin University}}
\email{rzli5524@mails.jlu.edu.cn}

\author{Hongjian Fang}
\affiliation{
  \institution{Beijing National Research Center for Information Science and Technology, Tsinghua University}}
\email{nuaalczd@gmail.com}

\author{Wanli Peng}
\authornote{Corresponding author.}
\affiliation{
  \institution{China Agricultural University}}
\email{wlpeng@cau.edu.cn}

\renewcommand{\shortauthors}{Gao, Hu, Li, Fang, and Peng}

\makeatletter
\def\@mkauthors{%
  \global\setbox\mktitle@bx=\vbox{%
    \unvbox\mktitle@bx
    \kern4pt
    \hbox to\textwidth{\hfil
      \begin{tabular}{@{}c@{}}
      {\fontsize{12}{14.5}\selectfont
      Jianxin Gao\textsuperscript{1,\@fnsymbol{1}}\quad
      Beini Hu\textsuperscript{2,\@fnsymbol{1}}\quad
      Runze Li\textsuperscript{3,\@fnsymbol{1}}\quad
      Hongjian Fang\textsuperscript{4}\quad
      Wanli Peng\textsuperscript{1,\@fnsymbol{2}}}\\[3pt]
      {\fontsize{10}{12.5}\selectfont
      \textsuperscript{1}China Agricultural University\qquad
      \textsuperscript{2}Beijing Normal University\qquad
      \textsuperscript{3}Jilin University}\\[1pt]
      {\fontsize{10}{12.5}\selectfont
      \textsuperscript{4}Beijing National Research Center for Information Science and Technology, Tsinghua University}\\[3pt]
      {\fontsize{9}{11}\selectfont
      \href{mailto:jxgao@cau.edu.cn}{jxgao@cau.edu.cn}\quad
      \href{mailto:beinihu160714@gmail.com}{beinihu160714@gmail.com}\quad
      \href{mailto:rzli5524@mails.jlu.edu.cn}{rzli5524@mails.jlu.edu.cn}}\\[1pt]
      {\fontsize{9}{11}\selectfont
      \href{mailto:nuaalczd@gmail.com}{nuaalczd@gmail.com}\quad
      \href{mailto:wlpeng@cau.edu.cn}{wlpeng@cau.edu.cn}}
      \end{tabular}%
    \hfil}%
    \kern8pt
  }%
}
\makeatother

\begin{abstract}
With the rapid progress of large language models (LLMs), an interesting question has emerged: how do LLM-based agents perform in Minecraft? To date, most Minecraft benchmarks evaluate them under fixed game mechanics, which are extensively documented online. High performance in these settings does not show whether an agent can continue making progress when familiar recipes, drops, and other rules change. In this paper, we introduce MirrorCraft, a paired benchmark for evaluating agents under hidden rule changes in Minecraft. Each Mirror world is a copy of its paired Vanilla world, with selected server-side rules silently modified by the datapack of one rule suite. Terrain, spawn, resource placement, task, interface, and action budget remain matched within every Vanilla-Mirror pair. MirrorCraft includes 10 Vanilla worlds in five controlled biomes, six rule suites, and three progression tasks, and we evaluate two LLMs and six agent configurations under a shared Mineflayer interface. We measure progress with server-verified advancement milestones and completion with success rate, and define the Rule Intervention Effect (RIE) as the performance change between paired Vanilla and Mirror worlds. Experiments with 8,640 episodes show that hidden rule changes have sharply different effects across suites. Among the configurations evaluated without rule descriptions, ReAct, Reflexion, and Voyager achieve the highest pooled Mirror Scores, but ReAct loses the most under compressed resource drops. Providing ReAct with the exact rules yields modest gains in average progress and completion across all three tasks, yet most of the difference between suites remains once the rules are known. MirrorCraft extends Minecraft evaluation beyond fixed mechanics and provides a controlled setting for studying how agents use gameplay outcomes when the rules of the current world differ from familiar ones.
\end{abstract}


\begin{document}

\makeatletter
\fancypagestyle{firstpagestyle}{%
  \fancyhf{}%
  \renewcommand{\headrulewidth}{0pt}%
  \renewcommand{\footrulewidth}{0pt}%
  \fancyfoot[C]{\footnotesize\thepage}}
\fancypagestyle{standardpagestyle}{%
  \fancyhf{}%
  \renewcommand{\headrulewidth}{0pt}%
  \renewcommand{\footrulewidth}{0pt}%
  \fancyhead[L]{\@headfootfont\shorttitle}%
  \fancyhead[R]{\@headfootfont\shortauthors}%
  \fancyfoot[C]{\footnotesize\thepage}}
\makeatother
\pagestyle{standardpagestyle}

\setlength{\emergencystretch}{1.5em}

\maketitle

\hypersetup{%
  pdftitle={MirrorCraft: Paired Evaluation under Hidden Rule Changes in Minecraft},
  pdfauthor={Jianxin Gao, Beini Hu, Runze Li, Hongjian Fang, Wanli Peng},
  pdfkeywords={LLM Agents, Minecraft, Benchmark, Hidden Rule Changes, Paired Evaluation}}

\section{Introduction}
\label{sec:intro}

Minecraft is widely played and documented online~\cite{minedojo}, so LLM-based agents may rely on mechanics memorized during pretraining rather than infer the rules from observed outcomes~\cite{adam}.
This distinction matters when a recipe, drop, or tool requirement changes: a wrong assumption can invalidate the plans built on it.
Agents outside games face similar risks when tools change between versions or teammates follow different conventions.
Rule changes can also lengthen or shorten task routes, so a performance difference combines the effect of the new route with the cost of not knowing the new rules.
Evaluation therefore needs three quantities: performance in the changed world, its difference from performance in the same world under standard rules, and the part of that difference removed by disclosing the new rules.
The latter two require worlds that differ only in their rules and a condition that discloses the change.

\begin{table}[t]
\caption{Representative benchmark protocols. ``Online'' denotes evaluation through interaction with the environment rather than on recorded episodes. ``Matched copies'' denotes separate standard and changed runs initialized from copies of the same world; MC denotes Minecraft.}
\label{tab:compare}
\centering
{\small
\setlength{\tabcolsep}{2.2pt}
\renewcommand{\arraystretch}{1.25}
\begin{tabular}{@{}lcccc@{}}
\toprule
Benchmark(s) & World & Online & Change & \shortstack{Matched\\copies}\\
\midrule
MineRL / MineDojo & MC & $\checkmark$ & -- & --\\
MCU / MineExplorer & MC & $\checkmark$ & -- & --\\
SciCrafter & MC & $\checkmark$ & -- & --\\
\addlinespace[1pt]
Mars & Crafter & $\checkmark$ & $\checkmark$ & --\\
NovelCraft & PolyCraft & -- & $\checkmark$ & --\\
NovelGym & Grid & $\checkmark$ & $\checkmark$ & --\\
\midrule
\mirror{} & MC & $\checkmark$ & $\checkmark$ & $\checkmark$\\
\bottomrule
\end{tabular}}
\end{table}

\begin{figure*}[!t]
\centering
\includegraphics[width=0.99\textwidth]{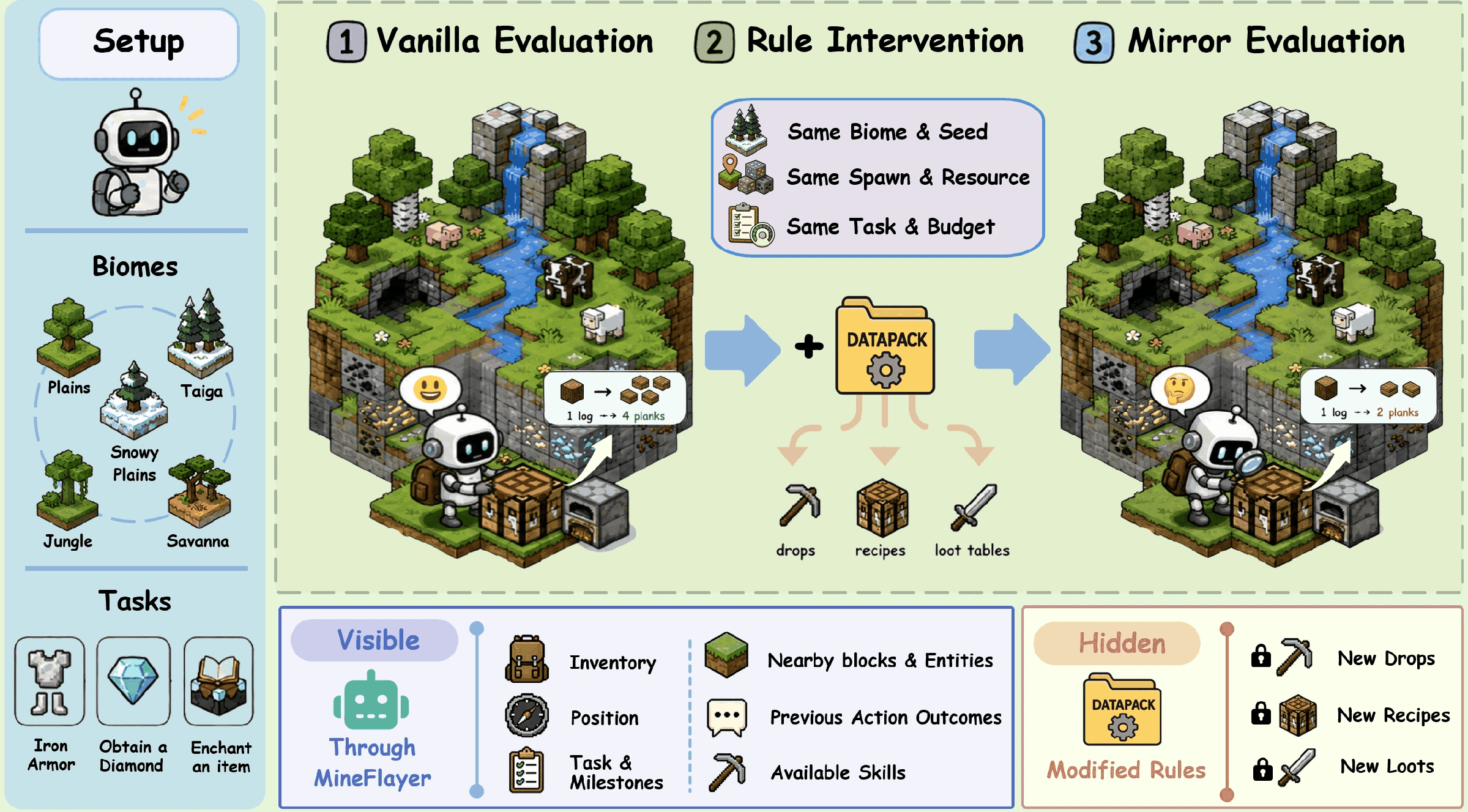}
\caption{Overview of benchmark construction and paired evaluation in Vanilla and Mirror worlds.
The paired Mirror world is copied from the Vanilla save, evaluated under matched controls, and loads the datapack of its rule suite.}
\Description{A three-stage diagram. The setup lists five biomes and three tasks. Stage one shows a Vanilla world in which one log yields four planks. Stage two shows a datapack that modifies drops, recipes, and loot tables while the biome, seed, spawn, resources, task, and budget stay the same. Stage three shows the Mirror copy of the same world in which one log yields two planks. The bottom row separates what the agent observes through Mineflayer from the modified rules that remain hidden.}
\label{fig:pipeline}
\end{figure*}

Existing Minecraft benchmarks such as MineRL, MineDojo, MCU, MineExplorer, and SciCrafter cover diverse tasks, generalization, exploration, and parameterized discovery, but they keep the game mechanics fixed~\cite{minerl,minedojo,mcu,minexplorer,scicrafter}.
Mars, NovelGym, and NovelCraft evaluate changed mechanics or novelties against a default condition, and XENON and ADAM study agents that revise task knowledge after recipes or dependencies are modified~\cite{mars,novelgym,novelcraft,xenon,adam}.
Table~\ref{tab:compare} distinguishes online interaction, changed mechanics, and matched world copies.
NovelGym compares performance before and after a novelty in one environment, retaining earlier experience; Mars samples a new world for each episode; and NovelCraft provides recorded episodes.
\mirror{} starts standard and changed runs from copies of the same save and resets agent state, excluding both world differences and carryover experience from the comparison.

We introduce \mirror{}, a paired benchmark for evaluating agents under hidden rule changes in Minecraft Java 1.19.
Each Vanilla world follows the standard rules and is copied into six Mirror worlds, each of which loads the datapack of a different rule suite.
A Vanilla world and its Mirror copies share terrain, spawn, placed resources, task, observation and action interfaces, and action budget.
Ordinary actions reveal their outcomes, but the agent is never told that any rule has changed (Figure~\ref{fig:pipeline}).

\mirror{} contains 10 Vanilla worlds in five controlled biomes, six rule suites, and three progression tasks: \emph{Iron Armor}, \emph{Diamond}, and \emph{Enchantment}.
The main study evaluates two models and six agent configurations through the same Mineflayer body and semantic skills, and a separate disclosure condition gives ReAct the exact Mirror rules.
Server-verified milestones yield a Score for partial progress, and the success rate (SR) records whether the final milestone is reached.
The two Rule Intervention Effects, $\mathrm{RIE}_{\mathrm{SC}}$ and $\mathrm{RIE}_{\mathrm{SR}}$, subtract the paired Mirror result from the Vanilla result for the same world, model, task, agent configuration, and action budget.
Together, Mirror performance, RIE, and the disclosure contrast address the three quantities above.
Within-suite comparisons then show how agent configurations respond to the same rule changes.

In the 8,640-episode main study, ReAct, Reflexion, and Voyager achieve the highest Mirror performance under the main action budgets, but the configurations with the smallest pooled RIE are not the strongest in Mirror worlds.
The scarcity and compressed-drop suites lower SR by 23.5 and 19.0 percentage points relative to Vanilla, whereas the three suites that expand yields or add byproducts, and thereby shorten routes, barely change it.
For ReAct, not being told the rules costs 5.6 points of SR with no detectable difference between suites, whereas the gap that remains with exact rules varies widely: disclosure removes ReAct's entire drop under route substitution, but a 32-point gap remains under compressed drops, although unpacking a compressed drop takes a single crafting call.
Within a suite, the configurations that perform well in Vanilla worlds all lose substantially under scarcity, whereas under compressed drops they differ significantly and ReAct loses the most.

Our contributions are a matched-world Minecraft benchmark with hidden rule changes; an evaluation protocol combining a common interface, server-verified outcomes, paired RIE measures, and a disclosure contrast; and an empirical comparison of agent configurations under the same interventions.
The study identifies compressed resource drops as a revealing case: adding at most one crafting call per resource produces a loss comparable to scarcity, separates the compared configurations, and leaves ReAct with a substantial gap even when the rules are disclosed.
\section{Related Work}
\label{sec:related}

\paragraph{Minecraft agents.}
Minecraft agents combine language-model planning with reusable skills, feedback, and structured knowledge~\cite{deps,voyager,gitm,jarvis1,odyssey,metagentp,li2026ppa}.
Goal-Oriented Graphs retrieve prerequisite chains from Minecraft knowledge sources~\cite{gogminecraft}.
The methods evaluated here use different forms of episode history: ReAct retains reasoning, actions, and observations; Reflexion adds verbal reflections; Voyager retrieves reusable procedures; and XENON and ADAM revise task relations from outcomes~\cite{react,reflexion,voyager,xenon,adam}.
We evaluate \direct{} and implementations of these five methods on the same Mineflayer interface~\cite{mineflayer}, with shared observations, semantic skills, action budgets, and state resets between episodes.
Agents that act on pixels instead emphasize perception and control with primitive actions~\cite{optimus1,jarvisvla,mrsteve}.

\paragraph{Minecraft capability benchmarks.}
Existing benchmarks emphasize task coverage, generalization, or specific capabilities.
MineRL and MineDojo provide diverse tasks and demonstrations~\cite{minerl,minedojo}; MCU studies compositional generalization, and CrafText tests instruction following under new phrasings and task configurations~\cite{mcu,craftext}.
SmartPlay probes abilities across six games, Crafter uses achievements to cover several abilities, and OpenHA compares action abstractions across more than 800 tasks~\cite{smartplay,crafter,openha}.
MineExplorer emphasizes exploration by filtering out knowledge-heavy tasks, while SciCrafter connects discovery with application through parameterized redstone construction~\cite{minexplorer,scicrafter}.
More focused environments test whether a crafting plan is feasible or whether an agent can complete spatial construction~\cite{plancraft,mineanybuild}.
These benchmarks broaden task coverage or isolate specific capabilities but generally keep the mechanics fixed, whereas \mirror{} changes selected mechanics that familiar progression tasks require.

\paragraph{Common interfaces and progress measures.}
BrowserGym provides a common observation and action interface for web agents, and AndroidWorld combines dynamic tasks with programmatic initialization and success checks~\cite{browsergym,androidworld}.
AgentBoard complements binary success with graded progress, and GameWorld compares raw and semantic game interfaces using outcomes verified from environment state~\cite{agentboard,gameworld}.
Trace analyses of multi-agent systems further show that correct answers can hide coordination breakdowns~\cite{cemri2025mast,so2026superficial}.
ProEvolve extends such evaluation to changes between versions of programmable tool environments~\cite{proevolve}.
\mirror{} likewise uses one runtime and server-verified outcomes, so the agent body and success judge stay fixed across configurations and between Vanilla and Mirror worlds.

\paragraph{Changed mechanics and novelty.}
Research on open-world AI studies how agents detect and accommodate sudden changes outside their prior experience~\cite{holder2025openworld}.
NovelGym combines changes to entities, recipes, actions, transition dynamics, and costs in a Minecraft-inspired gridworld, measuring novelty impact as the resulting performance change~\cite{novelgym}.
NovelCraft provides multimodal episodes from PolyCraft, a modified Minecraft environment for novelty detection and category discovery, while NovelGridworlds evaluates planning and learning after a novelty is introduced~\cite{novelcraft,novelgridworlds}.
Alchemy and XLand-MiniGrid hide causal structures or production rules that agents must infer through interaction~\cite{alchemy,xlandminigrid}.
Mars is the closest benchmark: it changes terrain, survival settings, and task dependencies in Crafter, a two-dimensional game inspired by Minecraft, to study reasoning when mechanics contradict familiar expectations~\cite{mars}.
XENON revises dependency and action knowledge from experience, while WALL-E learns symbolic rules from trajectories for world-model alignment~\cite{xenon,walle2025}.
Memory systems likewise reuse trajectories, insights, or plans from similar past tasks~\cite{zhang2025gmemory,wang2026meco}, but agents tend to follow retrieved records, so a wrong record can propagate errors~\cite{xiong2026memory,tian2026madmm}.
Counterfactual tasks alter the default assumptions of familiar tasks to test whether language models reason or recite~\cite{wu2024reasoning}, and studies of knowledge conflicts show that models can fall back on parametric knowledge when the context contradicts it~\cite{longpre2021entity,xie2024adaptive}.
Code over Words pairs identical Baba Is You states with contradictory rules stated in the prompt and finds that models struggle to set aside the priors that these rules contradict~\cite{xu2026codewords}.
These studies motivate evaluation under unfamiliar rules, but use different controls.
Several include a default or pre-novelty condition, and Code over Words pairs states under contradictory rules; none combines hidden rule changes with separate runs from copies of the same Minecraft world.
Like Mars, we compare ReAct with and without rule descriptions.
The unchanged world copy additionally separates the effect removed by disclosure from the gap that remains.
\section{The \mirror{} Benchmark}
\label{sec:bench}

Each Vanilla world is copied into six paired Mirror worlds, one per rule suite (M01--M06), and everything but the rules stays fixed within each pair.

\subsection{Vanilla and Mirror Worlds}

We construct two Vanilla worlds in each of five controlled biomes: Plains, Taiga, Snowy Plains, Jungle, and Savanna.
A biome datapack defines the surface blocks, vegetation, water margins, and all resources required by the tasks, and it is identical in each Vanilla world and its copies.
A deterministic compiler turns a versioned specification of terrain relief, surface materials, resource bands, vegetation, and passive mobs into each biome's datapack for Minecraft Java 1.19.0.
Added ore bands keep diamond and obsidian reachable in every biome without placing items in inventories or containers.
The Vanilla condition thus refers to the standard recipes and drops, not to unmodified world generation or game settings.
Two worlds per biome yield 10 Vanilla worlds with validated solution routes, and six copies of each yield 60 paired Mirror worlds.
Each episode starts from a fresh world copy, which resets inventory, advancements, entities, dropped items, and block state; agent state is cleared separately.

\subsection{Rule Suites}

Each rule suite changes the availability of materials or the route needed to complete a progression task.
A valid intervention stays undisclosed in the hidden setting, produces an observable outcome when triggered, affects a dependency used by at least one task, and preserves a complete route to every task.

A static program traces milestone dependencies and identifies the block drops, entity loot, and crafting or processing recipes that can be edited.
An LLM designer proposes candidate manifests, and an LLM judge screens them for relevance, observability, coherence, and diversity; both use GPT-5.6 Sol~\cite{gpt56sol} and receive only specifications and static reports, never agent outcomes.
The designer is instructed to make each changed transition encounterable while an agent pursues at least one task, to prefer two to six interpretable changes, and to keep every change learnable from action outcomes rather than from inspecting recipes or loot tables.
Candidates that fail compilation or receive a revision request from the judge return to the designer for another bounded round, and a manifest is accepted only when it compiles and the judge reports no hard failure and deems every task solvable.
The static program compiles accepted manifests into Java 1.19 datapacks, after which deterministic tests verify loading, declared outcomes, hidden-rule integrity, and complete solution routes.
No LLM evaluates agent episodes or determines success.
We froze all benchmark worlds and rule suites before inspecting any agent-evaluation results and did not use those results to select or revise them.

\begin{figure}[!t]
\centering
\includegraphics[width=\columnwidth]{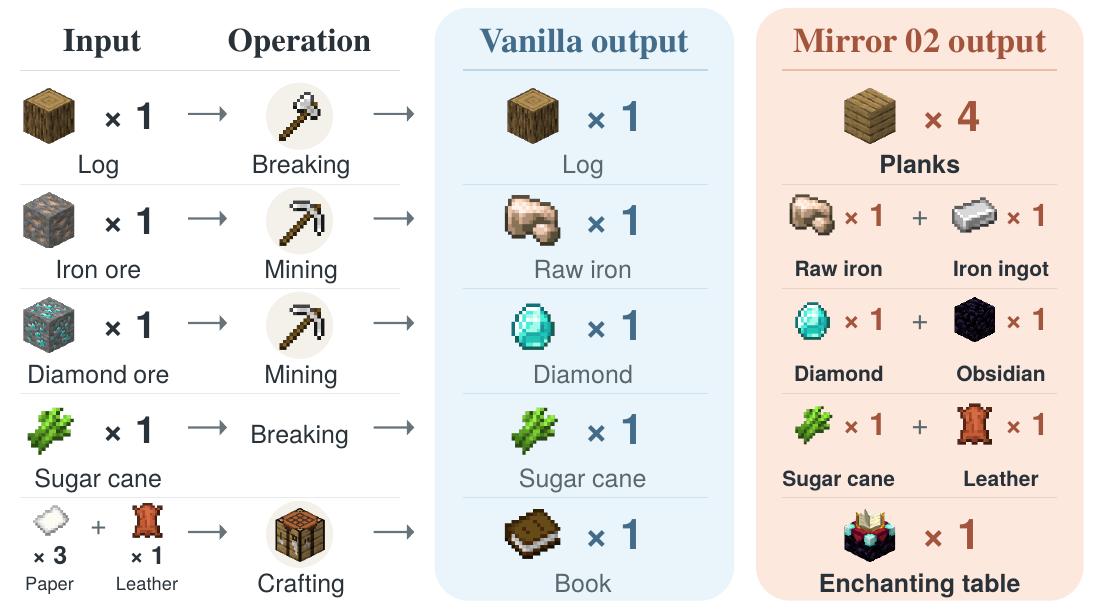}
\caption{Changed transitions in M02 (byproduct shortcuts). Each row applies an operation to an input and compares the Vanilla output with the M02 output.}
\Description{A table of Minecraft items with four columns: input, operation, Vanilla output, and Mirror 02 output. Breaking a log yields a log in Vanilla and four planks under M02. Mining iron ore yields raw iron in Vanilla and raw iron plus an iron ingot under M02. Mining diamond ore yields a diamond in Vanilla and a diamond plus obsidian under M02. Breaking sugar cane yields sugar cane in Vanilla and sugar cane plus leather under M02. Crafting three paper and one leather yields a book in Vanilla and an enchanting table under M02.}
\label{fig:m02}
\end{figure}

The six suites implement five forms of route change.
M01 creates quantity scarcity: one log yields two planks instead of four, and raw iron smelts into three nuggets instead of an ingot.
M02 and M06 turn useful task items into byproducts of ordinary actions, as the example below shows for M02.
M03 mainly expands yields: a log yields six planks, raw iron smelts into two ingots, and diamond ore drops two diamonds and one lapis lazuli. It also substitutes two outputs: three sugar canes yield a book, and the diamond pickaxe recipe yields a netherite pickaxe.
M04 replaces coal, iron, diamond, and lapis ore drops with their corresponding storage blocks, such as a block of raw iron. One crafting step unpacks a storage block into nine ordinary items; cows also drop a bookshelf.
M05 substitutes routes: the furnace recipe yields a blast furnace, and the iron and diamond pickaxe recipes yield hoes that must be smelted or blasted into the corresponding pickaxes.
M06 adds byproducts that the Enchantment route needs: logs drop planks and sticks, iron ore also drops paper, diamond ore also drops lapis lazuli, and lapis ore also drops obsidian.
Edits such as the bookshelf in M04 or the book crafted from sugar cane in M03 lie only on the Enchantment route; on the Iron Armor and Diamond routes, every changed transition belongs to the form that names its suite.

Figure~\ref{fig:m02} illustrates the construction with M02.
Tracing the milestone dependencies shows that every task needs planks and iron ingots, and that Enchantment also needs an enchanting table, which is normally crafted from a book, two diamonds, and four obsidian.
The accepted M02 manifest turns several of these requirements into byproducts or substituted outputs of ordinary actions, from planks dropped by logs to an enchanting table crafted from the book recipe.
Each edit lies on the route to at least one task and is observable, because mining reports the collected drops and crafting reports the produced item.
Each edit also preserves a route, because every familiar item either remains available or is replaced by the item it would have been used to make.
In the hidden setting, an agent gains from these shortcuts only if it notices the extra items and changes its plan; otherwise it may, for example, keep smelting raw iron even though mining already yields ingots.
As a result, M02 shortens the routes of all three tasks, most of all for Enchantment, whose enchanting table no longer requires diamonds or obsidian.

\subsection{Information and Action Interface}

Every configuration uses the same Mineflayer interface and receives the same observation fields: inventory, position, task milestones, nearby blocks and entities, action outcomes, available skills, equipped items, survival state, and biome (Figure~\ref{fig:pipeline}).
The observation also includes nearby dropped items and the next milestone.

In the hidden setting, the prompt is the same as in Vanilla episodes and never states that the rules may differ, so the agent can infer a modification only from the observed outcome of an affected action.
Crafting reports the produced item, mining reports the broken block and the collected drops, attacks report the collected loot, and processing reports the change in inventory.
The agent cannot query recipes, loot tables, processing rules, datapacks, or server registries.

The Mineflayer body provides skills for exploration, mining, collection, crafting, smelting, combat, enchanting, and limited obstacle recovery.
The \texttt{craft\_grid} skill takes an explicit grid and workstation rather than a desired output, so it exposes whatever item the current rules produce.
During a call, the body may equip an available tool, collect nearby drops, or clear at most one local obstruction; it cannot create missing items, select semantic subgoals, inspect rules, override the chosen action, or retry indefinitely.

\subsection{Progression Tasks and Scoring}

We evaluate three increasingly deep tasks along one Minecraft progression route: \emph{Iron Armor}, \emph{Diamond}, and \emph{Enchantment}.
Each task has an ordered sequence of server-verified milestones, which are built-in advancements of the game, and each milestone has a fixed value.
Iron Armor proceeds through stone (15), a stone pickaxe (35), an iron ingot (65), and any one iron armor piece (100).
Diamond proceeds through stone (10), a stone pickaxe (25), an iron ingot (50), an iron pickaxe (70), and a diamond (100), and Enchantment follows the same route with values 10, 25, 45, 60, and 75 before obsidian (90) and an enchanted item (100).
Score records the deepest completed milestone, and success rate (SR) records whether the final milestone, which has value 100, is reached.

Let $\mathcal{C}_t$ be the milestone set for task $t$, let $w_t(c)$ be the value of milestone $c$, and let $\mathcal{A}(e)$ be the advancements completed in episode $e$.
The episode Score and the aggregate SR are
\begin{align}
\operatorname{Score}(e,t) & = \max\left(\{0\}\cup\{w_t(c):c\in\mathcal{A}(e)\cap\mathcal{C}_t\}\right), \\
\operatorname{SR}(t) & = \frac{1}{N_t}\sum_{e=1}^{N_t}\mathbf{1}[\operatorname{Score}(e,t)=100].
\end{align}
Using only the deepest milestone avoids treating milestones on the same route as independent rewards, and a task completed through a changed route that skips an intermediate milestone still receives 100.
The milestones are identical across conditions, so Score measures progress along the Vanilla sequence. We therefore interpret it alongside SR, which does not depend on the route.
$N_t$ includes every valid episode, including deaths, exhausted budgets, and failures to produce an executable action, and the three tasks receive equal weight in pooled results.

\subsection{Rule Intervention Effects}
\label{sec:rie}

Score and SR measure performance within a Vanilla or Mirror world.
The Rule Intervention Effects quantify the difference in performance between a Vanilla world and a Mirror world copied from it, with the model, task, agent configuration, interfaces, and action budget held fixed.
For each Vanilla world, model, task, and agent configuration, we first average the three Vanilla repetitions.
We separately average the three repetitions in each paired Mirror world and compare that mean with its Vanilla counterpart.
Let $\mathcal P$ contain the resulting comparisons.
For comparison $p\in\mathcal P$, let $\bar S_p^{\mathrm V}$ and $\bar S_p^{\mathrm M}$ be the mean Scores in the Vanilla world and its paired Mirror world, and let $\bar Y_p^{\mathrm V}$ and $\bar Y_p^{\mathrm M}$ be the corresponding mean success percentages.
We define
\begin{align}
\mathrm{RIE}_{\mathrm{SC}} & = \frac{1}{|\mathcal P|}\sum_{p\in\mathcal P}\left(\bar S_p^{\mathrm V}-\bar S_p^{\mathrm M}\right), \\
\mathrm{RIE}_{\mathrm{SR}} & = \frac{1}{|\mathcal P|}\sum_{p\in\mathcal P}\left(\bar Y_p^{\mathrm V}-\bar Y_p^{\mathrm M}\right).
\end{align}
Positive values indicate lower performance in the Mirror world, and negative values indicate higher performance.
We report both measures because an intervention can affect partial progress and final completion differently.
RIE plays the role of the novelty impact used in open-world evaluation~\cite{novelgym}, but it compares two copies of the same world rather than periods before and after a novelty.

RIE includes every valid episode, even when the agent never attempts an affected action.
It therefore measures the overall effect of a rule suite, which combines changes in route requirements, exposure to modified outcomes, and the subsequent use of those outcomes.
The disclosure condition, the within-suite comparison, and the behavioral diagnostics below examine parts of it.

To measure the value of direct rule access, we follow Mars~\cite{mars}, which gives the modified rules to ReAct, and compare ReAct under two disclosure conditions in paired Mirror worlds.
ReAct receives no rule descriptions, whereas ReAct w/rules is told that the rules have changed and receives an exact description of them; the two otherwise share the task, interface, action budget, and implementation.
We define $\Delta\mathrm{Score}$ and $\Delta\mathrm{SR}$ as the disclosed result minus the hidden result.
Because every Mirror world has a Vanilla twin, ReAct's $\mathrm{RIE}_{\mathrm{SR}}$ splits into the mean of $\bar Y_p^{\mathrm V}-\bar Y_p^{\mathrm D}$ plus $\Delta\mathrm{SR}$, where $\bar Y_p^{\mathrm D}$ is the mean success percentage of ReAct w/rules in the Mirror world of comparison $p$, and likewise for Score.
$\Delta\mathrm{SR}$ is the cost of not being told that and how the rules changed, and the first term is the gap that remains when they are known, which combines route changes under the fixed budget with any difficulty in applying the disclosed rules.

Disclosure isolates the cost of hidden rules for ReAct.
Because a suite changes the rules identically for every configuration, differences between their RIEs within a suite show how they differ under the same change, both in how often their plans reach the changed transitions and in what they do afterwards.
Because $\mathrm{RIE}_{\mathrm{SR}}$ cannot exceed the Vanilla SR, a configuration that rarely succeeds in Vanilla worlds can only show a small $\mathrm{RIE}_{\mathrm{SR}}$.
We therefore compare, on Iron Armor and Diamond, the configurations whose Vanilla SR on both tasks is at least 50\%, a set that any threshold from 40\% to 60\% leaves unchanged, and test whether their RIEs differ within a suite with Cochran's $Q$ heterogeneity statistic.

\subsection{Behavior after Rule Encounters}
\label{sec:behavior}

Score, SR, and RIE characterize task outcomes but not the actions that follow a modified gameplay outcome.
A server-side evaluator, invisible to the agent, records every action that triggers such an outcome, and the first such action for each changed rule in an episode defines a rule encounter.
We map each action to a semantic signature: the workstation and grid for \texttt{craft\_grid}, the input and device for \texttt{smelt}, the block type for \texttt{mine}, the mob type for \texttt{attack}, the item and use for \texttt{place}, the item and option for \texttt{enchant}, and the skill name for all other skills, ignoring quantities, fuel, coordinates, and entity identity.
Among encounters followed by another budgeted action, the Immediate Repeat Rate (IRR) is the fraction whose next action has the same signature as the encounter action; whether such repetition helps depends on the rule.
Recovery Latency (RL) is the mean number of semantic actions from an encounter to the first Score increase above its pre-encounter value.
It is zero when the encounter action itself raises the Score and excludes encounters after which the Score never rises.
RL therefore describes how quickly progress resumes, conditional on its resuming.
Because RL also depends on how far a changed transition lies from the next milestone, we compare it within a suite.

\subsection{Integrity and Reachability Controls}
\label{sec:integrity}

Before evaluation, we verify the game version, datapack loading, advancements, spawn conditions, and placed resources in every Vanilla and Mirror world, together with an executable route for every task under every rule suite.
Reachability does not require equal route lengths: an intervention may lengthen or shorten a route while the action budget stays fixed.

For each Vanilla world, a scripted test executes the same action sequence there and in its paired Mirror worlds without triggering a changed rule, and the serialized observations must match.
The test also verifies that the interface cannot query server rules, rule text, or an identifier of the paired Mirror world.
Missing datapacks, unavailable advancements, and disconnections are treated as infrastructure faults and rerun with the same episode identifier, whereas deaths, exhausted budgets, ineffective strategies, and policy errors remain evaluation outcomes.
\section{Experiments}
\label{sec:experiments}

\subsection{Experimental Setup}

\begin{table*}[!t]
\caption{Results by configuration and disclosure condition, pooled over three tasks and 10 Vanilla worlds; Mirror results also pool M01--M06.
Vanilla and Mirror report Score\,/\,SR, with SR in percent and $\mathrm{RIE}_{\mathrm{SR}}$ in percentage points.
Actions and Cost are averaged over Mirror episodes, and Cost is in US dollars.}
\label{tab:main}
\centering
{\small
\setlength{\tabcolsep}{2.2pt}
\renewcommand{\arraystretch}{1.18}
\begin{tabular*}{\textwidth}{@{\extracolsep{\fill}}lcccccc@{\hspace{6pt}}cccccc@{}}
\toprule
& \multicolumn{6}{c}{Gemini} & \multicolumn{6}{c}{DeepSeek}\\
\cmidrule(lr){2-7}\cmidrule(lr){8-13}
Configuration & Vanilla & Mirror & $\mathrm{RIE}_{\mathrm{SC}}$ & $\mathrm{RIE}_{\mathrm{SR}}$ & Actions & Cost
& Vanilla & Mirror & $\mathrm{RIE}_{\mathrm{SC}}$ & $\mathrm{RIE}_{\mathrm{SR}}$ & Actions & Cost\\
\midrule
\direct{} & 70.6\,/\,13.3 & 65.1\,/\,12.4 & $+5.5$ & $+0.9$ & 56.1 & 0.05
& 51.3\,/\,25.6 & 49.1\,/\,9.3 & $+2.3$ & $+16.3$ & 56.0 & 0.01\\
ReAct & 94.4\,/\,66.7 & 89.7\,/\,54.8 & $+4.7$ & $+11.9$ & 50.8 & 0.10
& 90.4\,/\,61.1 & 83.8\,/\,51.1 & $+6.5$ & $+10.0$ & 51.5 & 0.04\\
Reflexion & 92.8\,/\,63.3 & 90.0\,/\,57.4 & $+2.9$ & $+5.9$ & 50.6 & 0.10
& 83.0\,/\,55.6 & 78.7\,/\,46.3 & $+4.3$ & $+9.3$ & 52.0 & 0.04\\
Voyager & 92.1\,/\,62.2 & 88.3\,/\,56.1 & $+3.8$ & $+6.1$ & 50.0 & 0.07
& 78.3\,/\,52.2 & 76.3\,/\,42.6 & $+2.0$ & $+9.6$ & 52.9 & 0.03\\
XENON & 73.6\,/\,12.2 & 69.6\,/\,12.4 & $+4.0$ & $-0.2$ & 56.0 & 0.12
& 83.9\,/\,52.2 & 83.1\,/\,43.1 & $+0.7$ & $+9.1$ & 52.1 & 0.05\\
ADAM & 75.2\,/\,24.4 & 71.2\,/\,18.3 & $+4.0$ & $+6.1$ & 56.3 & 0.17
& 82.1\,/\,51.1 & 81.5\,/\,44.8 & $+0.6$ & $+6.3$ & 52.9 & 0.05\\
\midrule
ReAct w/rules & -- & 92.0\,/\,61.7 & -- & -- & 50.1 & 0.10
& -- & 87.2\,/\,55.4 & -- & -- & 50.6 & 0.04\\
\bottomrule
\end{tabular*}}
\end{table*}

Our main study evaluates Gemini 3.1 Flash-Lite~\cite{gemini31flashlite} and DeepSeek-V4-Flash~\cite{deepseekv4flash}.
We additionally evaluate GPT-5.6 Luna~\cite{gpt56luna} on the Diamond task outside the main study, with two repetitions per map; its Vanilla runs cover all five biomes, and its Mirror runs cover Plains and Savanna.
For each combination of model, Vanilla world, task, and configuration, we run three repetitions in the Vanilla world and three in each of its six paired Mirror worlds, and ReAct w/rules also uses three repetitions per paired Mirror world.
We disable provider-side reasoning so that explicit reasoning comes only from the agent configurations, and sample responses at temperature 0.7 with at most 2,048 tokens.

All configurations receive the same task, ordered milestones, observation fields, latest action result, skill interface, and Mineflayer execution body in Vanilla and Mirror episodes.
Each attempted Mineflayer skill consumes one semantic action; Iron Armor and Diamond allow 50 actions, and Enchantment allows 75.

The six configurations differ in the reasoning, memory, review, retrieval, and probing mechanisms they use during an episode.
\direct{} selects each action from the current observation and recent verified outcomes without an additional memory module.
ReAct retains a trace of the 16 most recent thoughts, actions, and observations~\cite{react}, and Reflexion adds up to three temporary lessons when progress stalls or actions repeatedly fail~\cite{reflexion}.
Voyager builds and revises a small library of procedures that it can retrieve during an episode~\cite{voyager}.
Our XENON implementation maintains a graph of task dependencies and updates edges that conflict with observed outcomes~\cite{xenon}, and our ADAM implementation represents causal dependencies and can invoke an intervention probe when uncertainty blocks further progress~\cite{adam}.
Their memory, retrieval, review, and probe limits, given in \suppref{the supplementary material}{Appendix~\ref{app:baselines}}, remain fixed across models and worlds.

All internal state is cleared after each episode, including the graphs that XENON and ADAM would otherwise retain.
No configuration therefore carries rule knowledge across episodes.
We retain the original method names, but our XENON, ADAM, and Voyager implementations are within-episode adaptations of methods designed to learn across episodes.
Neither XENON nor ADAM receives evaluator labels that identify incorrect dependencies, and each intervention probe is an ordinary semantic action on the current inventory.
ReAct w/rules differs from ReAct only in receiving the exact rule descriptions on every model call.
Reflection, critique, retrieval, and analysis calls do not act in Minecraft and consume no actions, but we count their API cost.

We fix the time of day, use Peaceful difficulty, and give every agent the same Haste, Speed, Resistance, and Strength effects and an initial experience level of 100; these controls shorten mining, traversal, combat, and enchanting without supplying items, disclosing rules, or skipping task dependencies.
Each episode ends when the final milestone is reached, the budget is exhausted, the agent dies, or no executable semantic action can be produced.

Across the 10 Vanilla worlds and their paired Mirror worlds, three tasks, two models, and three repetitions, the six configurations produce 1,080 Vanilla episodes and 6,480 Mirror episodes with hidden rules.
ReAct w/rules contributes another 1,080 Mirror episodes, bringing the main study to 8,640 episodes.
Each combination of model, task, and configuration thus has 30 episodes in Vanilla worlds and 30 under each rule suite.
IRR and RL are averaged equally across M01--M06.
We report 95\% intervals computed from binomial variances stratified by model and task, which treat episodes as independent rather than clustered by world, and compute Cochran's $Q$ from the same variances for each model separately.
\suppref{The supplementary material reports}{The appendices report} per-task, per-suite, and per-biome results.
Agent configurations, datapack construction, task definitions, and world specifications are detailed in \suppref{Sections D--J of the supplementary material}{Appendices~\ref{app:baselines}--\ref{app:mirror-details}}; the supplied evaluation-agent prompt templates are reproduced in \suppref{Section L}{Appendix~\ref{app:agent-prompts}}.

\subsection{Results}
\label{sec:results}

\subsubsection{Absolute Performance and Change from Vanilla}

Among the six configurations without rule descriptions, ReAct, Reflexion, and Voyager form the leading group in Mirror worlds, with pooled Mirror Scores of 82.3 to 86.8 and SRs of 49.4\% to 53.0\% (Table~\ref{tab:main}).
Their order depends on the model: ReAct leads with DeepSeek, whereas with Gemini, Reflexion and Voyager both exceed ReAct in Mirror SR (57.4\% and 56.1\% versus 54.8\%).
ADAM and XENON have the smallest pooled $\mathrm{RIE}_{\mathrm{SC}}$ ($+2.3$ and $+2.4$) and XENON the smallest pooled $\mathrm{RIE}_{\mathrm{SR}}$ ($+4.4$ percentage points), yet their pooled Mirror SRs of 31.6\% and 27.8\% remain far below the leading group's.
With 95\% intervals of about $\pm5$ points, pooled $\mathrm{RIE}_{\mathrm{SR}}$ does not separate the six configurations.
With Gemini, XENON's near-zero $\mathrm{RIE}_{\mathrm{SR}}$ reflects a Vanilla SR of only 12.2\%, whereas with DeepSeek, XENON and ADAM combine the smallest $\mathrm{RIE}_{\mathrm{SC}}$ ($+0.7$ and $+0.6$) with the highest Mirror Scores after ReAct (83.1 and 81.5).
A small RIE alone does not establish robustness; it must be interpreted alongside absolute Mirror performance.

XENON and ADAM trail the leading group especially with Gemini.
XENON revises dependencies after observed failures, and ADAM probes uncertain links~\cite{xenon,adam}; because each attempt and probe is a semantic action, both mechanisms consume the action budget, whereas graph analysis does not.
With Gemini, they spend 56.0 and 56.3 actions per Mirror episode, close to the mean budget of 58.3, compared with 50.8 for ReAct.
A separate Gemini study with a 200-action cap shows that rankings depend on the action budget (Table~\ref{tab:action-checkpoints-main}).
It runs ReAct, Reflexion, XENON, and ADAM in its own worlds from the same five biomes and records cumulative SR at actions 100, 150, and 200 of each trajectory (\suppref{details in the supplementary material}{Appendix~\ref{app:horizon}}).
Between actions 100 and 200, XENON and ADAM gain 49.4 and 43.0 points of Mirror SR, compared with 9.5 and 8.5 for ReAct and Reflexion; XENON has the highest Mirror SR at 200 actions.
Their standing also depends on the model: with GPT-5.6 Luna, XENON and ADAM reach 80.0\% SR on Diamond in Vanilla worlds, compared with 13.3\% and 36.7\% with Gemini.

Enchantment extends the Vanilla \emph{Diamond} route with a book from leather and sugar cane, a diamond pickaxe to mine obsidian, an enchanting table from two more diamonds and four obsidian, lapis lazuli, and the final enchantment.
Even with 75 actions, the highest Mirror SR under hidden rules on this task is 15.6\% (ReAct with Gemini), and the pooled Mirror Score and SR of 69.1 and 3.1\% show that agents make substantial progress without completing the task.
Vanilla SR on this task is at most 20.0\%, which bounds its $\mathrm{RIE}_{\mathrm{SR}}$ at 20 points.
We therefore emphasize Score for Enchantment; differences in pooled SR mainly reflect Iron Armor and Diamond, and Enchantment leaves substantial headroom for stronger agents.

\subsubsection{Variation Across Rule Suites and Configurations}
\label{sec:suite-results}

\begin{table}[!tb]
\caption{Performance in paired Mirror worlds and RIE (Vanilla minus Mirror) by rule suite, pooled over models, tasks, the six configurations evaluated without rule descriptions, and 10 Vanilla worlds.
SR is a percentage; $\mathrm{RIE}_{\mathrm{SR}}$ is in percentage points.}
\label{tab:rule-rie}
\centering
{\small
\setlength{\tabcolsep}{1.8pt}
\renewcommand{\arraystretch}{1.12}
\begin{tabular*}{\columnwidth}{@{\extracolsep{\fill}}llrrrr@{}}
\toprule
Suite & Rule change & Score & SR & $\mathrm{RIE}_{\mathrm{SC}}$ & $\mathrm{RIE}_{\mathrm{SR}}$\\
\midrule
M01 & Quantity scarcity & 67.2 & 21.5 & $+13.5$ & $+23.5$\\
M02 & Byproduct shortcuts & 82.1 & 45.2 & $-1.5$ & $-0.2$\\
M03 & Yield expansion & 84.5 & 48.0 & $-3.8$ & $-3.0$\\
M04 & Compressed resource drops & 67.3 & 26.0 & $+13.3$ & $+19.0$\\
M05 & Route substitution & 78.6 & 38.1 & $+2.0$ & $+6.9$\\
M06 & Byproduct redistribution & 83.5 & 45.6 & $-2.9$ & $-0.6$\\
\midrule
Pooled & All suites & 77.2 & 37.4 & $+3.4$ & $+7.6$\\
\bottomrule
\end{tabular*}}
\end{table}

\begin{figure}[!tb]
\centering
\includegraphics[width=\columnwidth]{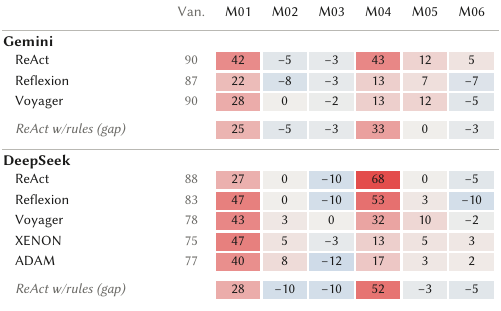}
\caption{$\mathrm{RIE}_{\mathrm{SR}}$ (Vanilla minus Mirror; red marks lower Mirror SR) by suite on Iron Armor and Diamond for configurations whose Vanilla SR (Van., which bounds RIE) is at least 50\% on both tasks; ReAct w/rules rows give the gap with exact rules.
Each cell compares 60 Vanilla and 60 Mirror episodes.
Configurations differ significantly only under M04 (Cochran's $Q$: $p=0.003$ with Gemini, $p<0.001$ with DeepSeek; $p>0.15$ elsewhere).}
\Description{A heatmap with rule suites M01 to M06 as columns and, separately for Gemini and DeepSeek, the compared configurations and ReAct with rules as rows, each cell labeled with its value. Column M01 is red for every configuration. Column M04 is dark red for ReAct with both models and light red for the other compared configurations. The other columns stay within twelve points of zero.}
\label{fig:suite-config}
\end{figure}

Quantity scarcity and compressed resource drops produce the largest declines from Vanilla, lowering SR by 23.5 and 19.0 percentage points, whereas yield expansion and the two byproduct suites, which shorten routes by construction, keep SR within 3.0 points of Vanilla; route substitution produces a smaller decline of 6.9 points (Table~\ref{tab:rule-rie}).
The pooled RIE therefore masks large differences between suites; by biome, $\mathrm{RIE}_{\mathrm{SR}}$ is positive in all five, from $+4.4$ in Jungle to $+13.0$ in Savanna.
Within a suite, the effect also depends on where the changed transitions lie on each task route.
Under route substitution, ReAct's SR on Iron Armor, whose route changes only in the smelting station, stays at its Vanilla level (90.0\% and 91.7\%, pooled over models), whereas on Diamond, which requires the hoe detour, it falls from 88.3\% to 75.0\%.

Configurations can also respond differently to the same change, so within each suite we compare the configurations with a Vanilla SR of at least 50\% on both Iron Armor and Diamond (Section~\ref{sec:rie}): ReAct, Reflexion, and Voyager with Gemini, and all but \direct{} with DeepSeek (Figure~\ref{fig:suite-config}).
Under quantity scarcity, all of them lose 22 to 47 points of SR on these tasks, their differences are not significant ($p=0.16$ with Gemini and $p=0.31$ with DeepSeek), and exact rules still leave ReAct 25 and 28 points below its Vanilla SR, compared with losses of 42 and 27 points without them.
Under compressed drops, the loss ranges from 13 points (Reflexion and Voyager with Gemini, XENON with DeepSeek) to 43 and 68 points for ReAct with Gemini and DeepSeek ($p=0.003$ and $p<0.001$, both significant after Bonferroni correction for 12 tests and with continuity-corrected variances), whereas the other four suites show no significant differences ($p>0.29$).
In each model, the two compared configurations that lose least under compressed drops keep 78\% to 85\% of their Vanilla SR, whereas ReAct keeps 52\% with Gemini and 23\% with DeepSeek, so this separation does not stem from differences in Vanilla SR; with DeepSeek, these two are XENON and ADAM, which revise explicit dependency graphs from observed outcomes.
For each model, the mean SR of the two least-affected configurations without rule descriptions still exceeds ReAct w/rules by $18.3\pm14.7$ points with Gemini and $24.2\pm14.8$ with DeepSeek (95\% intervals).

\subsubsection{Rule Disclosure and Behavioral Diagnostics}

\begin{table}[!tb]
\caption{ReAct SR (\%) with hidden or disclosed rules by suite, pooled over models and tasks (Vanilla SR 63.9\%).
$\mathrm{RIE}_{\mathrm{SR}}$ is the Gap (Vanilla minus Disclosed) plus $\Delta\mathrm{SR}$ (Disclosed minus Hidden), in percentage points.}
\label{tab:rule-disclosure}
\centering
{\small
\setlength{\tabcolsep}{2.3pt}
\renewcommand{\arraystretch}{1.12}
\begin{tabular*}{\columnwidth}{@{\extracolsep{\fill}}lccrrr@{}}
\toprule
Suite & Hidden & Disclosed & $\mathrm{RIE}_{\mathrm{SR}}$ & Gap & $\Delta\mathrm{SR}$\\
\midrule
M01 & 37.2 & 42.8 & $+26.7$ & $+21.1$ & $+5.6$\\
M02 & 67.2 & 70.0 & $-3.3$ & $-6.1$ & $+2.8$\\
M03 & 67.2 & 73.3 & $-3.3$ & $-9.4$ & $+6.1$\\
M04 & 22.2 & 31.7 & $+41.7$ & $+32.2$ & $+9.4$\\
M05 & 58.9 & 64.4 & $+5.0$ & $-0.6$ & $+5.6$\\
M06 & 65.0 & 68.9 & $-1.1$ & $-5.0$ & $+3.9$\\
\midrule
Pooled & 53.0 & 58.5 & $+10.9$ & $+5.4$ & $+5.6$\\
\bottomrule
\end{tabular*}}
\end{table}

For ReAct, the cost of hidden rules is modest and does not differ detectably between suites, whereas the gap that remains once the rules are known varies widely (Table~\ref{tab:rule-disclosure}).
Not being told the rules costs $5.6\pm2.8$ points of SR pooled over suites (95\% interval; 6.9 with Gemini and 4.3 with DeepSeek), and the six single-suite costs do not differ detectably (Cochran's $Q$, $p=0.84$).
The gap with exact rules instead ranges from $-9.4$ to $+32.2$ points and differs strongly between suites ($p<0.001$), so this gap, not the cost of hiding the rules, accounts for most of the variation in RIE between suites.
Under route substitution, disclosure closes ReAct's observed performance gap, so the RIE of M05 is approximately accounted for by the cost of hidden rules.
Under quantity scarcity, the gap is 21.1 points with each model; here low yields multiply the gathering and processing needed within the same budget.
Under compressed drops, a 32.2-point gap remains even though unpacking a storage block takes one crafting call and yields nine items.
With DeepSeek on Iron Armor, for example, ReAct succeeds in 93.3\% of Vanilla episodes, 13.3\% of M04 episodes, and 43.3\% when told the rules, whereas ADAM succeeds in 63.3\% of M04 episodes without being told.
The least-affected configurations retain 78\% to 85\% of their Vanilla SR through the same interface (Section~\ref{sec:suite-results}).
This is consistent with difficulty applying the disclosed rule, rather than an absence of an executable route.
Under the three suites that expand yields or add byproducts, disclosed ReAct exceeds its Vanilla SR by 5.0 to 9.4 points, and hidden rules forfeit 2.8 to 6.1 points of this gain.
In every condition, the interface lists the Vanilla milestones and the next unfinished one.
These lie on the changed routes of all suites except M02, whose book recipe makes diamonds unnecessary for Enchantment.
There, ReAct w/rules with DeepSeek reaches a mean Score of 80.2 but completes only 3.3\% of episodes, a pattern consistent with the milestone list steering an informed agent toward the Vanilla route.

The behavioral diagnostics, reported per suite in \suppref{the supplementary material}{Appendix~\ref{app:behavior-details}}, are consistent with this split.
Within M01 and M04, the suites with the longest RL under hidden rules, disclosure shortens ReAct's RL from 15.3 to 10.5 and from 9.1 to 6.0 actions.
M02 yields the highest IRR of any suite while keeping RL short, consistent with byproducts that reward repeating an action, whereas IRR stays below 5\% under M05, where repeating a changed craft only yields another hoe or blast furnace.
Pooled over suites, ReAct w/rules has a higher IRR than ReAct (9.9\% versus 7.1\%) but a shorter RL (5.4 versus 7.2 actions), so immediate repetition alone is not an error signal.

\begin{table}[t]
\caption{Cumulative Gemini SR (\%) at three checkpoints of the same trajectories in the horizon study, averaged over tasks.
Mirror worlds are copies of 10 of the 25 Vanilla worlds, so values are compared within each condition.}
\label{tab:action-checkpoints-main}
\centering
{\small
\setlength{\tabcolsep}{1.5pt}
\renewcommand{\arraystretch}{1.12}
\begin{tabular*}{\columnwidth}{@{\extracolsep{\fill}}lcccccc@{}}
\toprule
& \multicolumn{2}{c}{100 actions} & \multicolumn{2}{c}{150 actions} & \multicolumn{2}{c}{200 actions}\\
\cmidrule(lr){2-3}\cmidrule(lr){4-5}\cmidrule(lr){6-7}
Configuration & Vanilla & Mirror & Vanilla & Mirror & Vanilla & Mirror\\
\midrule
ReAct & 77.0 & 64.4 & 85.3 & 72.1 & 94.7 & 73.9\\
Reflexion & 72.7 & 68.6 & 83.7 & 74.4 & 93.7 & 77.1\\
XENON & 39.3 & 37.4 & 63.3 & 76.9 & 85.3 & 86.8\\
ADAM & 49.0 & 28.5 & 77.7 & 48.2 & 82.3 & 71.5\\
\bottomrule
\end{tabular*}}
\end{table}
\section{Discussion}
\label{sec:discussion}

RIE alone can mislead in two ways.
First, $\mathrm{RIE}_{\mathrm{SR}}$ is bounded by Vanilla SR: XENON with Gemini shows almost no change because it rarely succeeds in either world.
Second, pooling suites with effects of opposite signs can hide large losses under individual interventions.
Evaluations under changed rules should therefore report Mirror performance, RIE for each suite, and the action budget, and compare agents within a suite.
Reporting SR at several checkpoints of one trajectory adds no episodes and reveals whether a ranking reflects speed or eventual success, as the reversal between 100 and 200 actions in Table~\ref{tab:action-checkpoints-main} illustrates.

Telling ReAct the rules also does not ensure that it acts on them.
Disclosure removes ReAct's drop under route substitution but leaves most of it under compressed drops, although the disclosed route adds at most one crafting call per resource.
Mars likewise finds that knowing the modified rules does not suffice~\cite{mars}, and Code over Words and studies of knowledge conflicts find that models can fall back on familiar knowledge that stated rules or contexts contradict~\cite{xu2026codewords,longpre2021entity,xie2024adaptive}.
An evaluation that tests only whether an agent detects a change would miss such failures, which the Vanilla twin of each Mirror world makes measurable.

\section{Conclusion}
\mirror{} evaluates hidden rule changes against unchanged copies of the same Minecraft worlds, using a common interface and action budget.
Scarcity and compressed drops sharply reduce success, while route-shortening suites provide only small gains under hidden rules.
ReAct, Reflexion, and Voyager lead in absolute Mirror performance under the main budgets, but not in pooled RIE.
Among the compared configurations, significant within-suite differences appear only under compressed drops, where ReAct loses most.
For ReAct, disclosure yields modest gains, and most between-suite variation remains with exact rules.
The paired design thus distinguishes the effect removed by disclosure from the gap that remains.

Each harmful form of change is represented by one suite, disclosure is evaluated for ReAct as in Mars, our adaptations of Voyager, XENON, and ADAM learn within an episode, and the main study uses two cost-efficient models without provider-side reasoning.
Because suites are compiled from manifests for the same worlds, further suites for each form, disclosure for every configuration, an interface without the milestone list, and agents that learn across episodes can be evaluated under the same controls.
Since disclosure is set per agent, the same worlds could also host teams whose members know different versions of the rules, extending the benchmark to coordination under mismatched rule knowledge.
We will release the datapacks, code, and prompts.

\clearpage

\clearpage
\renewcommand{\floatpagefraction}{0.5}
\renewcommand{\dblfloatpagefraction}{0.5}
\setlength{\emergencystretch}{3em}
\ifarxiv\section*{Appendix Overview}\else\section*{Overview}\fi
\label{sec:supp-overview}
\ifarxiv\suppressfloats[t]\fi

\mainref{This document complements the main paper}{The appendices complement the main text} with disaggregated results and implementation details.
Section~\ref{app:luna} reports the full GPT-5.6 Luna evaluation, Section~\ref{app:horizon} reports the horizon study with a 200-action cap, and Section~\ref{app:aggregate-results} breaks the main results down by task, model, and biome, splits ReAct's RIE by rule suite and model, compares configurations within each rule suite, and reports 95\% intervals for the main estimates.
Sections~\ref{app:baselines} and~\ref{app:direct} specify the agent configurations and the input of \direct{}.
Section~\ref{app:datapacks} documents the automated construction of world-generation and rule-intervention datapacks, including the prompts of the LLM designer and judge.
Sections~\ref{app:checkpoints} and~\ref{app:behavior-details} give the task milestones and the behavioral diagnostics for each rule suite, Section~\ref{app:biome-details} describes the five biome setups, and Section~\ref{app:mirror-details} documents how each Mirror world is built and reports detailed results for every rule suite.
Section~\ref{sec:limits} discusses the limitations of the study.
Section~\ref{app:agent-prompts} reproduces the evaluation-agent prompt templates, their shared components, and the rule sheets for M01--M06.

Unless stated otherwise, Score, success rate (SR), and the Rule Intervention Effect (RIE) follow the definitions in \mainref{the main paper}{Section~\ref{sec:bench}}.
Vanilla--Mirror comparisons are paired by Vanilla world, model, task, agent configuration, and action budget.
Reported values follow the aggregation level stated for each analysis and are rounded to the displayed precision after aggregation.

\appendix

\section{Additional Model Evaluation}
\label{app:luna}

We additionally evaluate GPT-5.6 Luna on the Diamond task.
For each of the six configurations evaluated in both settings, Vanilla covers all five controlled biomes with two maps per biome and two repetitions per map.
Mirror covers Plains and Savanna with two source maps per biome, all six rule suites, and two repetitions; ReAct w/rules is evaluated only in Mirror.
Table~\ref{tab:luna} reports averages over the corresponding evaluation sets.
Because the Vanilla and Mirror aggregates cover different biome sets, we compare configurations within each condition and do not report RIE.

\begin{table}[t]
\caption{GPT-5.6 Luna on Diamond. The upper panel reports Vanilla and Mirror Score\,/\,SR; the lower panel reports Vanilla\,/\,Mirror Actions and Cost. SR is in percent and Cost in US dollars.}
\label{tab:luna}
\centering
{\footnotesize
\setlength{\tabcolsep}{2.0pt}
\renewcommand{\arraystretch}{1.05}
\begin{tabular*}{\columnwidth}{@{\extracolsep{\fill}}lcc@{}}
\toprule
Configuration & Vanilla Score\,/\,SR & Mirror Score\,/\,SR\\
\midrule
\direct{} & 64.8\,/\,45.0 & 60.4\,/\,39.6\\
ReAct & 80.0\,/\,65.0 & 82.8\,/\,68.8\\
Reflexion & 88.8\,/\,80.0 & 83.3\,/\,66.7\\
Voyager & 90.5\,/\,85.0 & 72.9\,/\,54.2\\
XENON & 90.0\,/\,80.0 & 73.0\,/\,47.9\\
ADAM & 90.0\,/\,80.0 & 77.6\,/\,62.5\\
ReAct w/rules & -- & 89.1\,/\,81.2\\
\bottomrule
\end{tabular*}
\vspace{1mm}

\begin{tabular*}{\columnwidth}{@{\extracolsep{\fill}}lcc@{}}
\toprule
Configuration & Actions (V/M) & Cost (V/M)\\
\midrule
\direct{} & 43.0\,/\,42.2 & 0.16\,/\,0.17\\
ReAct & 39.5\,/\,38.4 & 0.27\,/\,0.26\\
Reflexion & 34.1\,/\,33.6 & 0.22\,/\,0.23\\
Voyager & 35.9\,/\,37.6 & 0.21\,/\,0.23\\
XENON & 39.5\,/\,38.9 & 0.33\,/\,0.37\\
ADAM & 33.2\,/\,41.3 & 0.35\,/\,0.50\\
ReAct w/rules & --\,/\,34.0 & --\,/\,0.23\\
\bottomrule
\end{tabular*}}
\end{table}

Within the evaluated Mirror subset, Reflexion obtains the highest Score among configurations without rule descriptions (83.3), while ReAct obtains the highest SR (68.8\%).
Providing the exact rules increases ReAct's Score from 82.8 to 89.1 and its SR from 68.8\% to 81.2\%, while reducing its mean number of semantic actions from 38.4 to 34.0.
XENON and ADAM reach 80.0\% Vanilla SR with this model, compared with 13.3\% and 36.7\% on Diamond with Gemini.
We report these results separately from the two-model main study because they cover only the Diamond task and a different subset of worlds.
With 48 Mirror episodes per condition, the disclosure gain of 12.5 points has a 95\% interval of $\pm17.1$ points, so this check does not establish a disclosure effect on its own.

\section{Horizon Study}
\label{app:horizon}

To test whether the interaction horizon constrains completion, we run ReAct, Reflexion, XENON, and ADAM with Gemini under a 200-action cap and record cumulative SR at actions 100, 150, and 200 (Table~\ref{tab:action-checkpoints}).
The checkpoints share trajectories and therefore measure when episodes complete rather than results under separate budgets.
This horizon study uses its own worlds from the same five biomes and four runs per world and task: 25 Vanilla worlds (five per biome) and 60 Mirror worlds copied from 10 of them (two per biome, six suites each).
Because the two conditions cover different sets of worlds, we compare configurations within each condition.

\begin{table}[!tb]
\caption{Cumulative Gemini SR (\%) in the horizon study, averaged over tasks. Mirror worlds are copies of 10 of the 25 Vanilla worlds, so values are compared within each condition.}
\label{tab:action-checkpoints}
\centering
{\small
\setlength{\tabcolsep}{1.5pt}
\renewcommand{\arraystretch}{1.12}
\begin{tabular*}{\columnwidth}{@{\extracolsep{\fill}}lcccccc@{}}
\toprule
& \multicolumn{2}{c}{100 actions} & \multicolumn{2}{c}{150 actions} & \multicolumn{2}{c}{200 actions}\\
\cmidrule(lr){2-3}\cmidrule(lr){4-5}\cmidrule(lr){6-7}
Configuration & Vanilla & Mirror & Vanilla & Mirror & Vanilla & Mirror\\
\midrule
ReAct & 77.0 & 64.4 & 85.3 & 72.1 & 94.7 & 73.9\\
Reflexion & 72.7 & 68.6 & 83.7 & 74.4 & 93.7 & 77.1\\
XENON & 39.3 & 37.4 & 63.3 & 76.9 & 85.3 & 86.8\\
ADAM & 49.0 & 28.5 & 77.7 & 48.2 & 82.3 & 71.5\\
\bottomrule
\end{tabular*}}
\end{table}

At action 100, XENON and ADAM trail ReAct and Reflexion in both settings, but from action 100 to 200 their Mirror SRs rise by 49.4 and 43.0 percentage points, compared with 9.5 and 8.5 points for ReAct and Reflexion.
At action 200, XENON leads in Mirror SR (86.8\%) while ADAM still trails ReAct and Reflexion (71.5\%), and ReAct leads in Vanilla SR (94.7\%).
XENON's and ADAM's Vanilla SRs also rise sharply from action 100 to 200 (by 46.0 and 33.3 points), consistent with the actions that their attempts and probes consume.

\section{Additional Aggregate Results}
\label{app:aggregate-results}

Table~\ref{tab:main-detailed} disaggregates the main results by task and model.
Vanilla and Mirror entries aggregate the corresponding evaluations, with Mirror results pooled equally across the six rule suites.
The six configurations shared across settings receive no rule descriptions in Mirror; ReAct w/rules receives the exact descriptions.
For RIE, each per-world Mirror mean is subtracted from its matched Vanilla-world mean for the same model, task, and agent configuration before pooling.
Actions and cost are averaged over the corresponding Mirror evaluations and do not enter RIE.

\begin{table*}[p]
\caption{Vanilla and paired Mirror performance, RIE, and evaluation cost by task, model, and agent configuration.
Vanilla and Mirror report Score\,/\,SR with SR in percent; $\mathrm{RIE}_{\mathrm{SR}}$ is in percentage points and Cost in US dollars.}
\label{tab:main-detailed}
\centering
{\footnotesize
\setlength{\tabcolsep}{3.0pt}
\renewcommand{\arraystretch}{1.03}
\begin{tabular*}{\textwidth}{@{\extracolsep{\fill}}llcccccc@{}}
\toprule
\multicolumn{8}{c}{Gemini}\\
\cmidrule(lr){1-8}
Task & Configuration & Vanilla & Mirror & $\mathrm{RIE}_{\mathrm{SC}}$ & $\mathrm{RIE}_{\mathrm{SR}}$ & Actions & Cost\\
\midrule
\multirow{7}{*}{Iron Armor}
& \direct{} & 73.5\,/\,30.0 & 69.3\,/\,23.3 & $+4.2$ & $+6.7$ & 45.7 & 0.04\\
& ReAct & 95.3\,/\,86.7 & 90.8\,/\,75.6 & $+4.6$ & $+11.1$ & 38.9 & 0.07\\
& ReAct w/rules & -- & 93.6\,/\,81.7 & -- & -- & 37.4 & 0.07\\
& Reflexion & 95.3\,/\,86.7 & 95.4\,/\,87.8 & $-0.1$ & $-1.1$ & 36.3 & 0.07\\
& Voyager & 97.7\,/\,93.3 & 95.6\,/\,88.9 & $+2.1$ & $+4.4$ & 35.5 & 0.05\\
& XENON & 71.2\,/\,23.3 & 65.8\,/\,17.8 & $+5.4$ & $+5.6$ & 46.1 & 0.08\\
& ADAM & 74.7\,/\,33.3 & 69.5\,/\,27.2 & $+5.1$ & $+6.1$ & 46.6 & 0.12\\
\midrule
\multirow{7}{*}{Diamond}
& \direct{} & 65.8\,/\,10.0 & 66.2\,/\,13.9 & $-0.3$ & $-3.9$ & 47.7 & 0.04\\
& ReAct & 98.0\,/\,93.3 & 91.4\,/\,73.3 & $+6.6$ & $+20.0$ & 40.3 & 0.07\\
& ReAct w/rules & -- & 94.8\,/\,82.8 & -- & -- & 40.2 & 0.07\\
& Reflexion & 96.0\,/\,86.7 & 92.8\,/\,77.8 & $+3.2$ & $+8.9$ & 41.1 & 0.07\\
& Voyager & 96.0\,/\,86.7 & 92.2\,/\,75.6 & $+3.8$ & $+11.1$ & 40.6 & 0.05\\
& XENON & 67.7\,/\,13.3 & 68.0\,/\,16.7 & $-0.3$ & $-3.3$ & 47.2 & 0.08\\
& ADAM & 77.5\,/\,36.7 & 73.8\,/\,26.7 & $+3.8$ & $+10.0$ & 47.5 & 0.13\\
\midrule
\multirow{7}{*}{Enchantment}
& \direct{} & 72.5\,/\,0.0 & 59.8\,/\,0.0 & $+12.7$ & $0.0$ & 75.0 & 0.08\\
& ReAct & 90.0\,/\,20.0 & 87.0\,/\,15.6 & $+3.0$ & $+4.4$ & 73.3 & 0.15\\
& ReAct w/rules & -- & 87.5\,/\,20.6 & -- & -- & 72.6 & 0.15\\
& Reflexion & 87.2\,/\,16.7 & 81.8\,/\,6.7 & $+5.4$ & $+10.0$ & 74.3 & 0.16\\
& Voyager & 82.7\,/\,6.7 & 77.2\,/\,3.9 & $+5.5$ & $+2.8$ & 73.9 & 0.12\\
& XENON & 82.0\,/\,0.0 & 75.2\,/\,2.8 & $+6.8$ & $-2.8$ & 74.7 & 0.20\\
& ADAM & 73.3\,/\,3.3 & 70.3\,/\,1.1 & $+3.1$ & $+2.2$ & 74.9 & 0.26\\
\midrule
\multicolumn{8}{c}{DeepSeek}\\
\cmidrule(lr){1-8}
Task & Configuration & Vanilla & Mirror & $\mathrm{RIE}_{\mathrm{SC}}$ & $\mathrm{RIE}_{\mathrm{SR}}$ & Actions & Cost\\
\midrule
\multirow{7}{*}{Iron Armor}
& \direct{} & 71.7\,/\,50.0 & 64.4\,/\,21.1 & $+7.3$ & $+28.9$ & 44.8 & 0.01\\
& ReAct & 96.7\,/\,93.3 & 87.8\,/\,77.8 & $+8.9$ & $+15.6$ & 38.7 & 0.03\\
& ReAct w/rules & -- & 92.0\,/\,84.4 & -- & -- & 38.0 & 0.03\\
& Reflexion & 94.5\,/\,90.0 & 88.7\,/\,76.7 & $+5.8$ & $+13.3$ & 39.4 & 0.03\\
& Voyager & 95.5\,/\,93.3 & 91.6\,/\,81.7 & $+3.9$ & $+11.7$ & 39.5 & 0.02\\
& XENON & 86.7\,/\,83.3 & 88.2\,/\,73.9 & $-1.6$ & $+9.4$ & 38.8 & 0.04\\
& ADAM & 89.2\,/\,83.3 & 90.2\,/\,76.7 & $-1.1$ & $+6.7$ & 39.8 & 0.04\\
\midrule
\multirow{7}{*}{Diamond}
& \direct{} & 41.8\,/\,26.7 & 50.5\,/\,6.7 & $-8.7$ & $+20.0$ & 48.9 & 0.01\\
& ReAct & 90.8\,/\,83.3 & 86.5\,/\,72.2 & $+4.3$ & $+11.1$ & 41.9 & 0.03\\
& ReAct w/rules & -- & 88.9\,/\,75.0 & -- & -- & 40.8 & 0.03\\
& Reflexion & 86.3\,/\,76.7 & 84.4\,/\,62.2 & $+1.9$ & $+14.4$ & 42.3 & 0.03\\
& Voyager & 79.7\,/\,63.3 & 79.8\,/\,46.1 & $-0.1$ & $+17.2$ & 45.0 & 0.02\\
& XENON & 80.3\,/\,66.7 & 83.5\,/\,52.8 & $-3.1$ & $+13.9$ & 43.6 & 0.04\\
& ADAM & 83.2\,/\,70.0 & 84.0\,/\,57.2 & $-0.8$ & $+12.8$ & 44.9 & 0.04\\
\midrule
\multirow{7}{*}{Enchantment}
& \direct{} & 40.5\,/\,0.0 & 32.3\,/\,0.0 & $+8.3$ & $0.0$ & 74.3 & 0.02\\
& ReAct & 83.7\,/\,6.7 & 77.3\,/\,3.3 & $+6.4$ & $+3.3$ & 74.0 & 0.05\\
& ReAct w/rules & -- & 80.8\,/\,6.7 & -- & -- & 73.1 & 0.05\\
& Reflexion & 68.2\,/\,0.0 & 62.9\,/\,0.0 & $+5.2$ & $0.0$ & 74.4 & 0.06\\
& Voyager & 59.8\,/\,0.0 & 57.5\,/\,0.0 & $+2.3$ & $0.0$ & 74.3 & 0.04\\
& XENON & 84.7\,/\,6.7 & 77.8\,/\,2.8 & $+6.9$ & $+3.9$ & 74.0 & 0.08\\
& ADAM & 74.0\,/\,0.0 & 70.2\,/\,0.6 & $+3.8$ & $-0.6$ & 74.1 & 0.08\\
\bottomrule
\end{tabular*}}
\end{table*}

\paragraph{Task and model breakdown.}
For Enchantment, several configurations reach late milestones while rarely completing the final milestone, whereas Iron Armor and Diamond show closer agreement between progress and completion (Table~\ref{tab:main-detailed}).
For some combinations of task, model, and configuration, $\mathrm{RIE}_{\mathrm{SC}}$ is negative, meaning that the mean Mirror Score exceeds the Vanilla mean for the same worlds.
These cases arise when the gains in the beneficial suites outweigh the losses in the detrimental ones.

Table~\ref{tab:biome} pools models, tasks, and the six configurations evaluated without rule descriptions in both settings.

\begin{table}[t]
\caption{Vanilla and paired Mirror Score, SR, and RIE by biome. SR is in percent and $\mathrm{RIE}_{\mathrm{SR}}$ in percentage points.}
\label{tab:biome}
\centering
{\footnotesize
\setlength{\tabcolsep}{3.2pt}
\renewcommand{\arraystretch}{1.05}
\begin{tabular*}{\columnwidth}{@{\extracolsep{\fill}}lcccccc@{}}
\toprule
\multirow{2}{*}{Biome} & \multicolumn{2}{c}{Vanilla} & \multicolumn{2}{c}{Mirror} & \multirow{2}{*}{$\mathrm{RIE}_{\mathrm{SC}}$} & \multirow{2}{*}{$\mathrm{RIE}_{\mathrm{SR}}$}\\
\cmidrule(lr){2-3}\cmidrule(lr){4-5}
& Score & SR & Score & SR & & \\
\midrule
Plains & 81.5 & 46.3 & 78.9 & 40.4 & $+2.6$ & $+5.9$\\
Taiga & 84.9 & 50.0 & 82.0 & 42.1 & $+2.9$ & $+7.9$\\
Snowy Plains & 84.5 & 50.0 & 81.0 & 42.6 & $+3.5$ & $+7.4$\\
Jungle & 75.3 & 36.1 & 73.8 & 31.7 & $+1.4$ & $+4.4$\\
Savanna & 77.1 & 42.6 & 69.8 & 29.6 & $+7.3$ & $+13.0$\\
\bottomrule
\end{tabular*}}
\end{table}

\paragraph{Biome breakdown.}
Across the tested worlds in each biome, both RIE estimates are positive but vary in magnitude (Table~\ref{tab:biome}).
Savanna has the lowest Mirror Score and SR and the largest decreases from Vanilla to Mirror, whereas Taiga has the highest Mirror Score.
Because each biome contributes two Vanilla worlds, these results characterize the tested worlds of each biome.

\begin{table}[t]
\caption{Split of ReAct's $\mathrm{RIE}_{\mathrm{SR}}$ by rule suite and model, pooled over tasks. Gap is Vanilla minus Disclosed, and $\Delta\mathrm{SR}$ is Disclosed minus Hidden; all values are in percentage points.}
\label{tab:split-by-model}
\centering
{\footnotesize
\setlength{\tabcolsep}{2.5pt}
\renewcommand{\arraystretch}{1.05}
\begin{tabular*}{\columnwidth}{@{\extracolsep{\fill}}lrrrrrr@{}}
\toprule
& \multicolumn{3}{c}{Gemini} & \multicolumn{3}{c}{DeepSeek}\\
\cmidrule(lr){2-4}\cmidrule(lr){5-7}
Suite & $\mathrm{RIE}_{\mathrm{SR}}$ & Gap & $\Delta\mathrm{SR}$ & $\mathrm{RIE}_{\mathrm{SR}}$ & Gap & $\Delta\mathrm{SR}$\\
\midrule
M01 & $+33.3$ & $+21.1$ & $+12.2$ & $+20.0$ & $+21.1$ & $-1.1$\\
M02 & $-8.9$ & $-6.7$ & $-2.2$ & $+2.2$ & $-5.6$ & $+7.8$\\
M03 & $-2.2$ & $-6.7$ & $+4.4$ & $-4.4$ & $-12.2$ & $+7.8$\\
M04 & $+35.6$ & $+27.8$ & $+7.8$ & $+47.8$ & $+36.7$ & $+11.1$\\
M05 & $+10.0$ & $+1.1$ & $+8.9$ & $0.0$ & $-2.2$ & $+2.2$\\
M06 & $+3.3$ & $-6.7$ & $+10.0$ & $-5.6$ & $-3.3$ & $-2.2$\\
\midrule
Pooled & $+11.9$ & $+5.0$ & $+6.9$ & $+10.0$ & $+5.7$ & $+4.3$\\
\bottomrule
\end{tabular*}}
\end{table}

\paragraph{Split of ReAct's RIE by model.}
Table~\ref{tab:split-by-model} repeats the split of ReAct's $\mathrm{RIE}_{\mathrm{SR}}$ \mainref{from the main paper}{from Table~\ref{tab:rule-disclosure}} separately for each model.
The gap with exact rules is large under M01 and M04 and near zero under M05 with both models, whereas the cost of hidden rules for a single suite varies more between the models and is slightly negative in three of the twelve cells.
Pooled over suites, this cost is 6.9 points with Gemini and 4.3 points with DeepSeek.

\paragraph{Within-suite comparison of configurations.}
\mainref{Figure~3 of the main paper}{Figure~\ref{fig:suite-config}} reports $\mathrm{RIE}_{\mathrm{SR}}$ by suite on Iron Armor and Diamond for the compared configurations; all values follow from Table~\ref{tab:main-detailed} and Tables~\ref{tab:details-mirror01}--\ref{tab:details-mirror06}.
The compared set, the configurations whose Vanilla SR is at least 50\% on each of these two tasks, is the same for any threshold from 40\% to 60\%, whether the threshold is applied to each task or to the mean of the two tasks.
For each suite and model, Cochran's $Q$ weights each configuration's RIE by the inverse of its binomial variance and tests whether the RIEs differ.
Under M04, $Q=41.7$ with 4 degrees of freedom ($p<0.001$) with DeepSeek and $Q=11.4$ with 2 degrees of freedom ($p=0.003$) with Gemini, and both tests remain significant after Bonferroni correction for the 12 tests (6 suites and 2 models); under M01, $p=0.16$ with Gemini and $p=0.31$ with DeepSeek, and in the other four suites $p>0.29$.
Replacing the binomial variances with continuity-corrected (Agresti--Coull) variances leaves every conclusion unchanged: under M04, $p=0.003$ with Gemini and $p<0.001$ with DeepSeek, and $p\geq0.15$ for every other suite and model.
Under M04, XENON and ADAM keep 82\% and 78\% of their Vanilla SR with DeepSeek, Reflexion and Voyager keep 85\% with Gemini, ReAct keeps 23\% and 52\%, and ReAct w/rules keeps 42\% and 63\%.
In each model, the mean SR of the two configurations that lose least exceeds that of ReAct w/rules by $18.3\pm14.7$ points with Gemini and $24.2\pm14.8$ with DeepSeek; the four individual differences are $16.7\pm16.6$, $20.0\pm16.5$, $25.0\pm16.8$, and $23.3\pm17.3$ points.
Under M02, M03, and M06, the compared configurations stay between $-12$ and $+8$ points.
Like the intervals, these tests treat episodes within a world as independent.

\paragraph{Uncertainty of the main estimates.}
Table~\ref{tab:ci} gives 95\% intervals for the RIE of each suite (\mainref{Table~3 of the main paper}{Table~\ref{tab:rule-rie}}), ReAct's disclosure split (\mainref{Table~4 of the main paper}{Table~\ref{tab:rule-disclosure}}), and the RIE of each configuration pooled over models (\mainref{Table~2 of the main paper}{Table~\ref{tab:main}}).
The intervals are binomial and stratified by model and task, with 30 episodes per cell.
They treat episodes within a world as independent.
The pooled cost of not being told the rules, 5.6 points, excludes zero, whereas the six single-suite costs do not differ detectably (Cochran's $Q=2.1$ with 5 degrees of freedom, $p=0.84$; $p=0.87$ with continuity-corrected variances).
The gap with exact rules, in contrast, differs strongly between suites ($Q=196$ with 5 degrees of freedom, $p<0.001$); because all suites share ReAct's Vanilla SR, this test compares the disclosed SRs.
The pooled RIEs of the six configurations also overlap, so pooled $\mathrm{RIE}_{\mathrm{SR}}$ alone does not reliably rank the configurations.

\begin{table}[t]
\caption{Estimates with 95\% intervals (binomial, stratified by model and task), in percentage points.
Suite rows pool the six configurations evaluated without rule descriptions; the ReAct columns split its $\mathrm{RIE}_{\mathrm{SR}}$ into Gap (Vanilla minus Disclosed) and $\Delta\mathrm{SR}$ (Disclosed minus Hidden).}
\label{tab:ci}
\centering
{\footnotesize
\setlength{\tabcolsep}{2.2pt}
\renewcommand{\arraystretch}{1.08}
\begin{tabular*}{\columnwidth}{@{\extracolsep{\fill}}lrrrr@{}}
\toprule
& All six & \multicolumn{3}{c}{ReAct}\\
\cmidrule(lr){2-2}\cmidrule(lr){3-5}
Suite & $\mathrm{RIE}_{\mathrm{SR}}$ & $\mathrm{RIE}_{\mathrm{SR}}$ & Gap & $\Delta\mathrm{SR}$\\
\midrule
M01 & $+23.5\pm2.8$ & $+26.7\pm7.6$ & $+21.1\pm7.5$ & $+5.6\pm8.4$\\
M02 & $-0.2\pm2.8$ & $-3.3\pm6.3$ & $-6.1\pm5.9$ & $+2.8\pm5.6$\\
M03 & $-3.0\pm2.7$ & $-3.3\pm5.8$ & $-9.4\pm6.5$ & $+6.1\pm5.6$\\
M04 & $+19.0\pm2.9$ & $+41.7\pm7.1$ & $+32.2\pm7.5$ & $+9.4\pm8.0$\\
M05 & $+6.9\pm2.8$ & $+5.0\pm6.8$ & $-0.6\pm6.4$ & $+5.6\pm6.6$\\
M06 & $-0.6\pm2.7$ & $-1.1\pm6.7$ & $-5.0\pm6.3$ & $+3.9\pm6.5$\\
\midrule
Pooled & $+7.6\pm2.2$ & $+10.9\pm5.1$ & $+5.4\pm5.1$ & $+5.6\pm2.8$\\
\bottomrule
\end{tabular*}\par
\vspace{2mm}
\begin{tabular*}{\columnwidth}{@{\extracolsep{\fill}}lrlr@{}}
\toprule
\multicolumn{4}{@{}l}{Pooled $\mathrm{RIE}_{\mathrm{SR}}$ by configuration (both models)}\\
\midrule
\direct{} & $+8.6\pm5.4$ & Voyager & $+7.9\pm4.8$\\
ReAct & $+10.9\pm5.0$ & XENON & $+4.4\pm5.4$\\
Reflexion & $+7.6\pm5.1$ & ADAM & $+6.2\pm5.8$\\
\bottomrule
\end{tabular*}}
\end{table}

\section{Agent Configuration Details}
\label{app:baselines}

All configurations use the same serialized observations, semantic skills, and action budgets.
They also follow the same loop for observing, updating, deciding, and executing.
Beyond this common interface, they implement distinct reasoning, memory, review, retrieval, and probing mechanisms within each episode.
Each episode starts with empty internal state, and all traces, lessons, procedures, and graphs are discarded when it ends.
The main study uses \texttt{gemini-3.1-flash-lite} and \texttt{deepseek-v4-flash}.
All calls use temperature 0.7 and an output limit of 2,048 tokens.
All evaluated models, including GPT-5.6 Luna, were run with any provider-side extended-thinking or reasoning mode disabled.
This setting is distinct from the explicitly implemented reasoning traces and auxiliary agent procedures described below.

\paragraph{World and execution controls.}
\label{app:controls}
A common construction policy is instantiated as one world-generation datapack for each controlled biome, with a shared placement schedule for task resources.
Daytime is fixed, and Peaceful difficulty removes variation from hostile mobs and night.
Every agent receives the same Haste, Speed, Resistance, and Strength effects and begins at experience level 100.
These controls shorten mining, traversal, combat, and enchanting without choosing an agent subgoal or revealing a changed rule.
Iron Armor and Diamond use a budget of 50 semantic actions, while Enchantment uses 75.

\paragraph{Direct LLM}
\direct{} uses the current observation and a bounded record of verified action effects, without an auxiliary memory module.
Each model call selects and executes one semantic skill.

\paragraph{ReAct.}
ReAct maintains a trace of thoughts, actions, and observations for the 16 most recent transitions and uses this trace to select the next semantic skill.

\paragraph{ReAct w/rules.}
This configuration uses the same trace of 16 transitions and the same procedure for selecting actions as ReAct, but its prompt states that the rules have changed and includes the exact active rule descriptions in every model call.
All other configurations receive the same prompt in Vanilla and Mirror episodes, and it never states that the rules may differ.

\paragraph{Reflexion.}
Reflexion augments the ReAct trace with up to three temporary lessons.
It creates or revises lessons after repeated failure, conflicting evidence, stalled progress, or a scheduled review, with at most six reviews per episode and a retention window of 16 actions per lesson.

\paragraph{Voyager.}
Voyager maintains a procedure library of at most 16 entries within each episode.
It generates programs of four skills, critiques their execution, retrieves up to five procedures, and permits at most four repairs.

\paragraph{XENON}
XENON maintains a graph of task dependencies and a memory of observed effects, each with capacity 32.
It retrieves up to three entries, records action outcomes, corrects contradicted edges, and replans from the updated frontier after two failures.

\paragraph{ADAM}
ADAM maintains a causal graph of at most 48 nodes and a transition memory of at most 32 entries.
It learns from two comparable samples and may execute at most two probes when an uncertain dependency blocks progress; probing is disabled during the final 12 actions.

Neither XENON nor ADAM receives evaluator feedback that identifies an incorrect dependency.
An ADAM probe is an ordinary semantic action and consumes the same action budget as any other skill call.
Additional model calls used for reflection, critique, retrieval, or graph analysis do not act in Minecraft.

Deaths, policy failures, and exhausted budgets remain in the denominator.
Disconnected clients, missing datapacks, and unavailable advancements are rerun under the same episode identifier.
Before aggregation, we check the recorded biome, identifier of the Vanilla world, whether the episode uses Vanilla or Mirror rules, task, configuration, rule disclosure status, and action budget.
Score, SR, both RIEs, $\Delta\mathrm{Score}$, and $\Delta\mathrm{SR}$ are computed only after these checks.

\section{Direct LLM Input}
\label{app:direct}

\direct{} maps the current serialized observation to one semantic skill call from the common interface.
Its input contains inventory, position, task and milestones, nearby blocks and entities, outcomes of previous actions, and available skills.
The serialized observation also reports equipment, current biome, survival state, nearby dropped items, and the next unfinished milestone.
For \direct{}, previous outcomes include the latest action result and a bounded record of verified action effects.
The Direct LLM templates and shared observation components are reproduced in Sections~\ref{app:prompt-01} and~\ref{app:prompt-08}; the six rule sheets use the same M01--M06 identifiers as Section~\ref{app:mirror-details}.

\section{Automated Datapack Construction}
\label{app:datapacks}

Minecraft Java datapacks express recipes, loot tables, advancements, and resources for world generation as JSON files loaded by the server \cite{minecraftwiki_datapack,minecraftwiki_recipe,minecraftwiki_loot_table,minecraftwiki_worldgen}.
MirrorCraft instantiates two datapack families through one parameterized offline construction system: world-generation specifications define the controlled Vanilla environments, while rule-intervention manifests define the gameplay changes applied to their paired Mirror copies.
Both families share a versioned JSON protocol, compiler interface, feedback loop, and termination policy.
The domain flag \texttt{kind}\,$\in\{\texttt{biome},\texttt{mirror}\}$ selects the appropriate specification schema, allowlist, compiler, design criteria, and audit criteria.

\subsection{Shared Construction Workflow}

The workflow begins from a complete specification supplied by the user or generated by a deterministic seed-based sampler.
When LLM-assisted design is enabled, the Designer returns a complete replacement specification conditioned on the constraints, allowlist, current static report, and preceding Judge report.
Both the Designer and the Judge use GPT-5.6 Sol (API model \texttt{gpt-5.6-sol})~\cite{gpt56sol} for world-generation and rule-intervention datapacks alike.
The domain compiler validates and provisionally materializes each candidate; static failures and Judge revisions are returned for another bounded round.
A candidate is finalized when compilation succeeds and the Judge reports no hard failure and deems every selected task solvable, while static-only runs end after the compiler audit.
Construction-level acceptance is followed by the deterministic loading, outcome, information-integrity, and route controls described in \mainref{Section~3.7 of the main paper}{Section~\ref{sec:integrity}}.
All benchmark worlds and rule suites were frozen before any agent-evaluation results were inspected; those results were not used to select or revise them.

\subsection{World-Generation Datapacks}
\label{app:worldgen-construction}

For \texttt{kind=biome}, MirrorCraft represents each controlled biome as a versioned world-generation specification.
The specification factorizes a world into terrain, resources, ecology, and runtime controls.
Terrain fields determine the vertical range, sea level, relief noise, surface materials, and water generation.
Resource entries define one or more height bands per material, with separate controls for vein size, placement attempts, vertical range, and air-exposure discard.
Ecology fields select a benchmark or Vanilla feature backbone, or a minimal feature policy, and parameterize trees, vegetation, decorations, water features, and passive creatures.
Runtime fields configure difficulty, initial time, and day and weather cycles.

Pre-compilation validation checks the schema, numeric bounds, whitelisted identifiers, ore-band parameters, ecology fields, creature spawning, and the presence of resource categories required by the selected tasks.
A deterministic whitelist compiler then instantiates the world preset, biome, noise settings, density functions, configured features, placed features, and setup function for Minecraft Java 1.19.0 (pack format 10), and packages the resulting files deterministically for installation in the corresponding evaluation save.
A post-packaging audit parses every JSON document and checks the pack format, required world-generation resource types, custom resource identifiers, internal references, feature-stage structure, ore placements, noise settings, and artifact metadata.
The Judge then assesses the static plausibility of spawn safety, traversal, biome recognizability, resource density, and task completion from the specification and audit report.
Section~\ref{app:biome-details} reports the resulting terrain and placement controls.

\subsection{Rule-Intervention Datapacks}

For \texttt{kind=mirror}, the current specification is a task-aware rule manifest, and the injected catalog contains the allowed operators, editable targets, processing inputs and stations, output items, and dependency routes for the selected tasks.
The compiler constructs an abstract dependency graph from predefined task routes and task-relevant transformations, then identifies recipe, loot, drop, and processing transitions that can be modified without changing the map or action schema.
When enabled, the Designer proposes a compact set of interpretable interventions; the compiler materializes the manifest through standard datapack mechanisms and reports statically estimated relevance, abstract route reachability, and limited shadowing-based encounterability; and the Judge audits observability, coherence, interactions among changes, and task feasibility.
In Mirror episodes with hidden rules, agents receive neither rule descriptions nor access to a server rule query and can detect a changed rule only from the outcome of an affected action.
For rule-intervention construction, \texttt{kind=mirror}; the selected task set determines the dependency routes that must remain reachable.
After construction-level acceptance, the package, specification, and reports are finalized as a reproducible artifact; benchmark admission then follows the integrity controls in \mainref{Section~3.7 of the main paper}{Section~\ref{sec:integrity}}.

\subsection{World-Generation Prompts}

The world-generation prompts use the structured response contracts described below.
The Designer receives the user constraints, allowed catalog, current specification, static compiler/reachability report, and preceding Judge report; when no preceding report exists, its verdict is \texttt{not\_run}. The Judge receives the candidate specification and static compiler report.
The world-generation catalog contains the permitted base biomes, surface blocks, ores and ore schema, creatures, tree and feature policies, surface patches, decorations, and numeric bounds.

\paragraph{Designer prompt.}
{\itshape
You are the MirrorCraft LLM Designer. Design a complete, machine-readable candidate specification for Minecraft Java 1.19.0. The static compiler writes the datapack. Output no commands, source code, file paths, Markdown fences, comments, or prose outside one JSON object. The object contains \texttt{schema\_version}, \texttt{kind}, a brief \texttt{design\_summary}, one \texttt{resolved\_feedback} list item per addressed problem, and a complete replacement \texttt{spec}; the outer object and nested specification both use schema version \texttt{1.0} and kind \texttt{biome}. Preserve the requested \texttt{selected\_tasks}, keep every numeric field inside the supplied bounds, treat the static report as authoritative, and correct every hard failure.

Construct an Overworld single-biome evaluation world rather than a decorative showcase, with a traversable spawn surface and practically reachable resources for every selected task. The benchmark routes may require wood, stone, fuel, iron, diamond, lapis, obsidian, sugar cane, cows or leather, and experience; a listed resource is insufficient when its height range lies outside the generated terrain or its density is impractically low. Preserve the recognizable ecology of the base biome while varying terrain, surface blocks, resource distributions, vegetation, water, and passive mobs within the whitelist. Exclude Nether- or End-only assumptions, structures, chests, commands, and arbitrary identifiers. Avoid underwater spawn traps, impassable clutter, and settings that trivialize every task. Return one corrected full specification from the supplied constraints, catalog, current specification, static report, and preceding Judge report.
\par}

\paragraph{Judge prompt.}
{\itshape
You are the independent MirrorCraft LLM Judge. Audit the compiled Minecraft Java 1.19.0 candidate without redesigning the datapack or inspecting agent outcomes. Be conservative, use the static report as hard evidence, and output no Markdown or prose outside one JSON object. The object uses schema version \texttt{1.0} and kind \texttt{biome}, and contains \texttt{verdict}, \texttt{confidence}, \texttt{summary}, \texttt{hard\_failures}, \texttt{warnings}, \texttt{task\_assessments}, \texttt{change\_assessments}, and specific, machine-actionable \texttt{revision\_instructions}. Each task assessment reports the task identifier, solvability, ordered dependency route, and principal risk. Use \texttt{accept} only when the hard-failure list is empty, \texttt{revise} when a bounded specification change can repair the candidate, and \texttt{reject} when its central design must be replaced.

Audit spawn safety, traversal, the surface--sea relationship, biome recognizability, ore height ranges, every ore band, resource density, weighted tree placement, controlled decorations, wood and fuel availability, cows and leather, sugar cane, diamond, lapis, obsidian, and selected-task feasibility. A resource generated outside usable terrain does not count. Distinguish valid biome character from needless motor difficulty. Treat the ZIP as a deterministic serialization of the specification, and judge only the transitions and task routes represented by the supplied specification and static compiler report without assuming any unlisted resource or hidden interface.
\par}

\subsection{Rule-Intervention Prompts}

The rule-intervention prompts use the same structured response contracts.
The Designer receives the user constraints, allowed catalog, current specification, static compiler/reachability report, and preceding Judge report; when no preceding report exists, its verdict is \texttt{not\_run}. The Judge receives the candidate specification and static compiler report.
The rule-intervention catalog contains the permitted operators, editable targets, processing inputs and stations, output items, and dependency routes for the selected tasks.

\paragraph{Designer prompt.}
{\itshape
You are the MirrorCraft LLM Designer. Design a complete, machine-readable candidate specification for Minecraft Java 1.19.0. The static compiler writes the datapack. Output no commands, source code, file paths, Markdown fences, comments, or prose outside one JSON object. The object contains \texttt{schema\_version}, \texttt{kind}, a brief \texttt{design\_summary}, one \texttt{resolved\_feedback} list item per addressed problem, and a complete replacement \texttt{spec}; the outer object and nested specification both use schema version \texttt{1.0} and kind \texttt{mirror}. Preserve the requested \texttt{selected\_tasks}, keep every numeric field inside the supplied bounds, treat the static report as authoritative, and correct every hard failure.

Make each changed transition normally encounterable while an agent pursues at least one selected task, and keep every selected task solvable through Mineflayer-observable interaction. Prefer two to six interpretable changes, combining quantity, coproduct, or route modifications only when their interaction remains auditable. Restrict edits to the supplied operators, targets, stations, inputs, and outputs, without modifying unrelated food, decoration, redstone, village, Nether, or End mechanics. The changed rule must be learnable from actual action outcomes rather than recipe, loot-table, datapack, recipe-book, or hidden-state inspection. Avoid changes that make completion immediate or require an item absent from the allowed dependency graph. Return one corrected full specification from the supplied constraints, catalog, current specification, static report, and preceding Judge report.
\par}

\paragraph{Judge prompt.}
{\itshape
You are the independent MirrorCraft LLM Judge. Audit the compiled Minecraft Java 1.19.0 candidate without redesigning the datapack or inspecting agent outcomes. Be conservative, use the static report as hard evidence, and output no Markdown or prose outside one JSON object. The object uses schema version \texttt{1.0} and kind \texttt{mirror}, and contains \texttt{verdict}, \texttt{confidence}, \texttt{summary}, \texttt{hard\_failures}, \texttt{warnings}, \texttt{task\_assessments}, \texttt{change\_assessments}, and specific, machine-actionable \texttt{revision\_instructions}. Each task assessment reports the task identifier, solvability, ordered dependency route, and principal risk. Use \texttt{accept} only when the hard-failure list is empty, \texttt{revise} when a bounded specification change can repair the candidate, and \texttt{reject} when its central design must be replaced.

For every rule change, audit task relevance, encounterability before task completion, observable outcome, the complete modified dependency route, interactions with other changes, and difficulty. Reject changes unrelated to all selected routes and reject a pack when any selected task is unreachable. Treat extreme scarcity or shortcuts as revision requirements unless explicitly requested. The agent may observe action inputs and actual outputs but cannot inspect recipes, loot tables, datapack manifests, or the compiler's rule sheet. Treat the ZIP as a deterministic serialization of the specification, and judge only the transitions and task routes represented by the supplied specification and static compiler report without assuming any unlisted resource or hidden interface.
\par}

\begin{figure*}[t]
\centering
\includegraphics[width=0.92\textwidth]{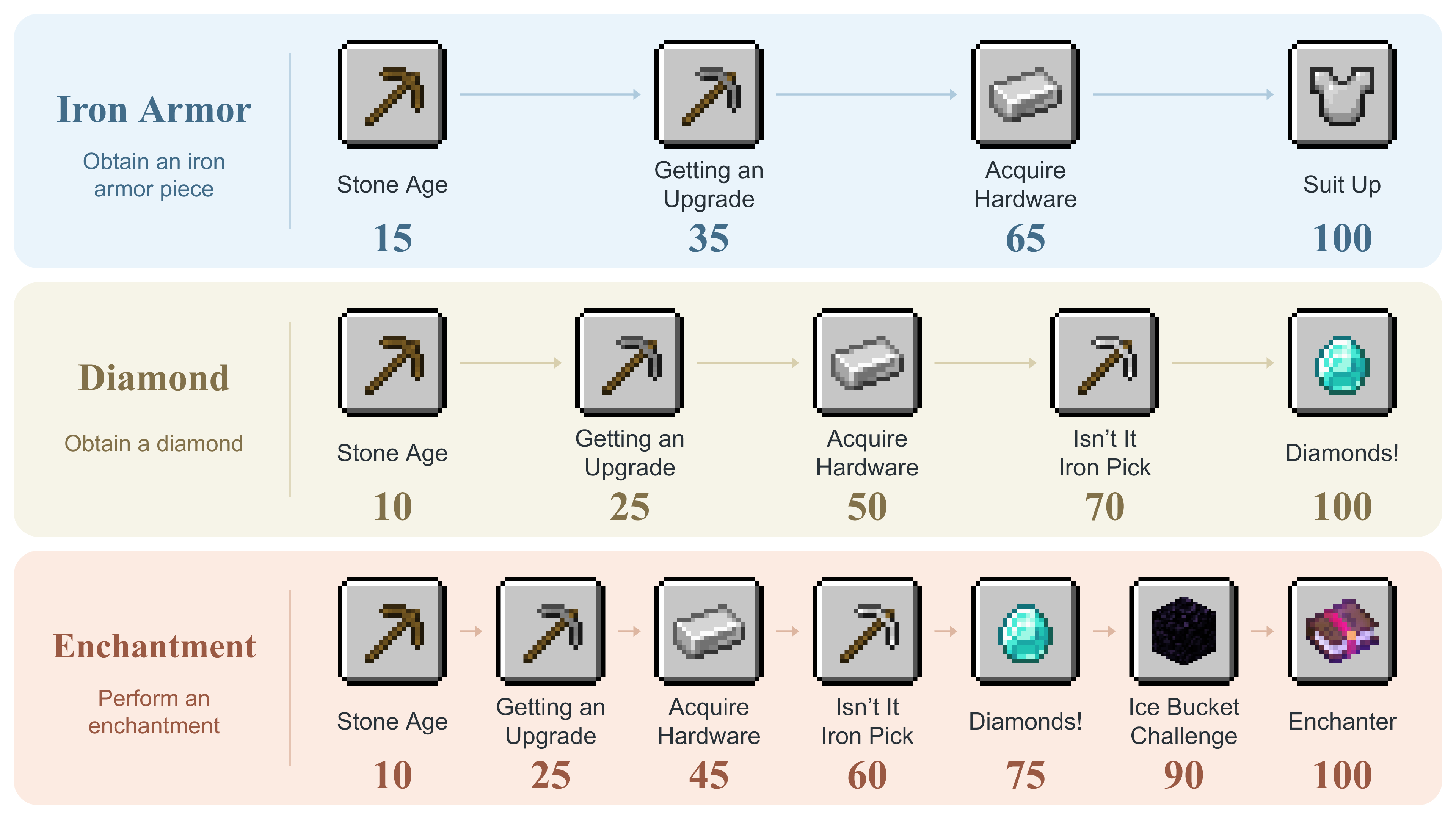}
\caption{Task milestones, which are built-in advancements verified by the server, and their Score values.}
\label{fig:checkpoints}
\end{figure*}

\section{Advancement Milestones}
\label{app:checkpoints}

MirrorCraft measures ordered task progress with the server-verified milestones in Figure~\ref{fig:checkpoints}.
Each task has a fixed milestone sequence that follows its main resource and tool dependencies.
An episode receives the value of its deepest completed milestone rather than the sum of all completed milestones.
The final milestone has value 100 and indicates task completion.
This design preserves partial progress when an episode reaches a late prerequisite but does not finish the full task.
All milestone identifiers below use the \texttt{minecraft:story/} namespace.

\paragraph{Iron Armor.}
{\raggedright
The task is completed by obtaining any one iron armor piece rather than a complete four-piece set.
The ordered codes used by the game are \texttt{mine\_stone}, \texttt{upgrade\_tools}, \texttt{smelt\_iron}, and \texttt{obtain\_armor}.\par}

\paragraph{Diamond.}
{\raggedright
The ordered codes are \texttt{mine\_stone}, \texttt{upgrade\_tools}, \texttt{smelt\_iron}, \texttt{iron\_tools}, and \texttt{mine\_diamond}.\par}

\paragraph{Enchantment.}
{\raggedright
The ordered codes are \texttt{mine\_stone}, \texttt{upgrade\_tools}, \texttt{smelt\_iron}, \texttt{iron\_tools}, \texttt{mine\_diamond}, \texttt{form\_obsidian}, and \texttt{enchant\_item}.\par}

The milestone sequence is shown to the agent, while the server independently verifies completion.

\section{Detailed Behavioral Diagnostics}
\label{app:behavior-details}

\begin{table*}[t]
\caption{IRR (\%) and RL (semantic actions) by rule suite and configuration, averaged over models, tasks, and Mirror worlds.
Average weights M01--M06 equally; only ReAct w/rules receives rule descriptions.}
\label{tab:irr-rl-detailed}
\centering
{\footnotesize
\setlength{\tabcolsep}{2.1pt}
\renewcommand{\arraystretch}{1.02}
\begin{tabular*}{\textwidth}{@{\extracolsep{\fill}}l*{7}{cc}@{}}
\toprule
& \multicolumn{2}{c}{M01}
& \multicolumn{2}{c}{M02}
& \multicolumn{2}{c}{M03}
& \multicolumn{2}{c}{M04}
& \multicolumn{2}{c}{M05}
& \multicolumn{2}{c}{M06}
& \multicolumn{2}{c}{Average}\\
\cmidrule(lr){2-3}
\cmidrule(lr){4-5}
\cmidrule(lr){6-7}
\cmidrule(lr){8-9}
\cmidrule(lr){10-11}
\cmidrule(lr){12-13}
\cmidrule(lr){14-15}
Configuration
& IRR & RL & IRR & RL & IRR & RL & IRR & RL
& IRR & RL & IRR & RL & IRR & RL\\
\midrule
\direct{}
& 0.7\% & 14.0 & 5.3\% & 3.4 & 0.3\% & 4.3
& 0.0\% & 7.2 & 0.2\% & 4.5 & 0.1\% & 3.8 & 1.1\% & 6.2\\
ReAct
& 10.7\% & 15.3 & 24.0\% & 4.1 & 1.8\% & 3.9
& 4.2\% & 9.1 & 1.1\% & 6.0 & 0.7\% & 5.0 & 7.1\% & 7.2\\
Reflexion
& 7.7\% & 15.2 & 20.8\% & 4.0 & 3.3\% & 3.7
& 3.1\% & 8.9 & 0.8\% & 5.6 & 0.5\% & 5.1 & 6.0\% & 7.1\\
Voyager
& 11.3\% & 15.0 & 35.9\% & 3.8 & 7.8\% & 3.7
& 11.5\% & 8.1 & 3.0\% & 5.4 & 1.8\% & 4.8 & 11.9\% & 6.8\\
XENON
& 11.7\% & 15.0 & 33.3\% & 3.3 & 5.3\% & 3.9
& 9.1\% & 7.5 & 4.6\% & 5.8 & 1.4\% & 4.3 & 10.9\% & 6.6\\
ADAM
& 7.0\% & 13.7 & 33.7\% & 4.0 & 4.7\% & 3.7
& 7.6\% & 7.6 & 2.0\% & 5.1 & 1.2\% & 4.5 & 9.4\% & 6.4\\
ReAct w/rules
& 18.9\% & 10.5 & 26.6\% & 4.2 & 3.4\% & 3.1
& 7.5\% & 6.0 & 1.7\% & 4.3 & 1.2\% & 4.0 & 9.9\% & 5.4\\
\bottomrule
\end{tabular*}}
\end{table*}

The results for each rule suite below retain the same encounter definitions used in the benchmark.
Each action $a$ is mapped to the semantic signature $\sigma(a)$ described in \mainref{Section~3.6 of the main paper}{Section~\ref{sec:behavior}}, which ignores crafting quantity, smelting fuel and quantity, mining quantity and coordinates, attack entity identity, and placement coordinates.
Let $\mathcal E^+$ contain the rule encounters followed by another budgeted action, and let $i_e$ be the index of the action that triggers encounter $e$.
Then
\begin{equation}
\operatorname{IRR}
=\frac{1}{|\mathcal E^+|}
\sum_{e\in\mathcal E^+}
\mathbf{1}\!\left[\sigma(a_{i_e+1})=\sigma(a_{i_e})\right].
\end{equation}
For encounter $e$, let $S_k$ be the milestone Score after action $k$ in the same episode, with $S_0$ denoting the initial Score, and let $s_e^-=S_{i_e-1}$ be the Score immediately before the encounter action.
We define $j_e=\min\{j\geq i_e:S_j>s_e^-\}$, so $j_e=i_e$ when the encounter action itself raises the Score, and let $\mathcal R$ contain the encounters for which $j_e$ exists.
Then
\begin{equation}
\operatorname{RL}
=\frac{1}{|\mathcal R|}
\sum_{e\in\mathcal R}(j_e-i_e).
\end{equation}
IRR thus compares the semantic signature of the action that triggers a rule with that of the next action counted against the budget, and RL counts actions until the first higher milestone and includes only encounters followed by such an increase.
The final columns weight M01--M06 equally.

M02 produces the highest IRR of any suite while retaining short RL.
This pattern is consistent with useful byproducts, for which repeating the same semantic action can remain productive.
M01 instead yields the longest RL for every configuration evaluated with hidden rules.
Because RL also depends on how far a changed transition lies from the next milestone, differences in RL between suites partly reflect the routes themselves, whereas comparisons within a suite, such as between ReAct and ReAct w/rules, hold this factor fixed.
Across suites, rule disclosure raises ReAct's IRR from 7.1\% to 9.9\% while reducing RL from 7.2 to 5.4 actions.
The two diagnostics therefore capture different behavior, and IRR should not be read as an error rate.

\begin{table*}[t]
\caption{Biome-specific world-generation parameters. The gradient maps density $1.0$ at the first height to $-1.0$ at the second; noise weights apply to the broad-contour and local-undulation fields. Tree counts are placement attempts per chunk, drawn from two values with the stated expectation.}
\label{tab:biome-params}
\centering
{\footnotesize
\setlength{\tabcolsep}{3pt}
\renewcommand{\arraystretch}{1.15}
\begin{tabular*}{\textwidth}{@{\extracolsep{\fill}}
  >{\raggedright\arraybackslash}p{0.085\textwidth}
  >{\centering\arraybackslash}p{0.075\textwidth}
  >{\centering\arraybackslash}p{0.065\textwidth}
  >{\raggedright\arraybackslash}p{0.185\textwidth}
  >{\raggedright\arraybackslash}p{0.12\textwidth}
  >{\raggedright\arraybackslash}p{0.13\textwidth}
  >{\raggedright\arraybackslash}p{0.21\textwidth}@{}}
\toprule
Biome & \shortstack{Gradient\\$Y$ bounds} & \shortstack{Noise\\weights} & Surface patches (noise intervals) & \shortstack[l]{Trees per chunk\\(expected)} & \shortstack[l]{Other attempts\\per chunk} & Rarity features (one per $N$ chunks)\\
\midrule
Plains & 50 to 90 & 0.56, 0.18 & podzol $[-.95,-.80]$; coarse dirt $[.51,.60]$, $[.66,.79]$; gravel $[.88,.95]$ & plains trees: 0 or 1 (0.5) & sugar cane 4; water lily 2 & surface lake 15; large ferns 12; mossy boulder 18; pumpkins 180\\
Taiga & 48 to 94 & 0.60, 0.20 & podzol $[-.95,-.80]$; coarse dirt $[.51,.60]$, $[.66,.79]$; moss $[.88,.95]$ & spruce: 14 or 15 (14.1) & sugar cane 4; large fern 2; water lily 1 & berries 12; surface lake 15; mossy boulder 8; huge brown mushroom 48; pumpkins 160\\
Snowy Plains & 50 to 88 & 0.45, 0.14 & podzol $[-.95,-.80]$; coarse dirt $[.51,.60]$, $[.66,.79]$; gravel $[.88,.95]$ & spruce: 0 or 1 (0.5) & sugar cane 6 & large ferns 14; berry bushes 24; pumpkins 200; mossy boulder 24; ice patch 18; ice spike 56; surface lake 18\\
Jungle & 46 to 94 & 0.65, 0.22 & moss $[-.95,-.80]$; coarse dirt $[.51,.60]$, $[.66,.79]$; gravel $[.88,.95]$ & jungle trees: 68 or 69 (68.1) & sugar cane 5; large fern 2; water lily 2 & melon 4; mossy boulder 10; huge brown mushroom 40; surface lake 14\\
Savanna & 46 to 96 & 0.68, 0.22 & coarse dirt $[-.95,-.80]$, $[.51,.60]$; gravel $[.66,.79]$; terracotta $[.88,.95]$ & acacia: 2 or 3 (2.5) & sugar cane 6; water lily 1 & melon 16; pumpkins 180; mossy boulder 14; small well 128; surface lake 16\\
\bottomrule
\end{tabular*}}
\end{table*}

\begin{table}[t]
\caption{Custom resource bands shared by all five biomes. Attempts count sampled origins per chunk, size is the configured ore-feature parameter, and the last column bounds the sampled origin rather than every generated block.}
\label{tab:ore-bands}
\centering
{\footnotesize
\setlength{\tabcolsep}{4pt}
\renewcommand{\arraystretch}{1.0}
\begin{tabular}{@{}lccc@{}}
\toprule
Band & Attempts per chunk & Size & Origin $Y$ (inclusive)\\
\midrule
Coal (accessible) & 96 & 20 & 44--112\\
Iron (accessible) & 84 & 16 & 40--92\\
Copper (accessible) & 72 & 20 & 40--96\\
Gold (accessible) & 48 & 12 & 32--72\\
Redstone (accessible) & 48 & 14 & 28--68\\
Lapis (accessible) & 36 & 12 & 30--68\\
Emerald (accessible) & 24 & 8 & 40--80\\
Diamond (main) & 40 & 10 & 36--68\\
Diamond (shallow) & 8 & 6 & 52--72\\
Obsidian (main) & 8 & 12 & 36--64\\
Obsidian (shallow) & 10 & 10 & 50--72\\
\bottomrule
\end{tabular}}
\end{table}

\section{Controlled Biome Datapacks}
\label{app:biome-details}

All five controlled biomes target Minecraft Java 1.19.0 (pack format 10).
Each datapack defines a single-biome Overworld preset spanning $Y=-64$ to $319$, with sea level 63, aquifers, and the standard \path{minecraft:cave} and \path{minecraft:cave_extra_underground} carvers; every Overworld climate point maps to the controlled biome, while the Nether and End keep their Vanilla generators.
The setup function selects Peaceful difficulty, fixes the time at noon and the weather to clear, retains natural mob spawning, and disables patrol, wandering-trader, and insomnia spawning.
The archive must be enabled alone when a new world is created, because existing chunks do not acquire its terrain, resources, or vegetation retroactively.

Terrain density combines a vertical gradient from $1.0$ to $-1.0$ with cached broad-contour and local-undulation noise fields, whose first octaves are $-8$ and $-5$ and whose amplitudes are $[1,.5,.25]$ and $[1,.5]$; the blended density is scaled by 0.64.
Table~\ref{tab:biome-params} lists the gradient bounds and noise weights of each biome, from the lowest relief in Snowy Plains to the widest vertical range in Savanna.
Grass over dirt is the default surface, and surface-noise intervals introduce the patches in the table.
Every biome retains Vanilla amethyst geodes, springs, glow lichen, stone variants, and sediments.
Tree placements reject positions with nonzero surface-water depth, and task resources are generated rather than placed at fixed coordinates.
Cows spawn with weight 24 in groups of four, and their base loot provides 0--2 leather before looting modifiers.
Boulders, mushrooms, wells, and ice spikes are loot-free placed features rather than structures, so they contain no chests or task rewards and are unaffected by the Generate Structures setting.

The biomes also keep their characteristic ecology.
Plains retains its ordinary flower, grass, mushroom, and pumpkin patches.
Snowy Plains has temperature 0.0 and snow precipitation, and the retained \path{minecraft:freeze_top_layer} feature applies the visible snow and ice; sheep, pigs, chickens, rabbits, bats, and glow squid remain in their spawn categories.
Jungle retains bamboo, vines, warm flowers, Jungle grass, mushrooms, pumpkins, and a melon patch, and parrots and pandas remain; its high tree count produces the dense canopy and short sight lines.
Savanna has temperature 2.0, downfall 0.0, and no precipitation, and horses and donkeys remain.

Every biome retains the ordinary Vanilla placements for coal, iron, copper, gold, redstone, lapis, and diamond and appends the 11 custom bands in Table~\ref{tab:ore-bands}, so the table is not a complete ore distribution; emerald and obsidian occur only through the custom bands.
Each band applies \texttt{count}, \texttt{in\_square}, an inclusive uniform height range, and a biome filter, and its configured feature replaces stone- or deepslate-replaceable blocks with air-exposure discard 0.0.
Attempts can fail or overlap, so configured size is not a realized block count.
The main and shallow diamond and obsidian bands reduce search and descent costs without placing task items in inventories or containers.
Setting \texttt{ore\_veins\_enabled} to false disables Minecraft's separate large noise-vein system, not the ordinary Vanilla ore placements.

\section{Rule Suite Manifests and Detailed Results}
\label{app:mirror-details}

\begin{figure*}[p]
\centering
\begingroup
\setlength{\tabcolsep}{0pt}%
\begin{tabular*}{\textwidth}{@{\extracolsep{\fill}}p{0.49\textwidth}p{0.49\textwidth}@{}}
{\centering
\includegraphics[width=0.97\linewidth,height=42mm,keepaspectratio]{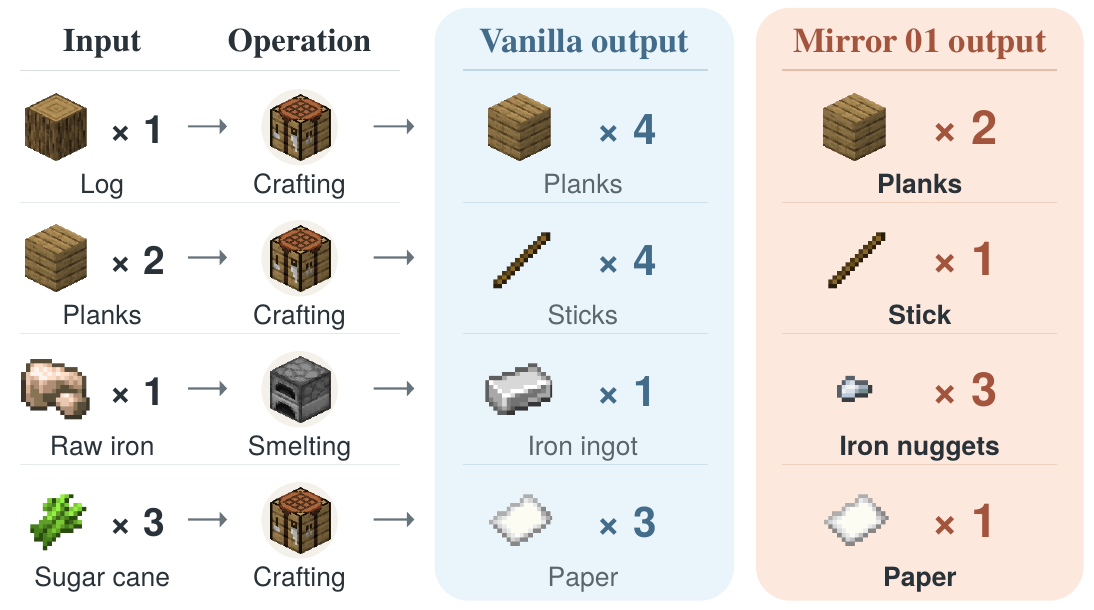}\par
\vspace{0.2mm}
{\small\bfseries M01: Quantity scarcity}\par
\vspace{0.2mm}
{\footnotesize\raggedright
\textbf{Rules.}
One trunk yields two matching planks; two planks yield one stick; raw iron smelts into three nuggets; three sugar canes yield one paper.
\textbf{Effect.}
Lower yields add gathering and repeated processing to otherwise familiar crafting and smelting routes.\par}}
&
{\centering
\includegraphics[width=0.97\linewidth,height=42mm,keepaspectratio]{figures/Mirror02.pdf}\par
\vspace{0.2mm}
{\small\bfseries M02: Byproduct shortcuts}\par
\vspace{0.2mm}
{\footnotesize\raggedright
\textbf{Rules.}
A trunk drops four matching planks; iron ore adds an ingot; diamond ore adds obsidian; sugar cane adds leather; the book recipe yields an enchanting table.
\textbf{Effect.}
Task resources become byproducts or substituted outputs, exposing useful noncanonical routes.\par}}
\tabularnewline[0.2mm]
{\centering
\includegraphics[width=0.97\linewidth,height=42mm,keepaspectratio]{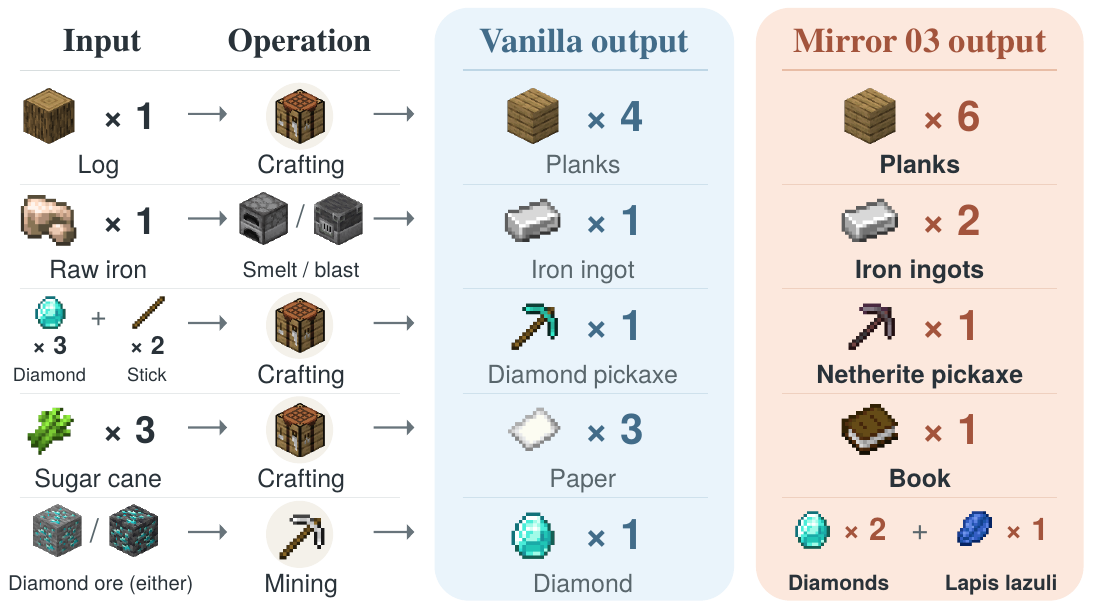}\par
\vspace{0.2mm}
{\small\bfseries M03: Yield expansion}\par
\vspace{0.2mm}
{\footnotesize\raggedright
\textbf{Rules.}
One trunk yields six matching planks; smelting or blasting raw iron yields two ingots; the diamond-pickaxe recipe yields a netherite pickaxe; three sugar canes yield one book; each diamond-ore variant yields two diamonds and one lapis lazuli.
\textbf{Effect.}
Expanded and substituted outputs shorten collection, crafting, and smelting routes.\par}}
&
{\centering
\includegraphics[width=0.97\linewidth,height=42mm,keepaspectratio]{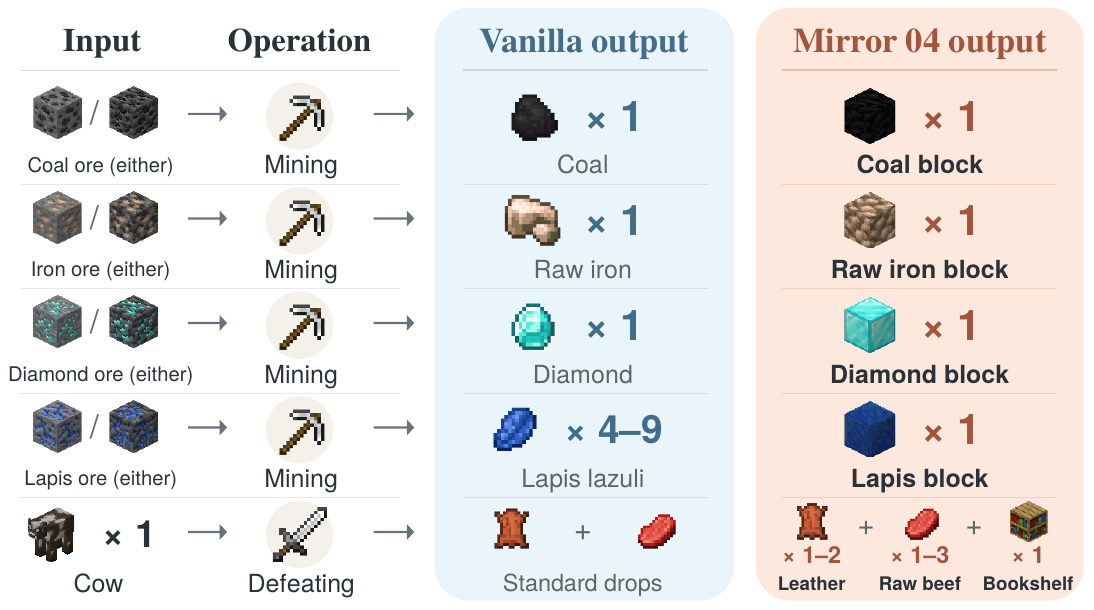}\par
\vspace{0.2mm}
{\small\bfseries M04: Compressed resource drops}\par
\vspace{0.2mm}
{\footnotesize\raggedright
\textbf{Rules.}
Coal, iron, diamond, and lapis ore variants respectively drop a coal block, raw-iron block, diamond block, and lapis block; cows drop one to two leather, one to three beef, and one bookshelf.
\textbf{Effect.}
Mining yields compressed resources that can be unpacked through crafting into ordinary task inputs, while cows create a direct bookshelf route.\par}}
\tabularnewline[0.2mm]
{\centering
\includegraphics[width=0.97\linewidth,height=42mm,keepaspectratio]{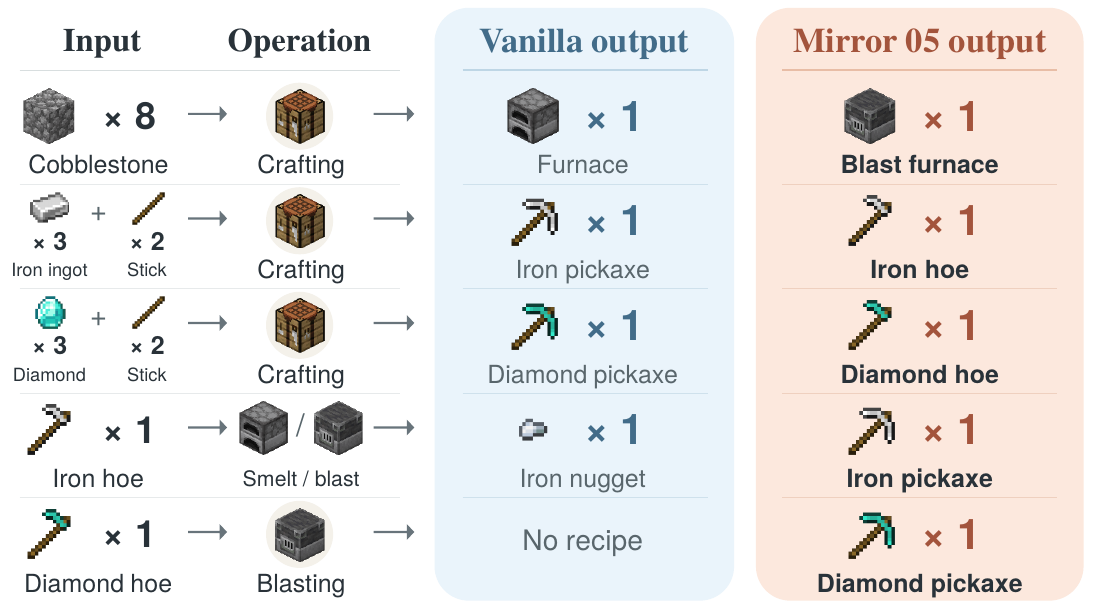}\par
\vspace{0.2mm}
{\small\bfseries M05: Route substitution}\par
\vspace{0.2mm}
{\footnotesize\raggedright
\textbf{Rules.}
The furnace recipe yields a blast furnace; iron- and diamond-pickaxe recipes yield their corresponding hoes; smelting or blasting an iron hoe yields an iron pickaxe, while blasting a diamond hoe yields a diamond pickaxe.
\textbf{Effect.}
Pickaxe production becomes an ordered detour through hoes and furnace processing.\par}}
&
{\centering
\includegraphics[width=0.97\linewidth,height=42mm,keepaspectratio]{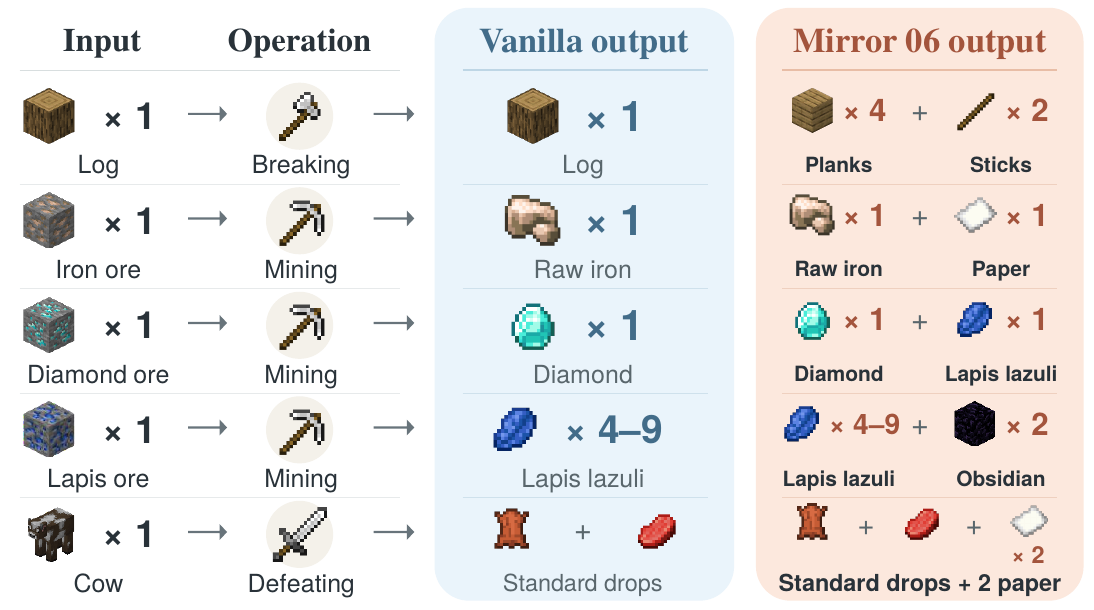}\par
\vspace{0.2mm}
{\small\bfseries M06: Byproduct redistribution}\par
\vspace{0.2mm}
{\footnotesize\raggedright
\textbf{Rules.}
Each trunk drops four matching planks and two sticks; iron ore adds one paper; diamond ore adds one lapis lazuli; lapis ore adds two obsidian; cows retain their standard drops and add two paper.
\textbf{Effect.}
Paper, lapis, obsidian, and sticks become byproducts of ordinary gathering.\par}}
\tabularnewline
\end{tabular*}
\endgroup
\caption{Changed transitions, disclosed rules, and route effects of the six rule suites.}
\label{fig:suite-cards}
\end{figure*}

This section documents how each Mirror world is built.
Each card in Figure~\ref{fig:suite-cards} shows the changed transitions of one rule suite, states its rules as they are disclosed to ReAct w/rules, and summarizes the effect on task routes.
Table~\ref{tab:main-detailed} gives the matched Vanilla references, and Tables~\ref{tab:details-mirror01}--\ref{tab:details-mirror06} report detailed Mirror results for every suite.

\begin{table*}[p]
\caption{Results for M01 (Quantity scarcity) by task, agent configuration, and model.}
\label{tab:details-mirror01}
\centering
{\footnotesize
\setlength{\tabcolsep}{3pt}
\renewcommand{\arraystretch}{1.05}
\begin{tabular*}{\textwidth}{@{\extracolsep{\fill}}llccc@{\hspace{10pt}}ccc@{}}
\toprule
& & \multicolumn{3}{c}{Gemini} & \multicolumn{3}{c}{DeepSeek}\\
\cmidrule(lr){3-5}\cmidrule(lr){6-8}
Task & Configuration & Score\,/\,SR & Actions & Cost
& Score\,/\,SR & Actions & Cost\\
\midrule
\multirow{7}{*}{Iron Armor} & Direct LLM & 54.0\,/\,0.0 & 50.0 & 0.04 & 40.3\,/\,0.0 & 48.0 & 0.01 \\
 & ReAct & 77.2\,/\,43.3 & 47.6 & 0.08 & 78.3\,/\,60.0 & 45.2 & 0.03 \\
 & ReAct w/rules & 84.8\,/\,56.7 & 46.5 & 0.07 & 84.5\,/\,63.3 & 44.9 & 0.03 \\
 & Reflexion & 85.2\,/\,63.3 & 45.5 & 0.07 & 76.3\,/\,46.7 & 46.9 & 0.03 \\
 & Voyager & 84.2\,/\,63.3 & 42.7 & 0.06 & 75.3\,/\,46.7 & 45.3 & 0.02 \\
 & XENON & 45.2\,/\,0.0 & 48.7 & 0.09 & 68.2\,/\,36.7 & 47.1 & 0.04 \\
 & ADAM & 57.5\,/\,10.0 & 49.1 & 0.14 & 71.2\,/\,36.7 & 46.7 & 0.04 \\
\midrule
\multirow{7}{*}{Diamond} & Direct LLM & 58.2\,/\,0.0 & 50.0 & 0.04 & 40.7\,/\,0.0 & 50.0 & 0.01 \\
 & ReAct & 84.7\,/\,53.3 & 47.0 & 0.08 & 88.3\,/\,63.3 & 47.4 & 0.03 \\
 & ReAct w/rules & 92.0\,/\,73.3 & 46.2 & 0.07 & 87.0\,/\,56.7 & 47.3 & 0.03 \\
 & Reflexion & 88.7\,/\,66.7 & 46.8 & 0.07 & 77.3\,/\,26.7 & 48.3 & 0.03 \\
 & Voyager & 87.3\,/\,60.0 & 47.0 & 0.06 & 72.2\,/\,23.3 & 49.6 & 0.02 \\
 & XENON & 59.0\,/\,10.0 & 49.9 & 0.09 & 69.7\,/\,20.0 & 49.7 & 0.05 \\
 & ADAM & 64.8\,/\,3.3 & 50.0 & 0.14 & 81.0\,/\,36.7 & 49.2 & 0.05 \\
\midrule
\multirow{7}{*}{Enchantment} & Direct LLM & 48.0\,/\,0.0 & 75.0 & 0.10 & 23.7\,/\,0.0 & 75.0 & 0.02 \\
 & ReAct & 82.3\,/\,3.3 & 75.0 & 0.17 & 73.0\,/\,0.0 & 75.0 & 0.06 \\
 & ReAct w/rules & 80.7\,/\,6.7 & 75.0 & 0.16 & 73.5\,/\,0.0 & 75.0 & 0.06 \\
 & Reflexion & 77.5\,/\,0.0 & 75.0 & 0.18 & 53.3\,/\,0.0 & 75.0 & 0.06 \\
 & Voyager & 64.8\,/\,0.0 & 75.0 & 0.13 & 44.2\,/\,0.0 & 75.0 & 0.05 \\
 & XENON & 69.3\,/\,0.0 & 75.0 & 0.22 & 73.5\,/\,0.0 & 75.0 & 0.09 \\
 & ADAM & 64.5\,/\,0.0 & 75.0 & 0.28 & 59.0\,/\,0.0 & 75.0 & 0.09 \\
\bottomrule
\end{tabular*}}
\end{table*}

\begin{table*}[p]
\caption{Results for M02 (Byproduct shortcuts) by task, agent configuration, and model.}
\label{tab:details-mirror02}
\centering
{\footnotesize
\setlength{\tabcolsep}{3pt}
\renewcommand{\arraystretch}{1.05}
\begin{tabular*}{\textwidth}{@{\extracolsep{\fill}}llccc@{\hspace{10pt}}ccc@{}}
\toprule
& & \multicolumn{3}{c}{Gemini} & \multicolumn{3}{c}{DeepSeek}\\
\cmidrule(lr){3-5}\cmidrule(lr){6-8}
Task & Configuration & Score\,/\,SR & Actions & Cost
& Score\,/\,SR & Actions & Cost\\
\midrule
\multirow{7}{*}{Iron Armor} & Direct LLM & 76.7\,/\,33.3 & 40.9 & 0.02 & 70.0\,/\,20.0 & 43.7 & 0.01 \\
 & ReAct & 97.7\,/\,93.3 & 25.9 & 0.04 & 100.0\,/\,100.0 & 26.1 & 0.02 \\
 & ReAct w/rules & 98.8\,/\,96.7 & 25.1 & 0.04 & 100.0\,/\,100.0 & 26.2 & 0.02 \\
 & Reflexion & 100.0\,/\,100.0 & 23.8 & 0.04 & 96.5\,/\,90.0 & 28.0 & 0.02 \\
 & Voyager & 100.0\,/\,100.0 & 22.9 & 0.03 & 97.7\,/\,93.3 & 27.5 & 0.01 \\
 & XENON & 72.3\,/\,26.7 & 42.5 & 0.05 & 97.7\,/\,93.3 & 28.9 & 0.02 \\
 & ADAM & 76.2\,/\,43.3 & 40.2 & 0.07 & 94.2\,/\,83.3 & 30.5 & 0.02 \\
\midrule
\multirow{7}{*}{Diamond} & Direct LLM & 76.0\,/\,26.7 & 43.5 & 0.02 & 51.8\,/\,0.0 & 50.0 & 0.01 \\
 & ReAct & 99.0\,/\,96.7 & 26.2 & 0.04 & 93.0\,/\,76.7 & 33.1 & 0.02 \\
 & ReAct w/rules & 98.0\,/\,93.3 & 26.3 & 0.04 & 99.0\,/\,96.7 & 26.4 & 0.02 \\
 & Reflexion & 97.0\,/\,90.0 & 29.0 & 0.04 & 93.0\,/\,76.7 & 34.3 & 0.02 \\
 & Voyager & 94.0\,/\,80.0 & 29.8 & 0.03 & 87.0\,/\,56.7 & 37.5 & 0.01 \\
 & XENON & 79.3\,/\,33.3 & 41.4 & 0.05 & 84.0\,/\,46.7 & 39.3 & 0.03 \\
 & ADAM & 83.7\,/\,50.0 & 41.9 & 0.08 & 86.0\,/\,53.3 & 38.9 & 0.03 \\
\midrule
\multirow{7}{*}{Enchantment} & Direct LLM & 63.5\,/\,0.0 & 75.0 & 0.05 & 32.8\,/\,0.0 & 70.9 & 0.01 \\
 & ReAct & 91.2\,/\,36.7 & 67.0 & 0.10 & 77.0\,/\,0.0 & 71.0 & 0.03 \\
 & ReAct w/rules & 91.0\,/\,30.0 & 67.1 & 0.10 & 80.2\,/\,3.3 & 69.5 & 0.03 \\
 & Reflexion & 83.5\,/\,10.0 & 72.0 & 0.10 & 59.5\,/\,0.0 & 71.1 & 0.03 \\
 & Voyager & 82.2\,/\,6.7 & 72.6 & 0.06 & 59.8\,/\,0.0 & 70.8 & 0.03 \\
 & XENON & 77.3\,/\,3.3 & 74.4 & 0.13 & 77.3\,/\,3.3 & 70.1 & 0.05 \\
 & ADAM & 76.3\,/\,3.3 & 75.0 & 0.16 & 74.0\,/\,0.0 & 70.7 & 0.05 \\
\bottomrule
\end{tabular*}}
\end{table*}

\begin{table*}[p]
\caption{Results for M03 (Yield expansion) by task, agent configuration, and model.}
\label{tab:details-mirror03}
\centering
{\footnotesize
\setlength{\tabcolsep}{3pt}
\renewcommand{\arraystretch}{1.05}
\begin{tabular*}{\textwidth}{@{\extracolsep{\fill}}llccc@{\hspace{10pt}}ccc@{}}
\toprule
& & \multicolumn{3}{c}{Gemini} & \multicolumn{3}{c}{DeepSeek}\\
\cmidrule(lr){3-5}\cmidrule(lr){6-8}
Task & Configuration & Score\,/\,SR & Actions & Cost
& Score\,/\,SR & Actions & Cost\\
\midrule
\multirow{7}{*}{Iron Armor} & Direct LLM & 78.2\,/\,43.3 & 42.4 & 0.03 & 74.5\,/\,30.0 & 43.2 & 0.01 \\
 & ReAct & 96.5\,/\,90.0 & 34.6 & 0.06 & 98.8\,/\,96.7 & 36.6 & 0.02 \\
 & ReAct w/rules & 98.8\,/\,96.7 & 32.7 & 0.06 & 100.0\,/\,100.0 & 34.2 & 0.02 \\
 & Reflexion & 98.8\,/\,96.7 & 34.2 & 0.06 & 98.8\,/\,96.7 & 34.3 & 0.03 \\
 & Voyager & 98.8\,/\,96.7 & 32.8 & 0.04 & 100.0\,/\,100.0 & 37.9 & 0.02 \\
 & XENON & 78.0\,/\,40.0 & 41.9 & 0.07 & 98.8\,/\,96.7 & 33.8 & 0.03 \\
 & ADAM & 80.3\,/\,46.7 & 45.2 & 0.11 & 97.7\,/\,93.3 & 36.7 & 0.03 \\
\midrule
\multirow{7}{*}{Diamond} & Direct LLM & 76.5\,/\,33.3 & 44.6 & 0.03 & 52.0\,/\,3.3 & 49.7 & 0.01 \\
 & ReAct & 99.0\,/\,96.7 & 37.2 & 0.06 & 100.0\,/\,100.0 & 39.5 & 0.02 \\
 & ReAct w/rules & 97.0\,/\,90.0 & 37.8 & 0.06 & 99.0\,/\,96.7 & 37.0 & 0.02 \\
 & Reflexion & 95.0\,/\,83.3 & 38.8 & 0.07 & 97.0\,/\,90.0 & 38.7 & 0.03 \\
 & Voyager & 96.0\,/\,86.7 & 38.8 & 0.04 & 85.7\,/\,56.7 & 42.7 & 0.02 \\
 & XENON & 68.7\,/\,10.0 & 48.3 & 0.08 & 86.5\,/\,60.0 & 41.8 & 0.04 \\
 & ADAM & 77.5\,/\,36.7 & 46.0 & 0.12 & 95.0\,/\,83.3 & 42.2 & 0.04 \\
\midrule
\multirow{7}{*}{Enchantment} & Direct LLM & 66.2\,/\,0.0 & 75.0 & 0.08 & 39.3\,/\,0.0 & 75.0 & 0.02 \\
 & ReAct & 91.5\,/\,20.0 & 73.8 & 0.13 & 83.0\,/\,0.0 & 75.0 & 0.04 \\
 & ReAct w/rules & 91.8\,/\,33.3 & 71.8 & 0.14 & 90.8\,/\,23.3 & 70.1 & 0.04 \\
 & Reflexion & 85.3\,/\,3.3 & 74.3 & 0.14 & 77.0\,/\,0.0 & 75.0 & 0.05 \\
 & Voyager & 85.7\,/\,16.7 & 71.0 & 0.10 & 66.5\,/\,0.0 & 75.0 & 0.04 \\
 & XENON & 80.8\,/\,13.3 & 74.0 & 0.18 & 82.3\,/\,3.3 & 74.7 & 0.07 \\
 & ADAM & 75.2\,/\,3.3 & 74.4 & 0.23 & 79.5\,/\,0.0 & 75.0 & 0.06 \\
\bottomrule
\end{tabular*}}
\end{table*}

\begin{table*}[p]
\caption{Results for M04 (Compressed resource drops) by task, agent configuration, and model.}
\label{tab:details-mirror04}
\centering
{\footnotesize
\setlength{\tabcolsep}{3pt}
\renewcommand{\arraystretch}{1.05}
\begin{tabular*}{\textwidth}{@{\extracolsep{\fill}}llccc@{\hspace{10pt}}ccc@{}}
\toprule
& & \multicolumn{3}{c}{Gemini} & \multicolumn{3}{c}{DeepSeek}\\
\cmidrule(lr){3-5}\cmidrule(lr){6-8}
Task & Configuration & Score\,/\,SR & Actions & Cost
& Score\,/\,SR & Actions & Cost\\
\midrule
\multirow{7}{*}{Iron Armor} & Direct LLM & 62.7\,/\,13.3 & 47.8 & 0.05 & 60.0\,/\,30.0 & 44.5 & 0.01 \\
 & ReAct & 82.7\,/\,53.3 & 44.3 & 0.08 & 50.5\,/\,13.3 & 46.6 & 0.04 \\
 & ReAct w/rules & 84.8\,/\,56.7 & 44.1 & 0.08 & 67.5\,/\,43.3 & 43.5 & 0.04 \\
 & Reflexion & 94.2\,/\,83.3 & 41.0 & 0.07 & 64.0\,/\,36.7 & 45.3 & 0.04 \\
 & Voyager & 93.0\,/\,80.0 & 40.2 & 0.05 & 77.7\,/\,53.3 & 44.9 & 0.03 \\
 & XENON & 60.3\,/\,6.7 & 48.7 & 0.09 & 75.2\,/\,46.7 & 43.3 & 0.05 \\
 & ADAM & 58.7\,/\,13.3 & 49.3 & 0.15 & 84.2\,/\,63.3 & 42.3 & 0.05 \\
\midrule
\multirow{7}{*}{Diamond} & Direct LLM & 55.7\,/\,0.0 & 50.0 & 0.05 & 47.0\,/\,23.3 & 47.0 & 0.01 \\
 & ReAct & 82.0\,/\,40.0 & 46.3 & 0.08 & 48.5\,/\,26.7 & 47.1 & 0.04 \\
 & ReAct w/rules & 87.0\,/\,56.7 & 47.3 & 0.09 & 58.0\,/\,30.0 & 46.4 & 0.04 \\
 & Reflexion & 89.0\,/\,63.3 & 46.3 & 0.08 & 52.8\,/\,23.3 & 48.6 & 0.04 \\
 & Voyager & 92.0\,/\,73.3 & 43.3 & 0.05 & 67.0\,/\,40.0 & 47.1 & 0.03 \\
 & XENON & 63.3\,/\,3.3 & 49.9 & 0.09 & 88.0\,/\,76.7 & 42.6 & 0.05 \\
 & ADAM & 64.5\,/\,10.0 & 49.5 & 0.15 & 70.0\,/\,56.7 & 45.8 & 0.05 \\
\midrule
\multirow{7}{*}{Enchantment} & Direct LLM & 51.3\,/\,0.0 & 75.0 & 0.10 & 22.5\,/\,0.0 & 74.8 & 0.02 \\
 & ReAct & 81.0\,/\,0.0 & 75.0 & 0.18 & 66.5\,/\,0.0 & 75.0 & 0.06 \\
 & ReAct w/rules & 81.3\,/\,3.3 & 75.0 & 0.18 & 69.3\,/\,0.0 & 75.0 & 0.07 \\
 & Reflexion & 75.7\,/\,6.7 & 75.0 & 0.19 & 53.7\,/\,0.0 & 75.0 & 0.07 \\
 & Voyager & 72.5\,/\,0.0 & 75.0 & 0.14 & 48.8\,/\,0.0 & 75.0 & 0.06 \\
 & XENON & 72.0\,/\,0.0 & 75.0 & 0.22 & 69.8\,/\,0.0 & 75.0 & 0.10 \\
 & ADAM & 65.0\,/\,0.0 & 75.0 & 0.31 & 62.2\,/\,0.0 & 75.0 & 0.09 \\
\bottomrule
\end{tabular*}}
\end{table*}

\begin{table*}[p]
\caption{Results for M05 (Route substitution) by task, agent configuration, and model.}
\label{tab:details-mirror05}
\centering
{\footnotesize
\setlength{\tabcolsep}{3pt}
\renewcommand{\arraystretch}{1.05}
\begin{tabular*}{\textwidth}{@{\extracolsep{\fill}}llccc@{\hspace{10pt}}ccc@{}}
\toprule
& & \multicolumn{3}{c}{Gemini} & \multicolumn{3}{c}{DeepSeek}\\
\cmidrule(lr){3-5}\cmidrule(lr){6-8}
Task & Configuration & Score\,/\,SR & Actions & Cost
& Score\,/\,SR & Actions & Cost\\
\midrule
\multirow{7}{*}{Iron Armor} & Direct LLM & 66.7\,/\,13.3 & 49.1 & 0.04 & 69.3\,/\,20.0 & 46.2 & 0.01 \\
 & ReAct & 95.3\,/\,86.7 & 42.1 & 0.07 & 98.8\,/\,96.7 & 40.9 & 0.03 \\
 & ReAct w/rules & 96.5\,/\,90.0 & 40.1 & 0.07 & 100.0\,/\,100.0 & 41.8 & 0.03 \\
 & Reflexion & 95.3\,/\,86.7 & 39.2 & 0.08 & 98.8\,/\,96.7 & 43.3 & 0.03 \\
 & Voyager & 97.7\,/\,93.3 & 39.3 & 0.05 & 100.0\,/\,100.0 & 42.0 & 0.02 \\
 & XENON & 67.8\,/\,16.7 & 47.6 & 0.09 & 95.3\,/\,86.7 & 41.3 & 0.04 \\
 & ADAM & 72.2\,/\,23.3 & 49.1 & 0.13 & 97.7\,/\,93.3 & 43.0 & 0.04 \\
\midrule
\multirow{7}{*}{Diamond} & Direct LLM & 58.7\,/\,3.3 & 50.0 & 0.04 & 50.7\,/\,0.0 & 50.0 & 0.01 \\
 & ReAct & 89.0\,/\,70.0 & 45.4 & 0.08 & 94.0\,/\,80.0 & 45.3 & 0.03 \\
 & ReAct w/rules & 97.0\,/\,90.0 & 45.1 & 0.08 & 95.0\,/\,83.3 & 46.4 & 0.03 \\
 & Reflexion & 90.0\,/\,73.3 & 45.7 & 0.08 & 88.3\,/\,63.3 & 44.7 & 0.03 \\
 & Voyager & 87.0\,/\,63.3 & 45.4 & 0.06 & 78.3\,/\,36.7 & 48.7 & 0.03 \\
 & XENON & 62.5\,/\,13.3 & 48.3 & 0.10 & 85.3\,/\,53.3 & 46.1 & 0.05 \\
 & ADAM & 72.0\,/\,20.0 & 49.7 & 0.15 & 86.0\,/\,53.3 & 47.7 & 0.05 \\
\midrule
\multirow{7}{*}{Enchantment} & Direct LLM & 59.0\,/\,0.0 & 75.0 & 0.09 & 34.7\,/\,0.0 & 75.0 & 0.02 \\
 & ReAct & 85.3\,/\,13.3 & 74.9 & 0.17 & 81.7\,/\,6.7 & 74.8 & 0.06 \\
 & ReAct w/rules & 87.7\,/\,16.7 & 74.8 & 0.17 & 85.2\,/\,6.7 & 75.0 & 0.06 \\
 & Reflexion & 81.8\,/\,3.3 & 75.0 & 0.18 & 65.8\,/\,0.0 & 75.0 & 0.07 \\
 & Voyager & 75.0\,/\,0.0 & 75.0 & 0.13 & 61.0\,/\,0.0 & 75.0 & 0.05 \\
 & XENON & 69.5\,/\,0.0 & 75.0 & 0.24 & 81.8\,/\,3.3 & 75.0 & 0.09 \\
 & ADAM & 65.5\,/\,0.0 & 75.0 & 0.29 & 72.5\,/\,0.0 & 75.0 & 0.08 \\
\bottomrule
\end{tabular*}}
\end{table*}

\begin{table*}[p]
\caption{Results for M06 (Byproduct redistribution) by task, agent configuration, and model.}
\label{tab:details-mirror06}
\centering
{\footnotesize
\setlength{\tabcolsep}{3pt}
\renewcommand{\arraystretch}{1.05}
\begin{tabular*}{\textwidth}{@{\extracolsep{\fill}}llccc@{\hspace{10pt}}ccc@{}}
\toprule
& & \multicolumn{3}{c}{Gemini} & \multicolumn{3}{c}{DeepSeek}\\
\cmidrule(lr){3-5}\cmidrule(lr){6-8}
Task & Configuration & Score\,/\,SR & Actions & Cost
& Score\,/\,SR & Actions & Cost\\
\midrule
\multirow{7}{*}{Iron Armor} & Direct LLM & 77.8\,/\,36.7 & 43.8 & 0.04 & 72.3\,/\,26.7 & 43.3 & 0.01 \\
 & ReAct & 95.3\,/\,86.7 & 38.7 & 0.07 & 100.0\,/\,100.0 & 36.5 & 0.03 \\
 & ReAct w/rules & 97.7\,/\,93.3 & 35.9 & 0.08 & 100.0\,/\,100.0 & 37.5 & 0.03 \\
 & Reflexion & 98.8\,/\,96.7 & 33.9 & 0.07 & 97.7\,/\,93.3 & 38.5 & 0.03 \\
 & Voyager & 100.0\,/\,100.0 & 35.0 & 0.05 & 98.8\,/\,96.7 & 39.1 & 0.02 \\
 & XENON & 70.8\,/\,16.7 & 47.0 & 0.09 & 94.2\,/\,83.3 & 38.2 & 0.04 \\
 & ADAM & 72.3\,/\,26.7 & 46.5 & 0.13 & 96.5\,/\,90.0 & 39.5 & 0.04 \\
\midrule
\multirow{7}{*}{Diamond} & Direct LLM & 72.0\,/\,20.0 & 48.1 & 0.05 & 60.8\,/\,13.3 & 46.7 & 0.01 \\
 & ReAct & 95.0\,/\,83.3 & 39.6 & 0.08 & 95.3\,/\,86.7 & 38.9 & 0.03 \\
 & ReAct w/rules & 98.0\,/\,93.3 & 38.3 & 0.07 & 95.3\,/\,86.7 & 41.1 & 0.03 \\
 & Reflexion & 97.0\,/\,90.0 & 40.0 & 0.08 & 98.0\,/\,93.3 & 39.4 & 0.03 \\
 & Voyager & 97.0\,/\,90.0 & 39.5 & 0.05 & 88.3\,/\,63.3 & 44.3 & 0.03 \\
 & XENON & 75.0\,/\,30.0 & 45.4 & 0.09 & 87.3\,/\,60.0 & 42.4 & 0.04 \\
 & ADAM & 80.0\,/\,40.0 & 47.7 & 0.14 & 86.0\,/\,60.0 & 45.6 & 0.05 \\
\midrule
\multirow{7}{*}{Enchantment} & Direct LLM & 71.0\,/\,0.0 & 75.0 & 0.09 & 40.5\,/\,0.0 & 75.0 & 0.02 \\
 & ReAct & 90.5\,/\,20.0 & 73.9 & 0.17 & 82.3\,/\,13.3 & 73.0 & 0.06 \\
 & ReAct w/rules & 92.3\,/\,33.3 & 72.0 & 0.16 & 85.7\,/\,6.7 & 73.9 & 0.06 \\
 & Reflexion & 86.7\,/\,16.7 & 74.3 & 0.18 & 68.3\,/\,0.0 & 75.0 & 0.07 \\
 & Voyager & 83.0\,/\,0.0 & 75.0 & 0.13 & 64.8\,/\,0.0 & 75.0 & 0.05 \\
 & XENON & 82.0\,/\,0.0 & 75.0 & 0.23 & 81.7\,/\,6.7 & 74.3 & 0.09 \\
 & ADAM & 75.0\,/\,0.0 & 75.0 & 0.30 & 73.8\,/\,3.3 & 74.1 & 0.09 \\
\bottomrule
\end{tabular*}}
\end{table*}

\clearpage
\section{Limitations}
\label{sec:limits}

\mirror{} evaluates decision making through a common semantic Mineflayer interface and does not measure visual perception or motor control.
The main study uses a fixed set of controlled worlds, so the biome analysis describes these worlds rather than arbitrary Minecraft seeds, and repeated trials add stochastic variation rather than environmental diversity.
Each rule suite combines several designed changes, so its effect reflects the suite as a whole rather than any individual edit, and the six suites do not exhaust the space of possible mechanics.
The three tasks share one progression route and its early milestones, so pooled results depend on the equal weighting of tasks.

Score records the deepest completed milestone on an ordinal scale, and we therefore interpret it together with SR and task-level results.
RIE averages over all valid episodes, including those that never trigger a changed rule, so it measures the total effect of a suite, which combines the changed route with how agents respond to it.
The disclosure contrast splits ReAct's RIE into the cost of hidden rules and the gap with exact rules.
Following Mars, we evaluate disclosure only for ReAct.
This condition combines a warning that the rules changed with their exact description, and the remaining gap does not separate longer routes from difficulty applying the disclosed rules.
The task interface lists the Vanilla milestones in every condition, so where a shortcut skips a listed milestone, as the M02 book recipe does for Enchantment, the list can steer agents toward the Vanilla route.
The within-suite comparison in Section~\ref{app:aggregate-results} shows how configurations differ under the same change, including how often their plans reach the changed transitions, but it does not replace disclosure for every configuration.
Our adaptations of Voyager, XENON, and ADAM clear their procedure libraries and graphs after every episode, so the results concern adaptation within an episode rather than the learning across episodes for which these methods were designed.
The intervals in Section~\ref{app:aggregate-results} are binomial and do not account for clustering by world, and with ten worlds and three repetitions, single-suite contrasts vary between models.
The main study uses two cost-efficient models with provider-side reasoning disabled, so explicit reasoning comes only from the agent configurations.
The additional GPT-5.6 Luna evaluation covers Diamond only, and the horizon study uses Gemini only.
Both additional studies cover different sets of Vanilla and Mirror worlds, so we compare configurations within each condition.
Finally, the current tasks do not cover building, collaboration, or other broader forms of Minecraft play.
\clearpage
\onecolumn
\section{Evaluation-Agent Prompt Templates}
\label{app:agent-prompts}

This section reproduces the accompanying prompt templates, including system and user messages, auxiliary-agent prompts, repair branches, and shared components.
Text inside the ruled blocks matches the accompanying Markdown files verbatim; line wrapping is typographic.
Placeholders such as \texttt{\{\{SKILLS\_BLOCK\}\}} and \texttt{\{\{INVENTORY\_JSON\}\}} denote content inserted at runtime, not additional instructions or observed values.
The Markdown files accompany the source, so the templates can be inspected without PDF line wrapping.

\paragraph{Rule suites and disclosure.}
MirrorCraft uses exactly six rule suites, M01--M06, documented in Section~\ref{app:mirror-details}.
The six rule sheets in Section~\ref{app:prompt-11} use these same identifiers and describe the corresponding rules provided to ReAct w/rules.
The reported disclosure comparison is ReAct with and without these rule descriptions.

\medskip
\noindent\textbf{Prompt index}\par\smallskip
{\small
\begin{tabular}{@{}lll@{}}
\toprule
Source file & Contents & Section\\
\midrule
\texttt{01\_base.md} & Direct LLM & \ref{app:prompt-01}\\
\texttt{02\_react.md} & ReAct & \ref{app:prompt-02}\\
\texttt{03\_reflexion.md} & Reflexion & \ref{app:prompt-03}\\
\texttt{04\_react\_rules.md} & ReAct with Rules & \ref{app:prompt-04}\\
\texttt{05\_voyager.md} & Voyager & \ref{app:prompt-05}\\
\texttt{06\_xenon.md} & XENON & \ref{app:prompt-06}\\
\texttt{07\_adam.md} & ADAM & \ref{app:prompt-07}\\
\texttt{08\_common\_observation\_history.md} & Shared Constraints, Observations, and History & \ref{app:prompt-08}\\
\texttt{09\_common\_skills.md} & Shared Skill Interface & \ref{app:prompt-09}\\
\texttt{10\_common\_task.md} & Task Templates & \ref{app:prompt-10}\\
\texttt{11\_common\_rule\_information.md} & Rule Information and Rule Sheets & \ref{app:prompt-11}\\
\texttt{12\_common\_output\_repair.md} & Output Repair & \ref{app:prompt-12}\\
\bottomrule
\end{tabular}}

\Needspace{9\baselineskip}
\subsection{Direct LLM}
\label{app:prompt-01}
\PromptSource{prompts/01_base.md}
\Needspace{5\baselineskip}
\subsubsection*{System}
\begin{Verbatim}[
    fontsize=\small,
    breaklines=true,
    breakanywhere=true,
    breakautoindent=true,
    breakindent=0pt,
    breaksymbolleft={},
    breaksymbolright={},
    breakanywheresymbolpre={},
    breakanywheresymbolpost={},
    frame=lines,
    framerule=0.3pt,
    framesep=4pt,
    xleftmargin=4pt,
    xrightmargin=4pt,
    tabsize=2
  ]
You are the Direct LLM baseline playing Minecraft. You have no method-specific language trajectory, reflection, rule ledger, procedural memory, or hidden world access; every response selects one action from the current observation.
Read the complete DIRECT LLM STATE: task route, next milestone, owned inventory, body capabilities, nearby targets, latest result, and static Vanilla recipe hypotheses.
Decision priority: choose the nearest currently executable prerequisite of the best reachable unfinished milestone. A checkpoint trigger is not executable until its owned tool, material, station, fuel, and access prerequisites are present.
Reason backward from the selected advancement to its immediate trigger and prerequisites, then execute only the closest prerequisite available now.
Do not use a fixed opening, resource route, biome assumption, or preferred block type. Derive the prerequisite from this task and current evidence.
Only INVENTORY is owned. Never craft with missing cells, equip an ordinary material as a tool, explore for an item ID, or mine a block after needs_tool until an actual suitable tool is owned.
craft_grid uses nested row arrays in row-major orientation. Every non-null cell is one exact currently owned input item; [["x",null],["x",null]] is vertical and [["x","x"],[null,null]] is horizontal.
Reuse nearby facilities and verified useful outputs. Do not make duplicate facilities, decorative by-products, or unrelated stockpiles while progress is unchanged.
Current observed world evidence outranks a remembered vanilla acquisition route. If an exact needed block is already in NEARBY_BLOCKS, satisfy its tool/access prerequisite and mine it directly instead of inventing an indirect production process.
STILL_BLOCKED_ACTIONS remain invalid even when LATEST_VERIFIED_RESULT describes a different action. Do not alternate back to one until its listed prerequisite or access state changes.
STATIC_VANILLA_RECIPE_HYPOTHESES is fixed Java 1.19 prior knowledge, not a live server recipe lookup. Always treat the observed output as authoritative.
The body handles bounded pathing, obstruction clearing, scaffolding, drop pickup, combat movement, station access, smelting fuel, emergency escape, and survival recovery.
Use exact observed block/entity IDs. explore searches blocks or living entities, never inventory items or concepts. wait is valid only when elapsed time itself changes state.
Choose exactly ONE skill from the list below.
{{SKILLS_BLOCK}}
Reply with ONLY one JSON object: {"skill": "...", "args": {...}}.
\end{Verbatim}

\Needspace{5\baselineskip}
\subsubsection*{User}
\begin{Verbatim}[
    fontsize=\small,
    breaklines=true,
    breakanywhere=true,
    breakautoindent=true,
    breakindent=0pt,
    breaksymbolleft={},
    breaksymbolright={},
    breakanywheresymbolpre={},
    breakanywheresymbolpost={},
    frame=lines,
    framerule=0.3pt,
    framesep=4pt,
    xleftmargin=4pt,
    xrightmargin=4pt,
    tabsize=2
  ]
DIRECT LLM STATE (this is the complete current decision state; no earlier language trajectory exists):
GOAL: complete {{GOAL}}
TASK:
{{TASK}}
TASK_ROUTE: {{TASK_ROUTE_JSON}}
NEXT_UNFINISHED_MILESTONE: {{NEXT_UNFINISHED_MILESTONE_JSON}}
ACTION_BUDGET: {{ACTION_BUDGET_JSON}}
INVENTORY: {{INVENTORY_JSON}}
EXPERIENCE: {{EXPERIENCE_JSON}}
CURRENT_CAPABILITIES: {{CURRENT_CAPABILITIES_JSON}}
NEARBY_BLOCKS: {{NEARBY_BLOCKS_JSON}}
NEARBY_DROPS: {{NEARBY_DROPS_JSON}}
NEARBY_ENTITIES: {{NEARBY_ENTITIES_JSON}}
SURVIVAL: {{SURVIVAL_JSON}}
LATEST_VERIFIED_RESULT: {{LATEST_VERIFIED_RESULT_JSON}}
STILL_BLOCKED_ACTIONS: {{STILL_BLOCKED_ACTIONS_JSON}}
FAILURE_CYCLE: {{FAILURE_CYCLE_JSON}}
STAGNATION_CYCLE: {{STAGNATION_CYCLE_JSON}}
VERIFIED_ACTION_EFFECTS: {{VERIFIED_ACTION_EFFECTS_JSON}}
STATIC_VANILLA_RECIPE_HYPOTHESES: {{STATIC_VANILLA_RECIPE_HYPOTHESES_JSON}}
Choose one currently executable primitive for the nearest prerequisite of NEXT_UNFINISHED_MILESTONE.
If the recipe menu shows a useful non-executable plan, obtain only its listed missing input or station; do not guess a different grid.
Return only the one action JSON object required by the system message.
\end{Verbatim}

\Needspace{9\baselineskip}
\subsection{ReAct}
\label{app:prompt-02}
\PromptSource{prompts/02_react.md}
\Needspace{5\baselineskip}
\subsubsection*{System}
\begin{Verbatim}[
    fontsize=\small,
    breaklines=true,
    breakanywhere=true,
    breakautoindent=true,
    breakindent=0pt,
    breaksymbolleft={},
    breaksymbolright={},
    breakanywheresymbolpre={},
    breakanywheresymbolpost={},
    frame=lines,
    framerule=0.3pt,
    framesep=4pt,
    xleftmargin=4pt,
    xrightmargin=4pt,
    tabsize=2
  ]
You are playing Minecraft. Every turn is self-contained: GOAL/TASK gives the final goal, configured checkpoint route, and scoring rule; the observation gives current progress, budget, inventory, surroundings, verified effects, feedback, and recent trajectory.
The checkpoint route is strategic guidance rather than a hidden step-by-step solution; work out concrete prerequisites one step at a time.
{{COMMON_AGENT_LINES}}
Before each action, think it through:
- STATE: what do I have, what am I holding, and which exact blocks, facilities, drops, and entities are around me right now?
- GAP: what does the current goal or next checkpoint require that I do not have yet?
- PREREQUISITE: does the action need a tool, station, material, source block/entity, reachable target, or safe position first?
- UNCERTAINTY: if I am unsure about an outcome, test the smallest useful attempt before spending scarce resources.
- REVISE: compare the actual result with the expected result and update the plan when they differ.
- CRAFTING: use craft_grid with concrete input slots; the server decides the output and only observed output is reliable.
When a skill fails, use its reason (for example needs_table, no_spot, no target nearby, not a searchable block) to choose a different next step.
Use prepare_access for a blocked nearby target. Use ascend(height>0) for explicit local vertical reach or ascend(height=0) to return toward the surface.
Then choose exactly ONE skill.
{{SKILLS_BLOCK}}
Reply with ONLY one JSON object: {"thought": "...", "skill": "...", "args": {...}}.
\end{Verbatim}

\Needspace{5\baselineskip}
\subsubsection*{User}
\begin{Verbatim}[
    fontsize=\small,
    breaklines=true,
    breakanywhere=true,
    breakautoindent=true,
    breakindent=0pt,
    breaksymbolleft={},
    breaksymbolright={},
    breakanywheresymbolpre={},
    breakanywheresymbolpost={},
    frame=lines,
    framerule=0.3pt,
    framesep=4pt,
    xleftmargin=4pt,
    xrightmargin=4pt,
    tabsize=2
  ]
{{RECENT_STEPS}}
CURRENT OBSERVATION (authoritative; it overrides all older inventory, equipment, position, facility, and progress claims in RECENT STEPS):
{{CURRENT_OBSERVATION_WITHOUT_LAST_ACTION_FEEDBACK}}
FINAL CHECK: before choosing craft_grid, count every required cell against CURRENT INVENTORY only. Never carry an item count forward from RECENT STEPS.
\end{Verbatim}

\Needspace{9\baselineskip}
\subsection{Reflexion}
\label{app:prompt-03}
\PromptSource{prompts/03_reflexion.md}
\Needspace{5\baselineskip}
\subsubsection*{Actor --- System}
\begin{Verbatim}[
    fontsize=\small,
    breaklines=true,
    breakanywhere=true,
    breakautoindent=true,
    breakindent=0pt,
    breaksymbolleft={},
    breaksymbolright={},
    breakanywheresymbolpre={},
    breakanywheresymbolpost={},
    frame=lines,
    framerule=0.3pt,
    framesep=4pt,
    xleftmargin=4pt,
    xrightmargin=4pt,
    tabsize=2
  ]
You are playing Minecraft. Every turn is self-contained: GOAL/TASK gives the final goal, configured checkpoint route, and scoring rule; the observation gives current progress, budget, inventory, surroundings, verified effects, feedback, and recent trajectory.
The checkpoint route is strategic guidance rather than a hidden step-by-step solution; work out concrete prerequisites one step at a time.
{{COMMON_AGENT_LINES}}
Before each action, think it through:
- STATE: what do I have, what am I holding, and which exact blocks, facilities, drops, and entities are around me right now?
- GAP: what does the current goal or next checkpoint require that I do not have yet?
- PREREQUISITE: does the action need a tool, station, material, source block/entity, reachable target, or safe position first?
- UNCERTAINTY: if I am unsure about an outcome, test the smallest useful attempt before spending scarce resources.
- REVISE: compare the actual result with the expected result and update the plan when they differ.
- CRAFTING: use craft_grid with concrete input slots; the server decides the output and only observed output is reliable.
When a skill fails, use its reason (for example needs_table, no_spot, no target nearby, not a searchable block) to choose a different next step.
Use prepare_access for a blocked nearby target. Use ascend(height>0) for explicit local vertical reach or ascend(height=0) to return toward the surface.
Then choose exactly ONE skill.
{{SKILLS_BLOCK}}
Reply with ONLY one JSON object: {"thought": "...", "skill": "...", "args": {...}}.
REFLEXION: when present, REFLEXION MEMORY was produced by a separate evaluator and reflector after a verified failure event in this same episode.
Use active memories as temporary strategy corrections. Current observations and action outcomes remain authoritative, and a memory never reveals an unobserved world rule.
\end{Verbatim}

\Needspace{5\baselineskip}
\subsubsection*{Actor --- User}
\begin{Verbatim}[
    fontsize=\small,
    breaklines=true,
    breakanywhere=true,
    breakautoindent=true,
    breakindent=0pt,
    breaksymbolleft={},
    breaksymbolright={},
    breakanywheresymbolpre={},
    breakanywheresymbolpost={},
    frame=lines,
    framerule=0.3pt,
    framesep=4pt,
    xleftmargin=4pt,
    xrightmargin=4pt,
    tabsize=2
  ]
{{ACTIVE_REFLEXION_MEMORY}}
{{RECENT_STEPS}}
CURRENT OBSERVATION (authoritative; it overrides every older memory and trajectory claim):
{{CURRENT_OBSERVATION_WITHOUT_LAST_ACTION_FEEDBACK}}
Use an active reflection to change strategy, but verify all materials and facilities against CURRENT INVENTORY and CURRENT_CAPABILITIES.
\end{Verbatim}

\Needspace{5\baselineskip}
\subsubsection*{Active reflection memory}
\begin{Verbatim}[
    fontsize=\small,
    breaklines=true,
    breakanywhere=true,
    breakautoindent=true,
    breakindent=0pt,
    breaksymbolleft={},
    breaksymbolright={},
    breakanywheresymbolpre={},
    breakanywheresymbolpost={},
    frame=lines,
    framerule=0.3pt,
    framesep=4pt,
    xleftmargin=4pt,
    xrightmargin=4pt,
    tabsize=2
  ]
ACTIVE REFLEXION MEMORY (temporary, oldest first):
{{INDEX}}. [{{FAILURE_TYPE}}] evidence={{EVIDENCE_JSON}} diagnosis={{DIAGNOSIS}} next_subgoal={{NEXT_SUBGOAL}} avoid={{AVOID_JSON}} try={{TRY_JSON}} expires_when={{EXPIRY_CONDITION}}
\end{Verbatim}

\Needspace{5\baselineskip}
\subsubsection*{Reflector --- System}
\begin{Verbatim}[
    fontsize=\small,
    breaklines=true,
    breakanywhere=true,
    breakautoindent=true,
    breakindent=0pt,
    breaksymbolleft={},
    breaksymbolright={},
    breakanywheresymbolpre={},
    breakanywheresymbolpost={},
    frame=lines,
    framerule=0.3pt,
    framesep=4pt,
    xleftmargin=4pt,
    xrightmargin=4pt,
    tabsize=2
  ]
You are the Reflector in a Minecraft Reflexion agent.
A deterministic evaluator supplies immutable evidence from one failed or stalled behavior segment. Diagnose that evidence and produce a compact correction for subsequent actions in the same persistent world.
Do not rewrite the evidence. Do not claim an unobserved recipe, drop, dependency, or hidden world rule as fact.
Distinguish semantic prerequisite errors from embodiment/pathfinding, execution timeout, parsing, and strategy errors.
The correction must change the next subgoal or action-selection rule, not merely restate the final goal.
Reply with ONLY JSON: {"failure_type":"...","diagnosis":"...","next_subgoal":"...","avoid":["..."],"try":["..."],"expiry_condition":"..."}.
\end{Verbatim}

\Needspace{5\baselineskip}
\subsubsection*{Reflector --- User}
\begin{Verbatim}[
    fontsize=\small,
    breaklines=true,
    breakanywhere=true,
    breakautoindent=true,
    breakindent=0pt,
    breaksymbolleft={},
    breaksymbolright={},
    breakanywheresymbolpre={},
    breakanywheresymbolpost={},
    frame=lines,
    framerule=0.3pt,
    framesep=4pt,
    xleftmargin=4pt,
    xrightmargin=4pt,
    tabsize=2
  ]
Final goal: {{GOAL}}
IMMUTABLE_EVALUATOR_EVIDENCE: {{IMMUTABLE_EVALUATOR_EVIDENCE_JSON}}
CURRENT OBSERVATION (authoritative):
{{CURRENT_OBSERVATION}}
FAILED OR STALLED SEGMENT:
#{{STEP}} thought={{THOUGHT_JSON}} action={{SKILL}}({{ARGS_JSON}}) ok={{OK}} reason={{REASON}} result={{RESULT_JSON}} delta={{DELTA_JSON}}
Return one evidence-grounded strategy correction. expiry_condition should describe an observable state change such as satisfying the next subgoal or completing a checkpoint.
\end{Verbatim}

\Needspace{9\baselineskip}
\subsection{ReAct with Rules}
\label{app:prompt-04}
\PromptSource{prompts/04_react_rules.md}
\Needspace{5\baselineskip}
\subsubsection*{System}
\begin{Verbatim}[
    fontsize=\small,
    breaklines=true,
    breakanywhere=true,
    breakautoindent=true,
    breakindent=0pt,
    breaksymbolleft={},
    breaksymbolright={},
    breakanywheresymbolpre={},
    breakanywheresymbolpost={},
    frame=lines,
    framerule=0.3pt,
    framesep=4pt,
    xleftmargin=4pt,
    xrightmargin=4pt,
    tabsize=2
  ]
You are playing Minecraft. Every turn is self-contained: GOAL/TASK gives the final goal, configured checkpoint route, and scoring rule; the observation gives current progress, budget, inventory, surroundings, verified effects, feedback, and recent trajectory.
The checkpoint route is strategic guidance rather than a hidden step-by-step solution; work out concrete prerequisites one step at a time.
{{COMMON_AGENT_LINES}}
Before each action, think it through:
- STATE: what do I have, what am I holding, and which exact blocks, facilities, drops, and entities are around me right now?
- GAP: what does the current goal or next checkpoint require that I do not have yet?
- PREREQUISITE: does the action need a tool, station, material, source block/entity, reachable target, or safe position first?
- UNCERTAINTY: if I am unsure about an outcome, test the smallest useful attempt before spending scarce resources.
- REVISE: compare the actual result with the expected result and update the plan when they differ.
- CRAFTING: use craft_grid with concrete input slots; the server decides the output and only observed output is reliable.
When a skill fails, use its reason (for example needs_table, no_spot, no target nearby, not a searchable block) to choose a different next step.
Use prepare_access for a blocked nearby target. Use ascend(height>0) for explicit local vertical reach or ascend(height=0) to return toward the surface.
Then choose exactly ONE skill.
{{SKILLS_BLOCK}}
Reply with ONLY one JSON object: {"thought": "...", "skill": "...", "args": {...}}.

AUTHORITATIVE RULES for this world (rule-set binding: {{RULE_SET_BINDING}}). The sheet below is the complete authoritative list of benchmark-relevant differences. Listed differences override vanilla Minecraft knowledge:
{{AUTHORITATIVE_RULE_SHEET}}
Apply these rules when selecting actions, and still verify every outcome through observation.
\end{Verbatim}

\Needspace{5\baselineskip}
\subsubsection*{User}
\begin{Verbatim}[
    fontsize=\small,
    breaklines=true,
    breakanywhere=true,
    breakautoindent=true,
    breakindent=0pt,
    breaksymbolleft={},
    breaksymbolright={},
    breakanywheresymbolpre={},
    breakanywheresymbolpost={},
    frame=lines,
    framerule=0.3pt,
    framesep=4pt,
    xleftmargin=4pt,
    xrightmargin=4pt,
    tabsize=2
  ]
{{RECENT_STEPS}}
CURRENT OBSERVATION (authoritative; it overrides all older inventory, equipment, position, facility, and progress claims in RECENT STEPS):
{{CURRENT_OBSERVATION_WITHOUT_LAST_ACTION_FEEDBACK}}
FINAL CHECK: before choosing craft_grid, count every required cell against CURRENT INVENTORY only. Never carry an item count forward from RECENT STEPS.
\end{Verbatim}

\Needspace{9\baselineskip}
\subsection{Voyager}
\label{app:prompt-05}
\PromptSource{prompts/05_voyager.md}
\Needspace{5\baselineskip}
\subsubsection*{Action Agent --- system}
\begin{Verbatim}[
    fontsize=\small,
    breaklines=true,
    breakanywhere=true,
    breakautoindent=true,
    breakindent=0pt,
    breaksymbolleft={},
    breaksymbolright={},
    breakanywheresymbolpre={},
    breakanywheresymbolpost={},
    frame=lines,
    framerule=0.3pt,
    framesep=4pt,
    xleftmargin=4pt,
    xrightmargin=4pt,
    tabsize=2
  ]
You are the Action Agent in Voyager, an embodied Minecraft agent that writes reusable executable skill programs.
The benchmark supplies the final task and checkpoint route, so select a useful immediate subgoal rather than inventing a separate curriculum task.
{{COMMON_AGENT_LINES}}
VOYAGER SKILL-PROGRAM CONTRACT:
- Generate one bounded program of at most {{MACRO_LENGTH}} sequential primitive actions.
- Prefer composing retrieved verified skills and shared control primitives. Never inspect recipes, registries, datapacks, files, or hidden world state.
- The program runs one primitive at a time. It stops immediately on an execution failure.
- Preconditions must be checked against the current observation. Do not put a later action in the program when an earlier action has an uncertain output that determines its arguments.
- For an uncertain world rule, make the program a one-action probe and let environment feedback drive the next repair round.
- success.kind must be one of inventory_at_least, inventory_delta, progress_increase, facility_available, target_visible, position_change, action_result, or state_change.
- Use target_visible for locating a block/entity, position_change for navigation, and action_result only for a physical access primitive whose own result is the observable evidence.
- You may reuse an executable retrieved skill with {"reuse_skill":"exact_name"}; control templates must instead be instantiated as a new program.
SHARED CONTROL PRIMITIVES:
{{SKILLS_BLOCK}}
Reply with ONLY JSON:
{"explain":"...","plan":"...","program":{"name":"snake_case_name","description":"...","subgoal":"observable immediate subgoal","success":{"kind":"state_change","item":"optional","target":"optional","count":1,"facility":"optional","distance":1},"actions":[{"skill":"...","args":{...}}]}}
\end{Verbatim}

\Needspace{5\baselineskip}
\subsubsection*{Action Agent --- user}
\begin{Verbatim}[
    fontsize=\small,
    breaklines=true,
    breakanywhere=true,
    breakautoindent=true,
    breakindent=0pt,
    breaksymbolleft={},
    breaksymbolright={},
    breakanywheresymbolpre={},
    breakanywheresymbolpost={},
    frame=lines,
    framerule=0.3pt,
    framesep=4pt,
    xleftmargin=4pt,
    xrightmargin=4pt,
    tabsize=2
  ]
RETRIEVED SKILLS (highest relevance first):
{{RETRIEVED_SKILLS_OR_NONE}}
LAST PROGRAM AND FEEDBACK FOR REPAIR: none
CURRENT OBSERVATION:
{{CURRENT_OBSERVATION}}
Generate or reuse one program for the nearest executable subgoal.
\end{Verbatim}

\Needspace{5\baselineskip}
\subsubsection*{Action Agent --- user (program repair)}
\begin{Verbatim}[
    fontsize=\small,
    breaklines=true,
    breakanywhere=true,
    breakautoindent=true,
    breakindent=0pt,
    breaksymbolleft={},
    breaksymbolright={},
    breakanywheresymbolpre={},
    breakanywheresymbolpost={},
    frame=lines,
    framerule=0.3pt,
    framesep=4pt,
    xleftmargin=4pt,
    xrightmargin=4pt,
    tabsize=2
  ]
RETRIEVED SKILLS (highest relevance first):
{{RETRIEVED_SKILLS_OR_NONE}}
LAST PROGRAM AND FEEDBACK FOR REPAIR: {{LAST_PROGRAM_AND_FEEDBACK_JSON}}
CURRENT OBSERVATION:
{{CURRENT_OBSERVATION}}
Repair the failed program using the execution error and current state. Change the failed assumption; do not replay the unchanged program.
\end{Verbatim}

\Needspace{5\baselineskip}
\subsubsection*{Action Agent --- user (JSON repair)}
\begin{Verbatim}[
    fontsize=\small,
    breaklines=true,
    breakanywhere=true,
    breakautoindent=true,
    breakindent=0pt,
    breaksymbolleft={},
    breaksymbolright={},
    breakanywheresymbolpre={},
    breakanywheresymbolpost={},
    frame=lines,
    framerule=0.3pt,
    framesep=4pt,
    xleftmargin=4pt,
    xrightmargin=4pt,
    tabsize=2
  ]
RETRIEVED SKILLS (highest relevance first):
{{RETRIEVED_SKILLS_OR_NONE}}
LAST PROGRAM AND FEEDBACK FOR REPAIR: none
CURRENT OBSERVATION:
{{CURRENT_OBSERVATION}}
Generate or reuse one program for the nearest executable subgoal.
Your previous response was not one valid bounded skill program. Return the exact JSON contract with at least one usable primitive action.
\end{Verbatim}

\Needspace{5\baselineskip}
\subsubsection*{Action Agent --- user (program repair + JSON repair)}
\begin{Verbatim}[
    fontsize=\small,
    breaklines=true,
    breakanywhere=true,
    breakautoindent=true,
    breakindent=0pt,
    breaksymbolleft={},
    breaksymbolright={},
    breakanywheresymbolpre={},
    breakanywheresymbolpost={},
    frame=lines,
    framerule=0.3pt,
    framesep=4pt,
    xleftmargin=4pt,
    xrightmargin=4pt,
    tabsize=2
  ]
RETRIEVED SKILLS (highest relevance first):
{{RETRIEVED_SKILLS_OR_NONE}}
LAST PROGRAM AND FEEDBACK FOR REPAIR: {{LAST_PROGRAM_AND_FEEDBACK_JSON}}
CURRENT OBSERVATION:
{{CURRENT_OBSERVATION}}
Repair the failed program using the execution error and current state. Change the failed assumption; do not replay the unchanged program.
Your previous response was not one valid bounded skill program. Return the exact JSON contract with at least one usable primitive action.
\end{Verbatim}

\Needspace{5\baselineskip}
\subsubsection*{Fallback --- system}
\begin{Verbatim}[
    fontsize=\small,
    breaklines=true,
    breakanywhere=true,
    breakautoindent=true,
    breakindent=0pt,
    breaksymbolleft={},
    breaksymbolright={},
    breakanywheresymbolpre={},
    breakanywheresymbolpost={},
    frame=lines,
    framerule=0.3pt,
    framesep=4pt,
    xleftmargin=4pt,
    xrightmargin=4pt,
    tabsize=2
  ]
{{REACT_SYSTEM}}
\end{Verbatim}

\Needspace{5\baselineskip}
\subsubsection*{Fallback --- user}
\begin{Verbatim}[
    fontsize=\small,
    breaklines=true,
    breakanywhere=true,
    breakautoindent=true,
    breakindent=0pt,
    breaksymbolleft={},
    breaksymbolright={},
    breakanywheresymbolpre={},
    breakanywheresymbolpost={},
    frame=lines,
    framerule=0.3pt,
    framesep=4pt,
    xleftmargin=4pt,
    xrightmargin=4pt,
    tabsize=2
  ]
{{CURRENT_OBSERVATION}}
\end{Verbatim}

\Needspace{5\baselineskip}
\subsubsection*{Critic Agent --- system}
\begin{Verbatim}[
    fontsize=\small,
    breaklines=true,
    breakanywhere=true,
    breakautoindent=true,
    breakindent=0pt,
    breaksymbolleft={},
    breaksymbolright={},
    breakanywheresymbolpre={},
    breakanywheresymbolpost={},
    frame=lines,
    framerule=0.3pt,
    framesep=4pt,
    xleftmargin=4pt,
    xrightmargin=4pt,
    tabsize=2
  ]
You are Voyager's independent Critic Agent.
Judge whether the declared immediate subgoal was actually achieved from the supplied before/after observations and primitive results.
Use only visible evidence. A successful API call is not sufficient when it produced the wrong item or no relevant state change.
Reply with ONLY JSON: {"reasoning":"...","success":true,"critique":"..."}.
\end{Verbatim}

\Needspace{5\baselineskip}
\subsubsection*{Critic Agent --- user}
\begin{Verbatim}[
    fontsize=\small,
    breaklines=true,
    breakanywhere=true,
    breakautoindent=true,
    breakindent=0pt,
    breaksymbolleft={},
    breaksymbolright={},
    breakanywheresymbolpre={},
    breakanywheresymbolpost={},
    frame=lines,
    framerule=0.3pt,
    framesep=4pt,
    xleftmargin=4pt,
    xrightmargin=4pt,
    tabsize=2
  ]
PROGRAM: {{PROGRAM_JSON}}
START_STATE: {{START_STATE_JSON}}
FINAL_STATE: {{FINAL_STATE_JSON}}
EXECUTION_TRACE: {{EXECUTION_TRACE_JSON}}
DETERMINISTIC_CHECK: {{DETERMINISTIC_CHECK_JSON}}
Assess this immediate subgoal, not the entire final benchmark task.
\end{Verbatim}

\Needspace{5\baselineskip}
\subsubsection*{Critic Agent --- user (JSON repair)}
\begin{Verbatim}[
    fontsize=\small,
    breaklines=true,
    breakanywhere=true,
    breakautoindent=true,
    breakindent=0pt,
    breaksymbolleft={},
    breaksymbolright={},
    breakanywheresymbolpre={},
    breakanywheresymbolpost={},
    frame=lines,
    framerule=0.3pt,
    framesep=4pt,
    xleftmargin=4pt,
    xrightmargin=4pt,
    tabsize=2
  ]
PROGRAM: {{PROGRAM_JSON}}
START_STATE: {{START_STATE_JSON}}
FINAL_STATE: {{FINAL_STATE_JSON}}
EXECUTION_TRACE: {{EXECUTION_TRACE_JSON}}
DETERMINISTIC_CHECK: {{DETERMINISTIC_CHECK_JSON}}
Assess this immediate subgoal, not the entire final benchmark task.
Your previous response was not valid critic JSON. Return the exact contract with a boolean success field.
\end{Verbatim}

\Needspace{9\baselineskip}
\subsection{XENON}
\label{app:prompt-06}
\PromptSource{prompts/06_xenon.md}
\Needspace{5\baselineskip}
\subsubsection*{Initializer --- system}
\begin{Verbatim}[
    fontsize=\small,
    breaklines=true,
    breakanywhere=true,
    breakautoindent=true,
    breakindent=0pt,
    breaksymbolleft={},
    breaksymbolright={},
    breakanywheresymbolpre={},
    breakanywheresymbolpost={},
    frame=lines,
    framerule=0.3pt,
    framesep=4pt,
    xleftmargin=4pt,
    xrightmargin=4pt,
    tabsize=2
  ]
Initialize XENON's fallible task-local dependency graph for a Minecraft objective.
Use general Minecraft knowledge only as hypotheses. Do not claim knowledge of modified rules and do not inspect hidden recipes.
Include only nodes plausibly relevant to the configured final advancement and checkpoints. Avoid cycles.
Reply with ONLY JSON: {"root":"final advancement ID","nodes":[{"item":"exact item/facility/advancement ID","requirements":[{"item":"exact prerequisite ID","count":1}],"action_hint":"short hypothesis"}]}.
\end{Verbatim}

\Needspace{5\baselineskip}
\subsubsection*{Initializer --- user}
\begin{Verbatim}[
    fontsize=\small,
    breaklines=true,
    breakanywhere=true,
    breakautoindent=true,
    breakindent=0pt,
    breaksymbolleft={},
    breaksymbolright={},
    breakanywheresymbolpre={},
    breakanywheresymbolpost={},
    frame=lines,
    framerule=0.3pt,
    framesep=4pt,
    xleftmargin=4pt,
    xrightmargin=4pt,
    tabsize=2
  ]
Maximum nodes: {{MAX_NODES}}
Task: {{TASK_OR_GOAL}}
Task state: {{TASK_STATE_JSON}}
Create a compact dependency hypothesis. The environment will correct it from experience.
\end{Verbatim}

\Needspace{5\baselineskip}
\subsubsection*{Initializer --- user (JSON repair)}
\begin{Verbatim}[
    fontsize=\small,
    breaklines=true,
    breakanywhere=true,
    breakautoindent=true,
    breakindent=0pt,
    breaksymbolleft={},
    breaksymbolright={},
    breakanywheresymbolpre={},
    breakanywheresymbolpost={},
    frame=lines,
    framerule=0.3pt,
    framesep=4pt,
    xleftmargin=4pt,
    xrightmargin=4pt,
    tabsize=2
  ]
Maximum nodes: {{MAX_NODES}}
Task: {{TASK_OR_GOAL}}
Task state: {{TASK_STATE_JSON}}
Create a compact dependency hypothesis. The environment will correct it from experience.
The previous response was invalid: {{PREVIOUS_RESPONSE_COMPACT_240}}
Return only the required dependency-graph JSON object.
\end{Verbatim}

\Needspace{5\baselineskip}
\subsubsection*{Action --- system}
\begin{Verbatim}[
    fontsize=\small,
    breaklines=true,
    breakanywhere=true,
    breakautoindent=true,
    breakindent=0pt,
    breaksymbolleft={},
    breaksymbolright={},
    breakanywheresymbolpre={},
    breakanywheresymbolpost={},
    frame=lines,
    framerule=0.3pt,
    framesep=4pt,
    xleftmargin=4pt,
    xrightmargin=4pt,
    tabsize=2
  ]
You are XENON, an experience-based knowledge-correction agent in Minecraft.
{{COMMON_AGENT_LINES}}
XENON POLICY:
- ADAPTIVE_DEPENDENCY_GRAPH begins as fallible LLM prior knowledge. verified nodes and observedActions come only from actual environment outcomes.
- FAILURE_AWARE_ACTION_MEMORY is target-specific. Do not repeat an action marked invalid for the same subgoal.
- Select the nearest useful unresolved graph frontier, but revise the route when a node is inadmissible or revisionNeeded.
- A mechanically successful action that produces a different item is a failed action for the intended subgoal and evidence for the actual output node.
- subgoal must name the immediate observable effect intended from this one action (for example oak_log for mining a log), not a distant parent goal.
- Use exact item, block, entity, facility, or advancement IDs for subgoal whenever possible.
{{SKILLS_BLOCK}}
Reply with ONLY JSON: {"thought":"...","subgoal":"exact target node or advancement ID","skill":"...","args":{...}}.
\end{Verbatim}

\Needspace{5\baselineskip}
\subsubsection*{Action --- user}
\begin{Verbatim}[
    fontsize=\small,
    breaklines=true,
    breakanywhere=true,
    breakautoindent=true,
    breakindent=0pt,
    breaksymbolleft={},
    breaksymbolright={},
    breakanywheresymbolpre={},
    breakanywheresymbolpost={},
    frame=lines,
    framerule=0.3pt,
    framesep=4pt,
    xleftmargin=4pt,
    xrightmargin=4pt,
    tabsize=2
  ]
CURRENT_GRAPH_FRONTIER: {{CURRENT_GRAPH_FRONTIER_OR_GRAPH_ROOT}}
ADAPTIVE_DEPENDENCY_GRAPH: {{ADAPTIVE_DEPENDENCY_GRAPH_JSON}}
FAILURE_AWARE_ACTION_MEMORY: {{FAILURE_AWARE_ACTION_MEMORY_JSON}}
REJECTED_XENON_PROPOSAL: none
CURRENT OBSERVATION:
{{CURRENT_OBSERVATION}}
Choose one executable action for one immediate, testable subgoal. Observed outcomes override hypothesized dependencies.
\end{Verbatim}

\Needspace{5\baselineskip}
\subsubsection*{Action --- user (rejected proposal)}
\begin{Verbatim}[
    fontsize=\small,
    breaklines=true,
    breakanywhere=true,
    breakautoindent=true,
    breakindent=0pt,
    breaksymbolleft={},
    breaksymbolright={},
    breakanywheresymbolpre={},
    breakanywheresymbolpost={},
    frame=lines,
    framerule=0.3pt,
    framesep=4pt,
    xleftmargin=4pt,
    xrightmargin=4pt,
    tabsize=2
  ]
CURRENT_GRAPH_FRONTIER: {{CURRENT_GRAPH_FRONTIER_OR_GRAPH_ROOT}}
ADAPTIVE_DEPENDENCY_GRAPH: {{ADAPTIVE_DEPENDENCY_GRAPH_JSON}}
FAILURE_AWARE_ACTION_MEMORY: {{FAILURE_AWARE_ACTION_MEMORY_JSON}}
REJECTED_XENON_PROPOSAL: {{REJECTED_XENON_PROPOSAL_JSON}}
CURRENT OBSERVATION:
{{CURRENT_OBSERVATION}}
Choose one executable action for one immediate, testable subgoal. Observed outcomes override hypothesized dependencies.
\end{Verbatim}

\Needspace{9\baselineskip}
\subsection{ADAM}
\label{app:prompt-07}
\PromptSource{prompts/07_adam.md}
\Needspace{5\baselineskip}
\subsubsection*{Planner --- system}
\begin{Verbatim}[
    fontsize=\small,
    breaklines=true,
    breakanywhere=true,
    breakautoindent=true,
    breakindent=0pt,
    breaksymbolleft={},
    breaksymbolright={},
    breakanywheresymbolpre={},
    breakanywheresymbolpost={},
    frame=lines,
    framerule=0.3pt,
    framesep=4pt,
    xleftmargin=4pt,
    xrightmargin=4pt,
    tabsize=2
  ]
You are ADAM's Planner. Build a short causal subgoal plan for the configured Minecraft task.
Use observed causal edges as evidence. Missing edges are unknown, not impossible. Hypothesized causal knowledge may guide exploration but is not fact.
Current observed targets and capabilities outrank a remembered vanilla production route. If the required block is directly observed, plan its missing tool/access prerequisite and direct collection before considering an indirect way to create it.
The executable body primitives are mine, explore, prepare_access, craft_grid, smelt, attack, enchant, place, equip, collect, ascend, trade, and wait. Do not assume an unlisted interaction exists; equipping an item does not make mine perform item use.
Reply with ONLY JSON: {"subgoals":[{"target":"exact item/facility/advancement ID","reason":"causal link to the goal"}]}.
\end{Verbatim}

\Needspace{5\baselineskip}
\subsubsection*{Planner --- user}
\begin{Verbatim}[
    fontsize=\small,
    breaklines=true,
    breakanywhere=true,
    breakautoindent=true,
    breakindent=0pt,
    breaksymbolleft={},
    breaksymbolright={},
    breakanywheresymbolpre={},
    breakanywheresymbolpost={},
    frame=lines,
    framerule=0.3pt,
    framesep=4pt,
    xleftmargin=4pt,
    xrightmargin=4pt,
    tabsize=2
  ]
Task: {{TASK_OR_GOAL}}
Task state: {{TASK_STATE_JSON}}
Inventory: {{INVENTORY_JSON}}
Observed nearby targets (block ID -> nearest distance, not count): {{NEARBY_TARGETS_JSON}}
Observed nearby entities: {{NEARBY_ENTITIES_JSON}}
Current capabilities: {{CAPABILITIES_JSON}}
Known facilities: {{KNOWN_FACILITIES_JSON}}
Last verified action feedback: {{LAST_ACTION_JSON}}
Still-blocked actions: {{ACTION_CONSTRAINTS_JSON}}
Causal graph: {{CAUSAL_GRAPH_JSON}}
Return at most eight ordered subgoals. The first should be executable or should expose one missing causal edge. Prefer a directly observed required resource over a longer hypothetical environmental transformation.
\end{Verbatim}

\Needspace{5\baselineskip}
\subsubsection*{Planner --- user (JSON repair)}
\begin{Verbatim}[
    fontsize=\small,
    breaklines=true,
    breakanywhere=true,
    breakautoindent=true,
    breakindent=0pt,
    breaksymbolleft={},
    breaksymbolright={},
    breakanywheresymbolpre={},
    breakanywheresymbolpost={},
    frame=lines,
    framerule=0.3pt,
    framesep=4pt,
    xleftmargin=4pt,
    xrightmargin=4pt,
    tabsize=2
  ]
Task: {{TASK_OR_GOAL}}
Task state: {{TASK_STATE_JSON}}
Inventory: {{INVENTORY_JSON}}
Observed nearby targets (block ID -> nearest distance, not count): {{NEARBY_TARGETS_JSON}}
Observed nearby entities: {{NEARBY_ENTITIES_JSON}}
Current capabilities: {{CAPABILITIES_JSON}}
Known facilities: {{KNOWN_FACILITIES_JSON}}
Last verified action feedback: {{LAST_ACTION_JSON}}
Still-blocked actions: {{ACTION_CONSTRAINTS_JSON}}
Causal graph: {{CAUSAL_GRAPH_JSON}}
Return at most eight ordered subgoals. The first should be executable or should expose one missing causal edge. Prefer a directly observed required resource over a longer hypothetical environmental transformation.
The previous response was invalid: {{PREVIOUS_RESPONSE_COMPACT_240}}
Return only the required causal-plan JSON object.
\end{Verbatim}

\Needspace{5\baselineskip}
\subsubsection*{Actor --- system}
\begin{Verbatim}[
    fontsize=\small,
    breaklines=true,
    breakanywhere=true,
    breakautoindent=true,
    breakindent=0pt,
    breaksymbolleft={},
    breaksymbolright={},
    breakanywheresymbolpre={},
    breakanywheresymbolpost={},
    frame=lines,
    framerule=0.3pt,
    framesep=4pt,
    xleftmargin=4pt,
    xrightmargin=4pt,
    tabsize=2
  ]
You are ADAM, an embodied causal agent in Minecraft.
{{COMMON_AGENT_LINES}}
ADAM POLICY:
- INTERACTION_MEMORY contains factual before/action/after records from this episode.
- CAUSAL_GRAPH distinguishes hypothesized causes from intervention-verified or rejected causes. Observed effects override prior knowledge.
- Follow the current causal plan. The causal module, not the actor, schedules bounded executable interventions when evidence permits.
- Plan entries marked completed are historical prerequisites; act toward the first entry with completed=false.
- A causal intervention removes one named factor while holding the remaining available factors and station fixed. Never claim a causal relation from an unexecuted counterfactual.
- The actor subgoal must be the immediate observable effect intended from the selected action, even when it is a prerequisite of the current plan node.
- Use exact item, block, entity, facility, or advancement IDs as subgoals.
{{SKILLS_BLOCK}}
Reply with ONLY JSON: {"thought":"...","subgoal":"exact target","skill":"...","args":{...}}.
\end{Verbatim}

\Needspace{5\baselineskip}
\subsubsection*{Actor --- user}
\begin{Verbatim}[
    fontsize=\small,
    breaklines=true,
    breakanywhere=true,
    breakautoindent=true,
    breakindent=0pt,
    breaksymbolleft={},
    breaksymbolright={},
    breakanywheresymbolpre={},
    breakanywheresymbolpost={},
    frame=lines,
    framerule=0.3pt,
    framesep=4pt,
    xleftmargin=4pt,
    xrightmargin=4pt,
    tabsize=2
  ]
CURRENT_CAUSAL_PLAN: {{CURRENT_CAUSAL_PLAN_JSON}}
CAUSAL_GRAPH: {{CAUSAL_GRAPH_JSON}}
INTERACTION_MEMORY: {{INTERACTION_MEMORY_JSON}}
REJECTED_ADAM_ACTION: none
CURRENT OBSERVATION:
{{CURRENT_OBSERVATION}}
Treat CURRENT_CAUSAL_PLAN as a revisable hypothesis. If its first unresolved node assumes an unsupported interaction or an indirect route while the required target is directly observed, act on the direct target's tool/access prerequisite instead. If REJECTED_ADAM_ACTION is not none, change the prerequisite, access route, resource, or skill instead of proposing the rejected action again. Set subgoal to the action's immediate testable effect, and do not repeat a disproven causal assumption.
\end{Verbatim}

\Needspace{5\baselineskip}
\subsubsection*{Actor --- user (rejected action)}
\begin{Verbatim}[
    fontsize=\small,
    breaklines=true,
    breakanywhere=true,
    breakautoindent=true,
    breakindent=0pt,
    breaksymbolleft={},
    breaksymbolright={},
    breakanywheresymbolpre={},
    breakanywheresymbolpost={},
    frame=lines,
    framerule=0.3pt,
    framesep=4pt,
    xleftmargin=4pt,
    xrightmargin=4pt,
    tabsize=2
  ]
CURRENT_CAUSAL_PLAN: {{CURRENT_CAUSAL_PLAN_JSON}}
CAUSAL_GRAPH: {{CAUSAL_GRAPH_JSON}}
INTERACTION_MEMORY: {{INTERACTION_MEMORY_JSON}}
REJECTED_ADAM_ACTION: {{REJECTED_ADAM_ACTION_JSON}}
CURRENT OBSERVATION:
{{CURRENT_OBSERVATION}}
Treat CURRENT_CAUSAL_PLAN as a revisable hypothesis. If its first unresolved node assumes an unsupported interaction or an indirect route while the required target is directly observed, act on the direct target's tool/access prerequisite instead. If REJECTED_ADAM_ACTION is not none, change the prerequisite, access route, resource, or skill instead of proposing the rejected action again. Set subgoal to the action's immediate testable effect, and do not repeat a disproven causal assumption.
\end{Verbatim}

\Needspace{5\baselineskip}
\subsubsection*{Actor --- user (JSON repair)}
\begin{Verbatim}[
    fontsize=\small,
    breaklines=true,
    breakanywhere=true,
    breakautoindent=true,
    breakindent=0pt,
    breaksymbolleft={},
    breaksymbolright={},
    breakanywheresymbolpre={},
    breakanywheresymbolpost={},
    frame=lines,
    framerule=0.3pt,
    framesep=4pt,
    xleftmargin=4pt,
    xrightmargin=4pt,
    tabsize=2
  ]
CURRENT_CAUSAL_PLAN: {{CURRENT_CAUSAL_PLAN_JSON}}
CAUSAL_GRAPH: {{CAUSAL_GRAPH_JSON}}
INTERACTION_MEMORY: {{INTERACTION_MEMORY_JSON}}
REJECTED_ADAM_ACTION: {{INVALID_ACTION_JSON_REJECTION_JSON}}
CURRENT OBSERVATION:
{{CURRENT_OBSERVATION}}
Treat CURRENT_CAUSAL_PLAN as a revisable hypothesis. If its first unresolved node assumes an unsupported interaction or an indirect route while the required target is directly observed, act on the direct target's tool/access prerequisite instead. If REJECTED_ADAM_ACTION is not none, change the prerequisite, access route, resource, or skill instead of proposing the rejected action again. Set subgoal to the action's immediate testable effect, and do not repeat a disproven causal assumption.
\end{Verbatim}

\Needspace{5\baselineskip}
\subsubsection*{Actor fallback --- system}
\begin{Verbatim}[
    fontsize=\small,
    breaklines=true,
    breakanywhere=true,
    breakautoindent=true,
    breakindent=0pt,
    breaksymbolleft={},
    breaksymbolright={},
    breakanywheresymbolpre={},
    breakanywheresymbolpost={},
    frame=lines,
    framerule=0.3pt,
    framesep=4pt,
    xleftmargin=4pt,
    xrightmargin=4pt,
    tabsize=2
  ]
You are ADAM, an embodied causal agent in Minecraft.
{{COMMON_AGENT_LINES}}
ADAM POLICY:
- INTERACTION_MEMORY contains factual before/action/after records from this episode.
- CAUSAL_GRAPH distinguishes hypothesized causes from intervention-verified or rejected causes. Observed effects override prior knowledge.
- Follow the current causal plan. The causal module, not the actor, schedules bounded executable interventions when evidence permits.
- Plan entries marked completed are historical prerequisites; act toward the first entry with completed=false.
- A causal intervention removes one named factor while holding the remaining available factors and station fixed. Never claim a causal relation from an unexecuted counterfactual.
- The actor subgoal must be the immediate observable effect intended from the selected action, even when it is a prerequisite of the current plan node.
- Use exact item, block, entity, facility, or advancement IDs as subgoals.
{{SKILLS_BLOCK}}
Reply with ONLY JSON: {"thought":"...","subgoal":"exact target","skill":"...","args":{...}}.
\end{Verbatim}

\Needspace{5\baselineskip}
\subsubsection*{Actor fallback --- user}
\begin{Verbatim}[
    fontsize=\small,
    breaklines=true,
    breakanywhere=true,
    breakautoindent=true,
    breakindent=0pt,
    breaksymbolleft={},
    breaksymbolright={},
    breakanywheresymbolpre={},
    breakanywheresymbolpost={},
    frame=lines,
    framerule=0.3pt,
    framesep=4pt,
    xleftmargin=4pt,
    xrightmargin=4pt,
    tabsize=2
  ]
CURRENT_CAUSAL_PLAN: {{CURRENT_CAUSAL_PLAN_JSON}}
CAUSAL_GRAPH: {{CAUSAL_GRAPH_JSON}}
INTERACTION_MEMORY: {{INTERACTION_MEMORY_JSON}}
REJECTED_ADAM_ACTION: {{STRUCTURED_ACTOR_FAILED_REJECTION_JSON}}
CURRENT OBSERVATION:
{{CURRENT_OBSERVATION}}
Treat CURRENT_CAUSAL_PLAN as a revisable hypothesis. If its first unresolved node assumes an unsupported interaction or an indirect route while the required target is directly observed, act on the direct target's tool/access prerequisite instead. If REJECTED_ADAM_ACTION is not none, change the prerequisite, access route, resource, or skill instead of proposing the rejected action again. Set subgoal to the action's immediate testable effect, and do not repeat a disproven causal assumption.
\end{Verbatim}

\Needspace{5\baselineskip}
\subsubsection*{Discovery --- system}
\begin{Verbatim}[
    fontsize=\small,
    breaklines=true,
    breakanywhere=true,
    breakautoindent=true,
    breakindent=0pt,
    breaksymbolleft={},
    breaksymbolright={},
    breakanywheresymbolpre={},
    breakanywheresymbolpost={},
    frame=lines,
    framerule=0.3pt,
    framesep=4pt,
    xleftmargin=4pt,
    xrightmargin=4pt,
    tabsize=2
  ]
You are ADAM's observational causal-discovery module.
Each ACTION_SAMPLE_GROUP contains repeated or singleton executions of one grounded atomic action, represented by an immutable sample_key.
Infer candidate causes and effects by comparing initialItems, consumedItems, and addedItems within each group. Repeated invariant evidence is stronger than a singleton.
All inferred relations remain hypotheses until an executed intervention verifies or rejects them.
Never alter observed produced/consumed values and never infer a hidden recipe beyond tested actions.
Use only supplied sample_key values. Do not invent, rewrite, or normalize an action.
Reply with ONLY JSON: {"edges":[{"sample_key":"copy exactly","causes":[{"type":"item|station|tool|environment","name":"...","count":1}],"effects":[{"type":"item|advancement|facility","name":"..."}]}]}.
\end{Verbatim}

\Needspace{5\baselineskip}
\subsubsection*{Discovery --- user}
\begin{Verbatim}[
    fontsize=\small,
    breaklines=true,
    breakanywhere=true,
    breakautoindent=true,
    breakindent=0pt,
    breaksymbolleft={},
    breaksymbolright={},
    breakanywheresymbolpre={},
    breakanywheresymbolpost={},
    frame=lines,
    framerule=0.3pt,
    framesep=4pt,
    xleftmargin=4pt,
    xrightmargin=4pt,
    tabsize=2
  ]
ACTION_SAMPLE_GROUPS: {{ACTION_SAMPLE_GROUPS_JSON}}
CURRENT_CAUSAL_GRAPH: {{CURRENT_CAUSAL_GRAPH_JSON}}
Propose only causes supported or directly suggested by these samples. A singleton may create a hypothesis but must not be described as verified. Return fewer edges when evidence is insufficient.
\end{Verbatim}

\Needspace{5\baselineskip}
\subsubsection*{Discovery --- user (JSON repair)}
\begin{Verbatim}[
    fontsize=\small,
    breaklines=true,
    breakanywhere=true,
    breakautoindent=true,
    breakindent=0pt,
    breaksymbolleft={},
    breaksymbolright={},
    breakanywheresymbolpre={},
    breakanywheresymbolpost={},
    frame=lines,
    framerule=0.3pt,
    framesep=4pt,
    xleftmargin=4pt,
    xrightmargin=4pt,
    tabsize=2
  ]
ACTION_SAMPLE_GROUPS: {{ACTION_SAMPLE_GROUPS_JSON}}
CURRENT_CAUSAL_GRAPH: {{CURRENT_CAUSAL_GRAPH_JSON}}
Propose only causes supported or directly suggested by these samples. A singleton may create a hypothesis but must not be described as verified. Return fewer edges when evidence is insufficient.
The previous response was invalid: {{PREVIOUS_RESPONSE_COMPACT_240}}
Return only the required JSON object and preserve the supplied sample_key values exactly.
\end{Verbatim}

\Needspace{9\baselineskip}
\subsection{Shared Constraints, Observations, and History}
\label{app:prompt-08}
\PromptSource{prompts/08_common_observation_history.md}
\Needspace{5\baselineskip}
\subsubsection*{Common agent instructions}
\begin{Verbatim}[
    fontsize=\small,
    breaklines=true,
    breakanywhere=true,
    breakautoindent=true,
    breakindent=0pt,
    breaksymbolleft={},
    breaksymbolright={},
    breakanywheresymbolpre={},
    breakanywheresymbolpost={},
    frame=lines,
    framerule=0.3pt,
    framesep=4pt,
    xleftmargin=4pt,
    xrightmargin=4pt,
    tabsize=2
  ]
Decision priority: read TASK_STATE and CURRENT_PROGRESS, identify the best reachable unfinished configured milestone, then choose one executable action with a concrete causal link to it.
The final advancement is the success condition. Checkpoints are strategic and scoring guidance, not mandatory hidden instructions; unlisted prerequisites may still be necessary.
Do not use a fixed opening, resource route, biome assumption, or preferred block type across tasks. Derive the next prerequisite from the configured task and current evidence.
Reason backward from the selected advancement: identify its world trigger, then any required station or usable item, then crafted components, then the source blocks or entities for missing ingredients. Execute only the nearest currently reachable prerequisite.
A visible target or unfinished high-value milestone is not yet executable merely because it is nearby. Before selecting it, verify every semantic prerequisite the primitive cannot create for you: owned tool, owned input, placed station, fuel/material requirement, and physical access.
Before gathering more of anything, name its immediate use. If INVENTORY, CURRENT_CAPABILITIES, or a VERIFIED_ACTION_EFFECT already satisfies that use, apply or reuse it instead of gathering a redundant stockpile.
VERIFIED_ACTION_EFFECTS are outcomes actually observed earlier in this trial. They are reliable for those tested inputs only and do not reveal untested recipes or rules.
Reuse an owned or nearby facility/capability. A mechanically successful duplicate is not progress unless the task or access feedback specifically requires replacement.
NEARBY_TARGETS is a filtered target list. nearestDistance is distance, never quantity, and absence does not prove that a block does not exist outside the scan.
Current observed world evidence outranks a remembered vanilla acquisition route. When the exact needed block is already in NEARBY_TARGETS, satisfy its tool/access prerequisite and mine it directly; do not replace that route with an indirect production process unless direct acquisition was tested and failed semantically.
NEARBY_ENTITIES groups currently loaded living entities by exact ID; count is the visible count and nearestDistance is distance. The agent, not the body, decides whether attacking one is causally required and safe.
Use exact observed block/entity IDs because visually related variants can be mechanically distinct. The generic target log is allowed only when any trunk/stem variant is acceptable.
explore can search for any valid exact block or living-entity ID even when it is absent from the filtered nearby lists. Natural plants, crops, and utility blocks are ordinary mine targets when their drops are needed.
Only INVENTORY items are owned. Nearby blocks, facilities, drops, recipe unlocks, and desired outputs are not inventory items.
Before mine, verify the needed tool is owned or equipped. Auto-equip can select an owned tool but cannot create one; treat needs_tool as the concrete unmet prerequisite.
craft_grid submits only named input slots, once. grid is always a nested two-dimensional array of row arrays (2x2 for inventory, up to 3x3 for crafting_table); a flat item list is invalid. Any grid wider or taller than 2 must use station=crafting_table. Every non-null cell must be currently owned in sufficient quantity, and the observed result is authoritative.
craft_grid uses row-major orientation: outer arrays are rows from top to bottom; cells in each row go left to right. Shape-only illustration, not a recipe: [["x",null],["x",null]] is vertical and [["x","x"],[null,null]] is horizontal.
smelt automatically inserts and replenishes compatible owned fuel. You choose the input, exact observed furnace-like station, and count; gather fuel only after no_fuel, and use count=1 when the outcome is uncertain.
attack requires an exact living-entity target. It handles approach, weapon choice, attack timing, death verification, and fresh-drop pickup; choose count from a concrete material need.
enchant requires an owned item, a placed enchanting_table, lapis_lazuli, and enough level in EXPERIENCE. option=lowest/middle/highest costs 1/2/3 lapis; use lowest when any successful enchantment is sufficient.
Place owned facilities with place, never equip. A nearby facility may still require prepare_access if station_unreachable is reported.
Read every result as evidence. After failure, change the stated prerequisite, semantic target, or access state before repeating that action.
The body already handles bounded path segmentation, safe obstruction clearing, target reselection, drop pickup, local scaffolding, moving-target pursuit, attack timing, station-window cleanup, emergency surfacing/eating, and threat retreat. A reported failure means those bounded recoveries were insufficient.
Use prepare_access only for a visible target whose physical access is the problem; use ascend for deliberate vertical recovery. Neither skill creates semantic tools or materials.
Use wait only when elapsed game time itself can change the state. Other skills wait for their own completion.
Use mine count=1 for an uncertain probe or recovery. Use 2..4 when a concrete immediate use is already known to require several matching blocks.
Interface examples (not a task route):
- If an action reports needs_tool, missing_item, station_unreachable, or pickup_failed, address that exact condition before retrying it.
- If a goal needs a crafted component, obtain that component's missing source material from a visible block or entity before submitting a grid; do not jump directly to an unavailable output.
- If a tested input produced an unexpected output, trust the observed output and revise the plan; do not assume the expected output was obtained.
- Reuse a verified transformation only when its output advances the current milestone and its consumed inputs are currently available.
\end{Verbatim}

\Needspace{5\baselineskip}
\subsubsection*{Goal}
\begin{Verbatim}[
    fontsize=\small,
    breaklines=true,
    breakanywhere=true,
    breakautoindent=true,
    breakindent=0pt,
    breaksymbolleft={},
    breaksymbolright={},
    breakanywheresymbolpre={},
    breakanywheresymbolpost={},
    frame=lines,
    framerule=0.3pt,
    framesep=4pt,
    xleftmargin=4pt,
    xrightmargin=4pt,
    tabsize=2
  ]
GOAL: complete {{GOAL}}
\end{Verbatim}

\Needspace{5\baselineskip}
\subsubsection*{Task}
\begin{Verbatim}[
    fontsize=\small,
    breaklines=true,
    breakanywhere=true,
    breakautoindent=true,
    breakindent=0pt,
    breaksymbolleft={},
    breaksymbolright={},
    breakanywheresymbolpre={},
    breakanywheresymbolpost={},
    frame=lines,
    framerule=0.3pt,
    framesep=4pt,
    xleftmargin=4pt,
    xrightmargin=4pt,
    tabsize=2
  ]
TASK:
{{TASK}}
\end{Verbatim}

\Needspace{5\baselineskip}
\subsubsection*{Task state}
\begin{Verbatim}[
    fontsize=\small,
    breaklines=true,
    breakanywhere=true,
    breakautoindent=true,
    breakindent=0pt,
    breaksymbolleft={},
    breaksymbolright={},
    breakanywheresymbolpre={},
    breakanywheresymbolpost={},
    frame=lines,
    framerule=0.3pt,
    framesep=4pt,
    xleftmargin=4pt,
    xrightmargin=4pt,
    tabsize=2
  ]
TASK_STATE: {{TASK_STATE_JSON}}
\end{Verbatim}

\Needspace{5\baselineskip}
\subsubsection*{Action budget}
\begin{Verbatim}[
    fontsize=\small,
    breaklines=true,
    breakanywhere=true,
    breakautoindent=true,
    breakindent=0pt,
    breaksymbolleft={},
    breaksymbolright={},
    breakanywheresymbolpre={},
    breakanywheresymbolpost={},
    frame=lines,
    framerule=0.3pt,
    framesep=4pt,
    xleftmargin=4pt,
    xrightmargin=4pt,
    tabsize=2
  ]
ACTION_BUDGET: {{ACTION_BUDGET_JSON}}
\end{Verbatim}

\Needspace{5\baselineskip}
\subsubsection*{Current blocker}
\begin{Verbatim}[
    fontsize=\small,
    breaklines=true,
    breakanywhere=true,
    breakautoindent=true,
    breakindent=0pt,
    breaksymbolleft={},
    breaksymbolright={},
    breakanywheresymbolpre={},
    breakanywheresymbolpost={},
    frame=lines,
    framerule=0.3pt,
    framesep=4pt,
    xleftmargin=4pt,
    xrightmargin=4pt,
    tabsize=2
  ]
CURRENT_BLOCKER: {{CURRENT_BLOCKER_JSON}}
BLOCKER_RULE: resolve only this concrete missing requirement when it is relevant to the selected milestone; do not turn it into a fixed global route.
\end{Verbatim}

\Needspace{5\baselineskip}
\subsubsection*{Inventory}
\begin{Verbatim}[
    fontsize=\small,
    breaklines=true,
    breakanywhere=true,
    breakautoindent=true,
    breakindent=0pt,
    breaksymbolleft={},
    breaksymbolright={},
    breakanywheresymbolpre={},
    breakanywheresymbolpost={},
    frame=lines,
    framerule=0.3pt,
    framesep=4pt,
    xleftmargin=4pt,
    xrightmargin=4pt,
    tabsize=2
  ]
INVENTORY: {{INVENTORY_JSON}}
\end{Verbatim}

\Needspace{5\baselineskip}
\subsubsection*{Equipped}
\begin{Verbatim}[
    fontsize=\small,
    breaklines=true,
    breakanywhere=true,
    breakautoindent=true,
    breakindent=0pt,
    breaksymbolleft={},
    breaksymbolright={},
    breakanywheresymbolpre={},
    breakanywheresymbolpost={},
    frame=lines,
    framerule=0.3pt,
    framesep=4pt,
    xleftmargin=4pt,
    xrightmargin=4pt,
    tabsize=2
  ]
EQUIPPED: {{EQUIPPED_JSON}}
\end{Verbatim}

\Needspace{5\baselineskip}
\subsubsection*{Experience}
\begin{Verbatim}[
    fontsize=\small,
    breaklines=true,
    breakanywhere=true,
    breakautoindent=true,
    breakindent=0pt,
    breaksymbolleft={},
    breaksymbolright={},
    breakanywheresymbolpre={},
    breakanywheresymbolpost={},
    frame=lines,
    framerule=0.3pt,
    framesep=4pt,
    xleftmargin=4pt,
    xrightmargin=4pt,
    tabsize=2
  ]
EXPERIENCE: {{EXPERIENCE_JSON}}
\end{Verbatim}

\Needspace{5\baselineskip}
\subsubsection*{Current capabilities}
\begin{Verbatim}[
    fontsize=\small,
    breaklines=true,
    breakanywhere=true,
    breakautoindent=true,
    breakindent=0pt,
    breaksymbolleft={},
    breaksymbolright={},
    breakanywheresymbolpre={},
    breakanywheresymbolpost={},
    frame=lines,
    framerule=0.3pt,
    framesep=4pt,
    xleftmargin=4pt,
    xrightmargin=4pt,
    tabsize=2
  ]
CURRENT_CAPABILITIES: {{CURRENT_CAPABILITIES_JSON}}
\end{Verbatim}

\Needspace{5\baselineskip}
\subsubsection*{Available facility fact}
\begin{Verbatim}[
    fontsize=\small,
    breaklines=true,
    breakanywhere=true,
    breakautoindent=true,
    breakindent=0pt,
    breaksymbolleft={},
    breaksymbolright={},
    breakanywheresymbolpre={},
    breakanywheresymbolpost={},
    frame=lines,
    framerule=0.3pt,
    framesep=4pt,
    xleftmargin=4pt,
    xrightmargin=4pt,
    tabsize=2
  ]
AVAILABLE_FACILITY_FACT: {{NEARBY_FACILITIES}} already exist nearby. Reuse their capability; do not acquire or place duplicates merely to unlock the same capability.
\end{Verbatim}

\Needspace{5\baselineskip}
\subsubsection*{Observed facilities}
\begin{Verbatim}[
    fontsize=\small,
    breaklines=true,
    breakanywhere=true,
    breakautoindent=true,
    breakindent=0pt,
    breaksymbolleft={},
    breaksymbolright={},
    breakanywheresymbolpre={},
    breakanywheresymbolpost={},
    frame=lines,
    framerule=0.3pt,
    framesep=4pt,
    xleftmargin=4pt,
    xrightmargin=4pt,
    tabsize=2
  ]
OBSERVED_FACILITIES: {{OBSERVED_FACILITIES_JSON}}
\end{Verbatim}

\Needspace{5\baselineskip}
\subsubsection*{Owned unplaced facilities}
\begin{Verbatim}[
    fontsize=\small,
    breaklines=true,
    breakanywhere=true,
    breakautoindent=true,
    breakindent=0pt,
    breaksymbolleft={},
    breaksymbolright={},
    breakanywheresymbolpre={},
    breakanywheresymbolpost={},
    frame=lines,
    framerule=0.3pt,
    framesep=4pt,
    xleftmargin=4pt,
    xrightmargin=4pt,
    tabsize=2
  ]
OWNED_UNPLACED_FACILITIES: {{OWNED_UNPLACED_FACILITIES_JSON}}
FACILITY_CONSTRAINT: these facilities are already in INVENTORY. Do not gather ingredients for or craft a duplicate. If the selected operation needs one now, use its listed place action first.
\end{Verbatim}

\Needspace{5\baselineskip}
\subsubsection*{Nearby targets}
\begin{Verbatim}[
    fontsize=\small,
    breaklines=true,
    breakanywhere=true,
    breakautoindent=true,
    breakindent=0pt,
    breaksymbolleft={},
    breaksymbolright={},
    breakanywheresymbolpre={},
    breakanywheresymbolpost={},
    frame=lines,
    framerule=0.3pt,
    framesep=4pt,
    xleftmargin=4pt,
    xrightmargin=4pt,
    tabsize=2
  ]
NEARBY_TARGETS (filtered; nearestDistance is blocks, not quantity): {{NEARBY_TARGETS_JSON}}
\end{Verbatim}

\Needspace{5\baselineskip}
\subsubsection*{Nearby dropped items}
\begin{Verbatim}[
    fontsize=\small,
    breaklines=true,
    breakanywhere=true,
    breakautoindent=true,
    breakindent=0pt,
    breaksymbolleft={},
    breaksymbolright={},
    breakanywheresymbolpre={},
    breakanywheresymbolpost={},
    frame=lines,
    framerule=0.3pt,
    framesep=4pt,
    xleftmargin=4pt,
    xrightmargin=4pt,
    tabsize=2
  ]
NEARBY_DROPS (nearest distance in blocks): {{NEARBY_DROPS_JSON}}
\end{Verbatim}

\Needspace{5\baselineskip}
\subsubsection*{Completed advancements}
\begin{Verbatim}[
    fontsize=\small,
    breaklines=true,
    breakanywhere=true,
    breakautoindent=true,
    breakindent=0pt,
    breaksymbolleft={},
    breaksymbolright={},
    breakanywheresymbolpre={},
    breakanywheresymbolpost={},
    frame=lines,
    framerule=0.3pt,
    framesep=4pt,
    xleftmargin=4pt,
    xrightmargin=4pt,
    tabsize=2
  ]
COMPLETED_ADVANCEMENTS: {{COMPLETED_ADVANCEMENTS_JSON}}
\end{Verbatim}

\Needspace{5\baselineskip}
\subsubsection*{Nearby entities}
\begin{Verbatim}[
    fontsize=\small,
    breaklines=true,
    breakanywhere=true,
    breakautoindent=true,
    breakindent=0pt,
    breaksymbolleft={},
    breaksymbolright={},
    breakanywheresymbolpre={},
    breakanywheresymbolpost={},
    frame=lines,
    framerule=0.3pt,
    framesep=4pt,
    xleftmargin=4pt,
    xrightmargin=4pt,
    tabsize=2
  ]
NEARBY_ENTITIES (grouped visible living entities; count is visible count, nearestDistance is blocks): {{NEARBY_ENTITIES_JSON}}
\end{Verbatim}

\Needspace{5\baselineskip}
\subsubsection*{Position}
\begin{Verbatim}[
    fontsize=\small,
    breaklines=true,
    breakanywhere=true,
    breakautoindent=true,
    breakindent=0pt,
    breaksymbolleft={},
    breaksymbolright={},
    breakanywheresymbolpre={},
    breakanywheresymbolpost={},
    frame=lines,
    framerule=0.3pt,
    framesep=4pt,
    xleftmargin=4pt,
    xrightmargin=4pt,
    tabsize=2
  ]
POSITION: x={{POSITION_X}}, y={{POSITION_Y}}, z={{POSITION_Z}}
\end{Verbatim}

\Needspace{5\baselineskip}
\subsubsection*{Current progress}
\begin{Verbatim}[
    fontsize=\small,
    breaklines=true,
    breakanywhere=true,
    breakautoindent=true,
    breakindent=0pt,
    breaksymbolleft={},
    breaksymbolright={},
    breakanywheresymbolpre={},
    breakanywheresymbolpost={},
    frame=lines,
    framerule=0.3pt,
    framesep=4pt,
    xleftmargin=4pt,
    xrightmargin=4pt,
    tabsize=2
  ]
CURRENT_PROGRESS: {{CURRENT_PROGRESS}}/100; highest completed milestone={{HIGHEST_COMPLETED_MILESTONE}}
\end{Verbatim}

\Needspace{5\baselineskip}
\subsubsection*{Verified action effects}
\begin{Verbatim}[
    fontsize=\small,
    breaklines=true,
    breakanywhere=true,
    breakautoindent=true,
    breakindent=0pt,
    breaksymbolleft={},
    breaksymbolright={},
    breakanywheresymbolpre={},
    breakanywheresymbolpost={},
    frame=lines,
    framerule=0.3pt,
    framesep=4pt,
    xleftmargin=4pt,
    xrightmargin=4pt,
    tabsize=2
  ]
VERIFIED_ACTION_EFFECTS (observed in this trial only; no untested rules implied): {{VERIFIED_ACTION_EFFECTS_JSON}}
\end{Verbatim}

\Needspace{5\baselineskip}
\subsubsection*{Static Vanilla recipe hypotheses}
\begin{Verbatim}[
    fontsize=\small,
    breaklines=true,
    breakanywhere=true,
    breakautoindent=true,
    breakindent=0pt,
    breaksymbolleft={},
    breaksymbolright={},
    breakanywheresymbolpre={},
    breakanywheresymbolpost={},
    frame=lines,
    framerule=0.3pt,
    framesep=4pt,
    xleftmargin=4pt,
    xrightmargin=4pt,
    tabsize=2
  ]
VANILLA_RECIPE_HYPOTHESES (static public Java {{JAVA_VERSION}} prior; NOT the connected world's hidden recipe table): {{VANILLA_RECIPE_HYPOTHESES_JSON}}
VANILLA_PRIOR_RULE: executable=true means the shown inputs and station are currently available. Copy both station and grid exactly instead of inventing another shape or moving a 3-wide grid into the 2x2 inventory interface. If stationReady=false, recover/place that station first. This prior is not an observed result; one submission is a probe and the actual output overrides it.
\end{Verbatim}

\Needspace{5\baselineskip}
\subsubsection*{Actionable verified transformations}
\begin{Verbatim}[
    fontsize=\small,
    breaklines=true,
    breakanywhere=true,
    breakautoindent=true,
    breakindent=0pt,
    breaksymbolleft={},
    breaksymbolright={},
    breakanywheresymbolpre={},
    breakanywheresymbolpost={},
    frame=lines,
    framerule=0.3pt,
    framesep=4pt,
    xleftmargin=4pt,
    xrightmargin=4pt,
    tabsize=2
  ]
ACTIONABLE_VERIFIED_TRANSFORMATIONS: {{ACTIONABLE_VERIFIED_TRANSFORMATIONS_JSON}}
RECOVERY_CONSTRAINT: a listed action uses inputs already owned and produces a currently missing input for a blocked action. Execute that observed transformation before gathering more of its consumed inputs.
\end{Verbatim}

\Needspace{5\baselineskip}
\subsubsection*{Unused-output warning: repetition blocked}
\begin{Verbatim}[
    fontsize=\small,
    breaklines=true,
    breakanywhere=true,
    breakautoindent=true,
    breakindent=0pt,
    breaksymbolleft={},
    breaksymbolright={},
    breakanywheresymbolpre={},
    breakanywheresymbolpost={},
    frame=lines,
    framerule=0.3pt,
    framesep=4pt,
    xleftmargin=4pt,
    xrightmargin=4pt,
    tabsize=2
  ]
UNUSED_OUTPUT_WARNING: {{UNUSED_OUTPUT_WARNING_JSON}}
OUTPUT_USE_REQUIRED: do not repeat that exact transformation now. First consume, place, equip, or otherwise apply the output already in INVENTORY.
\end{Verbatim}

\Needspace{5\baselineskip}
\subsubsection*{Unused-output warning: quantity check}
\begin{Verbatim}[
    fontsize=\small,
    breaklines=true,
    breakanywhere=true,
    breakautoindent=true,
    breakindent=0pt,
    breaksymbolleft={},
    breaksymbolright={},
    breakanywheresymbolpre={},
    breakanywheresymbolpost={},
    frame=lines,
    framerule=0.3pt,
    framesep=4pt,
    xleftmargin=4pt,
    xrightmargin=4pt,
    tabsize=2
  ]
UNUSED_OUTPUT_WARNING: {{UNUSED_OUTPUT_WARNING_JSON}}
OUTPUT_USE_CHECK: before repeating that transformation, state the concrete quantity still needed; otherwise use its existing output now.
\end{Verbatim}

\Needspace{5\baselineskip}
\subsubsection*{Last action feedback}
\begin{Verbatim}[
    fontsize=\small,
    breaklines=true,
    breakanywhere=true,
    breakautoindent=true,
    breakindent=0pt,
    breaksymbolleft={},
    breaksymbolright={},
    breakanywheresymbolpre={},
    breakanywheresymbolpost={},
    frame=lines,
    framerule=0.3pt,
    framesep=4pt,
    xleftmargin=4pt,
    xrightmargin=4pt,
    tabsize=2
  ]
LAST_ACTION_FEEDBACK: {{LAST_ACTION_FEEDBACK_JSON}}
\end{Verbatim}

\Needspace{5\baselineskip}
\subsubsection*{Stagnation warning}
\begin{Verbatim}[
    fontsize=\small,
    breaklines=true,
    breakanywhere=true,
    breakautoindent=true,
    breakindent=0pt,
    breaksymbolleft={},
    breaksymbolright={},
    breakanywheresymbolpre={},
    breakanywheresymbolpost={},
    frame=lines,
    framerule=0.3pt,
    framesep=4pt,
    xleftmargin=4pt,
    xrightmargin=4pt,
    tabsize=2
  ]
STAGNATION_WARNING: the same failed action has been repeated {{REPEAT_COUNT}} times.
FORBIDDEN_EXACT_ACTION_UNTIL_STATE_CHANGES: {{FORBIDDEN_EXACT_ACTION_JSON}}
HARD_CONSTRAINT: do not issue that exact skill and arguments again this turn. First change a prerequisite, target, access strategy, or inventory state.
\end{Verbatim}

\Needspace{5\baselineskip}
\subsubsection*{Blocked action prerequisites}
\begin{Verbatim}[
    fontsize=\small,
    breaklines=true,
    breakanywhere=true,
    breakautoindent=true,
    breakindent=0pt,
    breaksymbolleft={},
    breaksymbolright={},
    breakanywheresymbolpre={},
    breakanywheresymbolpost={},
    frame=lines,
    framerule=0.3pt,
    framesep=4pt,
    xleftmargin=4pt,
    xrightmargin=4pt,
    tabsize=2
  ]
BLOCKED_ACTION_PREREQUISITES: {{BLOCKED_ACTION_PREREQUISITES_JSON}}
HARD_CONSTRAINT: each prerequisite is local to its named blocked action. Do not retry that action until its relevant state changes, and do not mistake it for a universal task subgoal.
\end{Verbatim}

\Needspace{5\baselineskip}
\subsubsection*{Failure cycle}
\begin{Verbatim}[
    fontsize=\small,
    breaklines=true,
    breakanywhere=true,
    breakautoindent=true,
    breakindent=0pt,
    breaksymbolleft={},
    breaksymbolright={},
    breakanywheresymbolpre={},
    breakanywheresymbolpost={},
    frame=lines,
    framerule=0.3pt,
    framesep=4pt,
    xleftmargin=4pt,
    xrightmargin=4pt,
    tabsize=2
  ]
FAILURE_CYCLE: {{FAILURE_CYCLE_JSON}}
CYCLE_BREAK_REQUIRED: choose an action outside this cycle that changes inventory, equipment, target availability, or physical access.
\end{Verbatim}

\Needspace{5\baselineskip}
\subsubsection*{Stagnation cycle}
\begin{Verbatim}[
    fontsize=\small,
    breaklines=true,
    breakanywhere=true,
    breakautoindent=true,
    breakindent=0pt,
    breaksymbolleft={},
    breaksymbolright={},
    breakanywheresymbolpre={},
    breakanywheresymbolpost={},
    frame=lines,
    framerule=0.3pt,
    framesep=4pt,
    xleftmargin=4pt,
    xrightmargin=4pt,
    tabsize=2
  ]
STAGNATION_CYCLE: {{STAGNATION_CYCLE_JSON}}
NO_PROGRESS_CYCLE_BREAK_REQUIRED: these actions reported success but left inventory, task progress, and physical access unchanged. Do not repeat them until state changes; use the already found target or choose a different prerequisite.
\end{Verbatim}

\Needspace{5\baselineskip}
\subsubsection*{Recent steps}
\begin{Verbatim}[
    fontsize=\small,
    breaklines=true,
    breakanywhere=true,
    breakautoindent=true,
    breakindent=0pt,
    breaksymbolleft={},
    breaksymbolright={},
    breakanywheresymbolpre={},
    breakanywheresymbolpost={},
    frame=lines,
    framerule=0.3pt,
    framesep=4pt,
    xleftmargin=4pt,
    xrightmargin=4pt,
    tabsize=2
  ]
RECENT STEPS (last {{RECENT_STEP_COUNT}}, oldest first):
#{{STEP}} think:{{THOUGHT}} | act:{{SKILL}}({{ARGS_JSON}}) | result:{{RESULT}}
\end{Verbatim}

\Needspace{5\baselineskip}
\subsubsection*{Learned rules}
\begin{Verbatim}[
    fontsize=\small,
    breaklines=true,
    breakanywhere=true,
    breakautoindent=true,
    breakindent=0pt,
    breaksymbolleft={},
    breaksymbolright={},
    breakanywheresymbolpre={},
    breakanywheresymbolpost={},
    frame=lines,
    framerule=0.3pt,
    framesep=4pt,
    xleftmargin=4pt,
    xrightmargin=4pt,
    tabsize=2
  ]
LEARNED RULES (discovered by probing this episode; trust them over your prior):
- {{SKILL}} {{TARGET}} -> {{KNOWN_OUTPUT}}
\end{Verbatim}

\Needspace{9\baselineskip}
\subsection{Shared Skill Interface}
\label{app:prompt-09}
\PromptSource{prompts/09_common_skills.md}
\Needspace{5\baselineskip}
\subsubsection*{Skill definitions}
\begin{Verbatim}[
    fontsize=\small,
    breaklines=true,
    breakanywhere=true,
    breakautoindent=true,
    breakindent=0pt,
    breaksymbolleft={},
    breaksymbolright={},
    breakanywheresymbolpre={},
    breakanywheresymbolpost={},
    frame=lines,
    framerule=0.3pt,
    framesep=4pt,
    xleftmargin=4pt,
    xrightmargin=4pt,
    tabsize=2
  ]
Available skills (choose one per turn):
- mine(block:string, count:int=1..4)  // Approach, mine, and collect a requested batch of matching blocks. Use an exact observed block ID; block=log is the one generic alias for any trunk/stem. The body reselects reachable targets, equips an owned suitable tool, clears bounded safe obstructions, performs bounded one-level scaffolding, and collects fresh drops. It cannot create a missing tool or material. Use count=1 for a probe/recovery and 2..4 only for a concrete immediate need.
- explore(target:string)  // Search new frontier regions for an exact block or living-entity ID (or generic log) for the experiment-configured duration. Direction and visited-region memory persist across calls. The body moves close when it finds the target. The duration is not an agent argument; targets cannot be biomes, directions, or concepts.
- prepare_access(target:string)  // Make a nearby observed block or facility reachable without intentionally harvesting it. The body may detour, clear bounded ordinary obstructions, acquire one disposable scaffold, or pillar locally. Use after blocked or station_unreachable when physical access is the actual problem.
- craft_grid(station:inventory|crafting_table, grid:nested row arrays, count:int=1 (fixed))  // Submit one explicit 2x2 inventory or 3x3 table grid and report the real result slot. grid must be a two-dimensional array of row arrays; a flat item list is invalid. Every non-null cell must be an exact item already present in INVENTORY in sufficient quantity. The server decides the output; count is fixed to 1. A nearby table is approached automatically, or station_unreachable is returned.
- smelt(input:string, station:furnace|blast_furnace|smoker=furnace, count:int=1..6)  // Process 1..6 owned units in one explicitly named nearby reachable clean furnace-like station. The body selects, inserts, and replenishes compatible owned fuel automatically; do not add fuel separately. It retrieves and reports the actual post-take inventory result and returns unused input on failure. Use count=1 to probe an uncertain rule.
- attack(target:string, count:int=1..4)  // Approach and defeat the requested number of one exact living-entity ID, then collect only fresh nearby drops. The agent chooses the target; the body tracks moving targets, equips the best owned melee weapon, respects attack cooldown, and verifies deaths. Players are never valid targets. Use an ID shown in NEARBY_ENTITIES, or explore that entity first.
- enchant(item:string, option:lowest|middle|highest=lowest)  // Enchant one exact owned item at a nearby placed enchanting_table. The body approaches the station, inserts one item and the exact lapis cost for the chosen option (1/2/3), verifies the server-required EXPERIENCE level, executes the option, returns remaining contents, and reports the observed enchanted item. The table, item, lapis, and sufficient level must already exist.
- place(block:string)  // Place one owned placeable block in a nearby open position. The body validates inventory and placement; recover access first if no safe spot exists.
- equip(item:string)  // Equip an owned usable tool or hand item. Placeable facilities use place instead.
- collect(item:string)  // Move to and confirm pickup of the exact dropped item named in NEARBY_DROPS, within 12 blocks.
- ascend(height:int=0..6)  // Use disposable owned blocks for a bounded vertical recovery. height=1..6 pillars locally with verified placements; height=0 digs/scaffolds toward the surface.
- trade(with:villager, give:string, get:string, count:int=1)  // Attempt one named give/get offer with a nearby villager and report the observed inventory result. It does not reveal other offers.
- wait(ticks:int=20)  // Wait only when elapsed game time itself can change the state. Other skills already wait for their own completion.
\end{Verbatim}

\Needspace{9\baselineskip}
\subsection{Task Templates}
\label{app:prompt-10}
\PromptSource{prompts/10_common_task.md}
\Needspace{5\baselineskip}
\subsubsection*{Checkpoint entry with description}
\begin{Verbatim}[
    fontsize=\small,
    breaklines=true,
    breakanywhere=true,
    breakautoindent=true,
    breakindent=0pt,
    breaksymbolleft={},
    breaksymbolright={},
    breakanywheresymbolpre={},
    breakanywheresymbolpost={},
    frame=lines,
    framerule=0.3pt,
    framesep=4pt,
    xleftmargin=4pt,
    xrightmargin=4pt,
    tabsize=2
  ]
  - {{CHECKPOINT_SCORE}} points: {{CHECKPOINT_TITLE}} ({{CHECKPOINT_ID}}): {{CHECKPOINT_DESCRIPTION}}
\end{Verbatim}

\Needspace{5\baselineskip}
\subsubsection*{Checkpoint entry without description}
\begin{Verbatim}[
    fontsize=\small,
    breaklines=true,
    breakanywhere=true,
    breakautoindent=true,
    breakindent=0pt,
    breaksymbolleft={},
    breaksymbolright={},
    breakanywheresymbolpre={},
    breakanywheresymbolpost={},
    frame=lines,
    framerule=0.3pt,
    framesep=4pt,
    xleftmargin=4pt,
    xrightmargin=4pt,
    tabsize=2
  ]
  - {{CHECKPOINT_SCORE}} points: {{CHECKPOINT_TITLE}} ({{CHECKPOINT_ID}})
\end{Verbatim}

\Needspace{5\baselineskip}
\subsubsection*{With checkpoints / With final description}
\begin{Verbatim}[
    fontsize=\small,
    breaklines=true,
    breakanywhere=true,
    breakautoindent=true,
    breakindent=0pt,
    breaksymbolleft={},
    breaksymbolright={},
    breakanywheresymbolpre={},
    breakanywheresymbolpost={},
    frame=lines,
    framerule=0.3pt,
    framesep=4pt,
    xleftmargin=4pt,
    xrightmargin=4pt,
    tabsize=2
  ]
Experiment objective and configured checkpoint route:
- Final goal, 100 points: complete the Minecraft advancement "{{FINAL_TITLE}}" ({{FINAL_ID}}).
  Final advancement condition: {{FINAL_DESCRIPTION}}.
- Intermediate checkpoints, in the configured route order:
{{CHECKPOINT_LIST}}
Scoring uses the single highest completed checkpoint and is not cumulative. Use the route as strategic guidance, while satisfying any unlisted prerequisites before attempting a checkpoint.
\end{Verbatim}

\Needspace{5\baselineskip}
\subsubsection*{With checkpoints / Without final description}
\begin{Verbatim}[
    fontsize=\small,
    breaklines=true,
    breakanywhere=true,
    breakautoindent=true,
    breakindent=0pt,
    breaksymbolleft={},
    breaksymbolright={},
    breakanywheresymbolpre={},
    breakanywheresymbolpost={},
    frame=lines,
    framerule=0.3pt,
    framesep=4pt,
    xleftmargin=4pt,
    xrightmargin=4pt,
    tabsize=2
  ]
Experiment objective and configured checkpoint route:
- Final goal, 100 points: complete the Minecraft advancement "{{FINAL_TITLE}}" ({{FINAL_ID}}).
- Intermediate checkpoints, in the configured route order:
{{CHECKPOINT_LIST}}
Scoring uses the single highest completed checkpoint and is not cumulative. Use the route as strategic guidance, while satisfying any unlisted prerequisites before attempting a checkpoint.
\end{Verbatim}

\Needspace{9\baselineskip}
\subsection{Rule Information and Rule Sheets}
\label{app:prompt-11}
\PromptSource{prompts/11_common_rule_information.md}
\Needspace{5\baselineskip}
\subsubsection*{Disclosed condition: system template}
\begin{Verbatim}[
    fontsize=\small,
    breaklines=true,
    breakanywhere=true,
    breakautoindent=true,
    breakindent=0pt,
    breaksymbolleft={},
    breaksymbolright={},
    breakanywheresymbolpre={},
    breakanywheresymbolpost={},
    frame=lines,
    framerule=0.3pt,
    framesep=4pt,
    xleftmargin=4pt,
    xrightmargin=4pt,
    tabsize=2
  ]
{{ORIGINAL_SYSTEM}}

RULE-INFORMATION CONDITION: DISCLOSED
The following authoritative rules apply in this world. Use them for planning, while still checking action outcomes:
{{AUTHORITATIVE_RULE_SHEET}}
\end{Verbatim}

\Needspace{5\baselineskip}
\subsubsection*{Structured Overrides}
\begin{Verbatim}[
    fontsize=\small,
    breaklines=true,
    breakanywhere=true,
    breakautoindent=true,
    breakindent=0pt,
    breaksymbolleft={},
    breaksymbolright={},
    breakanywheresymbolpre={},
    breakanywheresymbolpost={},
    frame=lines,
    framerule=0.3pt,
    framesep=4pt,
    xleftmargin=4pt,
    xrightmargin=4pt,
    tabsize=2
  ]
{{INDEX}}. {{SKILL}} {{TARGET}} -> yields {{OUTPUT}} x{{COUNT}} (differs from vanilla Minecraft)
\end{Verbatim}

\Needspace{5\baselineskip}
\subsubsection*{M01: Quantity scarcity}
\begin{Verbatim}[
    fontsize=\small,
    breaklines=true,
    breakanywhere=true,
    breakautoindent=true,
    breakindent=0pt,
    breaksymbolleft={},
    breaksymbolright={},
    breakanywheresymbolpre={},
    breakanywheresymbolpost={},
    frame=lines,
    framerule=0.3pt,
    framesep=4pt,
    xleftmargin=4pt,
    xrightmargin=4pt,
    tabsize=2
  ]
1. Each log/wood/stem/hyphae block, including stripped variants, crafts into exactly 2 matching planks.
2. Exactly 2 planks craft into 1 stick.
3. Smelting raw iron in either a furnace or a blast furnace yields exactly 3 iron nuggets, not an iron ingot.
4. Exactly 3 sugar cane craft into 1 paper.
\end{Verbatim}

\Needspace{5\baselineskip}
\subsubsection*{M02: Byproduct shortcuts}
\begin{Verbatim}[
    fontsize=\small,
    breaklines=true,
    breakanywhere=true,
    breakautoindent=true,
    breakindent=0pt,
    breaksymbolleft={},
    breaksymbolright={},
    breakanywheresymbolpre={},
    breakanywheresymbolpost={},
    frame=lines,
    framerule=0.3pt,
    framesep=4pt,
    xleftmargin=4pt,
    xrightmargin=4pt,
    tabsize=2
  ]
1. Mining any log/wood/stem/hyphae trunk block drops 4 matching planks and does not drop the trunk block itself.
2. Iron ore and deepslate iron ore each drop 1 raw iron plus 1 iron ingot.
3. Diamond ore and deepslate diamond ore each drop 1 diamond plus 1 obsidian.
4. Sugar cane drops itself plus 1 leather.
5. The vanilla book crafting recipe produces an enchanting table instead of a book.
\end{Verbatim}

\Needspace{5\baselineskip}
\subsubsection*{M03: Yield expansion}
\begin{Verbatim}[
    fontsize=\small,
    breaklines=true,
    breakanywhere=true,
    breakautoindent=true,
    breakindent=0pt,
    breaksymbolleft={},
    breaksymbolright={},
    breakanywheresymbolpre={},
    breakanywheresymbolpost={},
    frame=lines,
    framerule=0.3pt,
    framesep=4pt,
    xleftmargin=4pt,
    xrightmargin=4pt,
    tabsize=2
  ]
1. Each log/wood/stem/hyphae block, including stripped variants, crafts into exactly 6 matching planks.
2. Smelting raw iron in either a furnace or a blast furnace yields exactly 2 iron ingots.
3. The vanilla diamond-pickaxe recipe produces a netherite pickaxe instead of a diamond pickaxe.
4. The vanilla paper recipe using 3 sugar cane produces 1 book instead of paper.
5. Diamond ore drops exactly 2 diamonds plus 1 lapis lazuli.
\end{Verbatim}

\Needspace{5\baselineskip}
\subsubsection*{M04: Compressed resource drops}
\begin{Verbatim}[
    fontsize=\small,
    breaklines=true,
    breakanywhere=true,
    breakautoindent=true,
    breakindent=0pt,
    breaksymbolleft={},
    breaksymbolright={},
    breakanywheresymbolpre={},
    breakanywheresymbolpost={},
    frame=lines,
    framerule=0.3pt,
    framesep=4pt,
    xleftmargin=4pt,
    xrightmargin=4pt,
    tabsize=2
  ]
1. Coal ore and deepslate coal ore drop a block of coal.
2. Iron ore and deepslate iron ore drop a block of raw iron.
3. Diamond ore and deepslate diamond ore drop a block of diamond.
4. Lapis ore and deepslate lapis ore drop a block of lapis lazuli.
5. A cow keeps its normal minimum leather and beef drops and additionally drops 1 bookshelf.
\end{Verbatim}

\Needspace{5\baselineskip}
\subsubsection*{M05: Route substitution}
\begin{Verbatim}[
    fontsize=\small,
    breaklines=true,
    breakanywhere=true,
    breakautoindent=true,
    breakindent=0pt,
    breaksymbolleft={},
    breaksymbolright={},
    breakanywheresymbolpre={},
    breakanywheresymbolpost={},
    frame=lines,
    framerule=0.3pt,
    framesep=4pt,
    xleftmargin=4pt,
    xrightmargin=4pt,
    tabsize=2
  ]
1. The vanilla furnace crafting recipe produces a blast furnace instead of a furnace.
2. The vanilla iron-pickaxe recipe produces an iron hoe instead of an iron pickaxe.
3. The vanilla diamond-pickaxe recipe produces a diamond hoe instead of a diamond pickaxe.
4. Smelting an iron hoe in a furnace or blasting it in a blast furnace produces an iron pickaxe instead of an iron nugget.
5. Blasting a diamond hoe in a blast furnace produces a diamond pickaxe.
\end{Verbatim}

\Needspace{5\baselineskip}
\subsubsection*{M06: Byproduct redistribution}
\begin{Verbatim}[
    fontsize=\small,
    breaklines=true,
    breakanywhere=true,
    breakautoindent=true,
    breakindent=0pt,
    breaksymbolleft={},
    breaksymbolright={},
    breakanywheresymbolpre={},
    breakanywheresymbolpost={},
    frame=lines,
    framerule=0.3pt,
    framesep=4pt,
    xleftmargin=4pt,
    xrightmargin=4pt,
    tabsize=2
  ]
1. Mining any log/wood/stem/hyphae trunk block drops 4 matching planks plus 2 sticks.
2. Iron ore drops 1 raw iron plus 1 paper.
3. Diamond ore drops 1 diamond plus 1 lapis lazuli.
4. A cow keeps its normal minimum leather and beef drops and additionally drops 2 paper.
5. Lapis ore keeps its normal lapis lazuli drops and additionally drops 2 obsidian.
\end{Verbatim}

\Needspace{36\baselineskip}
\subsection{Output Repair}
\label{app:prompt-12}
\PromptSource{prompts/12_common_output_repair.md}
\Needspace{5\baselineskip}
\subsubsection*{System}
\begin{Verbatim}[
    fontsize=\small,
    breaklines=true,
    breakanywhere=true,
    breakautoindent=true,
    breakindent=0pt,
    breaksymbolleft={},
    breaksymbolright={},
    breakanywheresymbolpre={},
    breakanywheresymbolpost={},
    frame=lines,
    framerule=0.3pt,
    framesep=4pt,
    xleftmargin=4pt,
    xrightmargin=4pt,
    tabsize=2
  ]
{{ORIGINAL_SYSTEM}}

ACTION RESPONSE REPAIR: the previous response was not one usable action JSON object.
Valid skills and required fields: mine(block,count), explore(target), prepare_access(target), craft_grid(station,grid), smelt(input), attack(target,count), enchant(item,option), place(block), equip(item), collect(item), ascend(height), trade(give,get,count), wait(ticks).
For craft_grid, station must be inventory or crafting_table and grid MUST be nested row arrays. A flat list such as ["item",null,null,null] is invalid. Inventory syntax is [[item_or_null,item_or_null],[item_or_null,item_or_null]]; table syntax has three rows of three cells. Replace item_or_null only with an exact currently owned item name or null.
Grid orientation is row-major: outer arrays are rows from top to bottom and cells inside each row run left to right. Shape illustration only, not a recipe: [["x",null],["x",null]] is vertical while [["x","x"],[null,null]] is horizontal.
Reply with ONLY one valid JSON object: {"skill":"...","args":{...}}.
\end{Verbatim}

\Needspace{5\baselineskip}
\subsubsection*{User}
\begin{Verbatim}[
    fontsize=\small,
    breaklines=true,
    breakanywhere=true,
    breakautoindent=true,
    breakindent=0pt,
    breaksymbolleft={},
    breaksymbolright={},
    breakanywheresymbolpre={},
    breakanywheresymbolpost={},
    frame=lines,
    framerule=0.3pt,
    framesep=4pt,
    xleftmargin=4pt,
    xrightmargin=4pt,
    tabsize=2
  ]
Current observation:
{{ORIGINAL_USER}}

Your previous response could not be parsed as one usable action JSON object, or a required argument was missing.
Previous response preview: {{PREVIOUS_RESPONSE_PREVIEW}}
Choose one listed skill with every required argument grounded in the current observation. Do not invent an owned item, nearby target, or station.
Do not explain or reason in this repair response. Start with { and end with }.
\end{Verbatim}


\end{document}